\pdftrailerid{}
\documentclass[11pt]{article}

\usepackage{microsoft-tech-report}
\usepackage{amsmath,amsfonts}
\usepackage{wrapfig}
\usepackage{capt-of}
\usepackage{hyperref}
\usepackage{booktabs}
\usepackage{array}
\usepackage{enumitem}
\usepackage{bbm}

\hypersetup{
  colorlinks=true,
  linkcolor=msftblue,
  citecolor=msftblue,
  urlcolor=msftblue,
  pdfcreator={},
  pdfproducer={},
  pdftitle={LensErase: A Foundation Model for Erase Anything},
  pdfauthor={Yunuo Chen, Xiuyu Wu, Qinhong Yang, Jinjing Zhao, Yanchen Dong, Sirui Zhang, Dong Chen, Fangyun Wei, Lei Shi, Ji Li, Dongdong Chen, Guo Lu, Yan Lu, Baining Guo}
}
\tcbset{lenseraseabstract/.style={
  enhanced,frame hidden,colback=msftcard,colframe=msftline,
  boxrule=0.5pt,arc=8pt,left=0.4cm,right=0.4cm,
  top=0.4cm,bottom=0.4cm,before skip=0pt,after skip=1em,
  drop shadow={opacity=0.06}
}}

\begin{document}
\thispagestyle{empty}

\noindent
\begin{minipage}[c]{0.5\linewidth}
  \raggedright\raisebox{-0.5\height}{\msftbrandmark}
\end{minipage}\begin{minipage}[c]{0.5\linewidth}
  \raggedleft{\msftdatefont\small\color{msftgray}October 2026}
\end{minipage}\par
\vspace{0.35em}
\noindent{\color{msftline}\rule{\linewidth}{0.8pt}\par}

\vspace{0.5em}
\begin{center}
  {\msfttitlefont\fontsize{21}{25}\selectfont\color{msftdark}
   LensErase: A Foundation Model\\for Erase Anything\par}
  \vspace{0.55em}
  {\normalsize\rmfamily\color{msftdark}
   Yunuo Chen$^{1,*}$ \hspace{0.5em}
   Xiuyu Wu$^{2,*}$ \hspace{0.5em}
   Qinhong Yang$^{3}$ \hspace{0.5em}
   Jinjing Zhao$^{4}$ \hspace{0.5em}
   Yanchen Dong$^{5}$\\[0.2em]
   Sirui Zhang$^{3}$ \hspace{0.5em}
   Dong Chen$^{6,\ddagger}$ \hspace{0.5em}
   Fangyun Wei$^{6,\dagger}$ \hspace{0.5em}
   Lei Shi$^{6}$ \hspace{0.5em}
   Ji Li$^{6}$\\[0.2em]
   Dongdong Chen$^{6}$ \hspace{0.6em}
   Guo Lu$^{1,\dagger}$ \hspace{0.6em}
   Yan Lu$^{6}$ \hspace{0.6em}
   Baining Guo$^{6}$\par}
  \vspace{0.45em}
  {\fontsize{9}{11}\selectfont\rmfamily\color{msftgray}
   $^{1}$ Shanghai Jiao Tong University \quad $^{2}$ Nankai University \quad
   $^{3}$ USTC\\
   $^{4}$ University of Sydney \quad $^{5}$ Peking University \quad $^{6}$ Microsoft\par}
  \vspace{0.25em}
  {\footnotesize\rmfamily\itshape\color{msftgray}
   $^{*}$ Equal contribution. \quad $^{\dagger}$ Corresponding authors. \quad $^{\ddagger}$ Project lead.\par}
\end{center}
\vspace{0.25em}
\noindent\begin{minipage}{\textwidth}
  \centering
\includegraphics[width=0.92\linewidth]{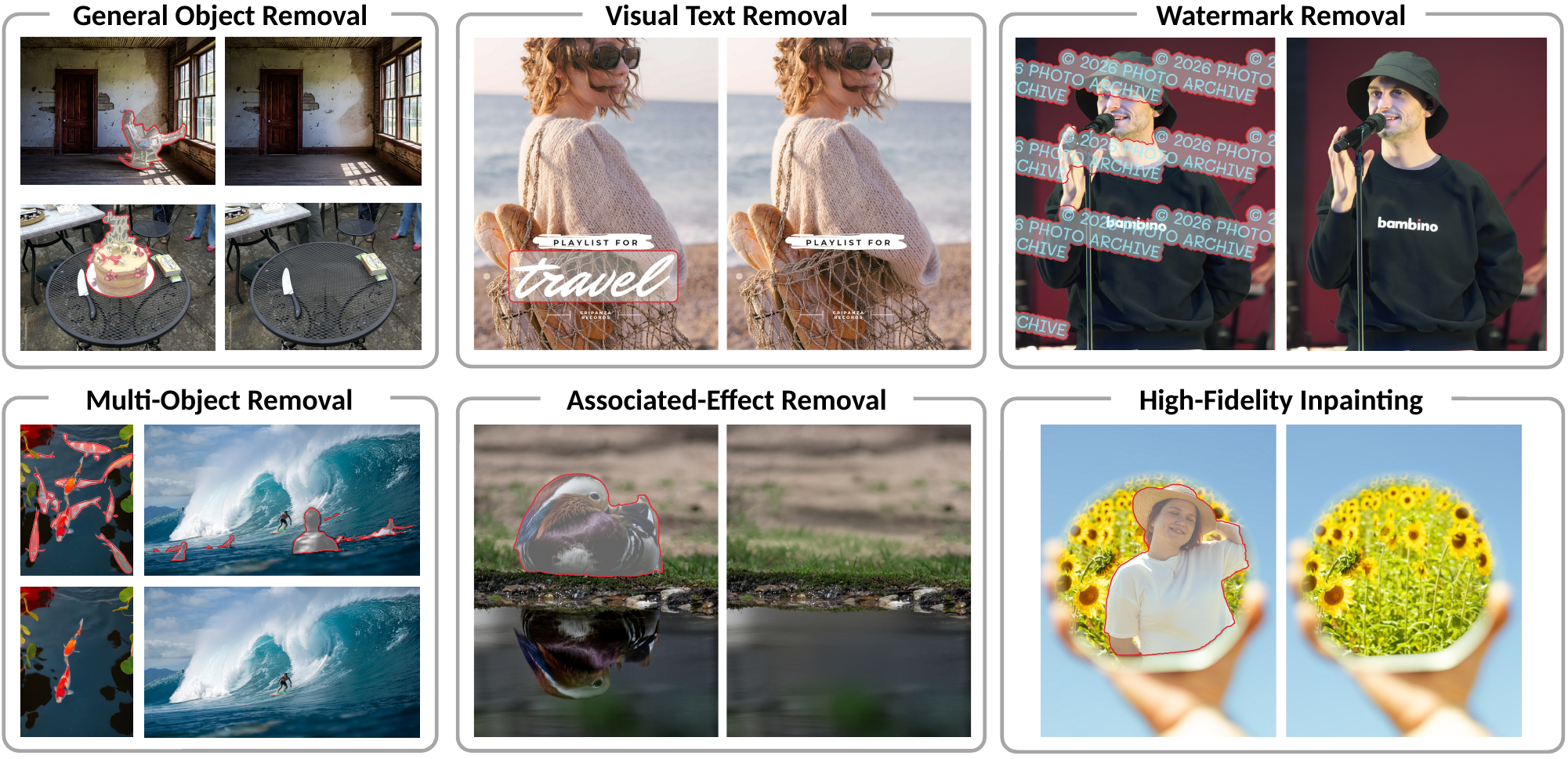}
\vspace{2pt}
  \captionof{figure}{\textbf{LensErase across diverse removal scenarios.}
  Each pair shows a source image with highlighted removal targets and the
  corresponding LensErase result. Examples cover general objects, visual text,
  watermarks, multiple objects, coupled effects, and high-fidelity inpainting.}
  \label{fig:teaser}
  \vspace{8pt}
\end{minipage}

\begin{tcolorbox}[lenseraseabstract]
\setlength{\parindent}{0pt}
\begin{abstract}
We introduce \textit{LensErase}, a general-purpose mask-based \emph{Erase Anything} foundation model designed to cover a broad spectrum of real-world erase scenarios. LensErase supports diverse objects, visual text, watermarks, associated visual effects, multiple-object removal, and a wide range of object scales, aspect ratios, and image resolutions. First, to enable such broad coverage, we construct \textit{LensErase-100M}, a large-scale dataset containing $101.07$ million paired source--mask--target examples spanning diverse erase scenarios. Second, we pretrain LensErase from the text-to-image model \textit{Lens}, adapting its strong generative prior to mask-based erasing. To efficiently incorporate the source image and erase mask, we propose \textit{Mask-Aware RoPE}, a 3D rotary positional encoding that directly injects mask information into the source-image tokens without introducing a separate mask token sequence. Finally, we refine the model through supervised fine-tuning and reinforcement learning with contract-guided, region-aware rewards, and apply \textit{TraceFusion} at inference to preserve high-fidelity content outside the effective edit region. Across six public benchmarks and the proposed \textit{LensErase-Bench}, LensErase achieves substantial improvements under both reference-based objective metrics and reference-free VLM evaluation. We hope LensErase can serve as a foundational model for erase-related applications and systems.
\end{abstract}
\vspace{0.4em}
{\small\msftmetalabel{Project Page}\url{https://lensfamily.github.io/LensErase/}\par}
\end{tcolorbox}
\section{Introduction}

Object erasing~\citep{winter2024objectdrop,jiang2025smarteraser,zhao2026precise} is a fundamental image-editing capability with broad applications in photography, content creation, graphic design, and automated visual-production pipelines. Given an input image and a user-specified region, an ideal erase model should understand \emph{what} the user intends to remove, eliminate all visual evidence associated with the target, and plausibly reconstruct content that is geometrically, semantically, and visually consistent with the surrounding scene, while preserving the remainder of the image.

However, building a single erase model that works reliably across the diverse requests encountered in real-world applications remains challenging. Handling targets across a wide range of \emph{spatial scales}, from tiny objects occupying less than $1\%$ of an image to dominant foreground objects covering nearly half of the scene, requires both precise local editing and reconstruction of large missing regions. Erasing diverse \emph{target types} poses distinct reconstruction challenges: text and logos require removing fine strokes without damaging nearby details, while translucent watermarks require recovering underlying content from blended pixels. \emph{Multi-object removal} requires clearing every selected instance while maintaining scene consistency across spatially separated edits. Removing \emph{associated visual effects}, such as reflections and shadows, requires recognizing and removing target-dependent changes that may extend beyond the selection mask~\citep{winter2024objectdrop,zhao2026precise}.

We introduce \textbf{LensErase}, a general-purpose \emph{Erase Anything} model conditioned only on a source image and an erase mask. To enable broad coverage, we construct \textbf{LensErase-100M}, a large-scale dataset containing $101.07$ million paired training examples, each consisting of a source image, an erase mask, and a target image. LensErase-100M is two to three orders of magnitude larger than representative object-removal datasets, which contain approximately $100$K--$1$M training examples~\citep{wei2025omnieraser,li2025rorem,jiang2025smarteraser}. Beyond scale, LensErase-100M is systematically constructed to cover diverse object categories, visual text, watermark types, object scales, image resolutions, and multi-object settings. We use domain-specific construction and validation pipelines to ensure reliable source--target supervision. General Data is constructed by inserting scene-compatible objects into clean real-world images and reversing the pairs, preserving the original images as removal targets. Specialized pipelines use alpha compositing for watermarks, layer rendering for posters, verified teacher edits for visual text, and foreground compositing for large objects.

LensErase builds upon a pretrained text-to-image model, \textit{Lens}~\citep{chen2026lens}, leveraging its strong generative prior to reconstruct realistic and contextually coherent content in erased regions. Since both the condition and output are visual, we initialize both the \textit{image branch} and \textit{condition branch} from the pretrained \textit{Lens} image branch. The condition branch receives the intact source image, allowing the model to identify the target and its associated effects, such as shadows and reflections. \textit{Mask-Aware RoPE} injects region identity directly into the source-image tokens without introducing a separate mask token sequence. Region-weighted supervision further emphasizes the intended edit during pretraining, producing \textbf{LensErase-Base}. We then refine the model through supervised fine-tuning on curated high-quality, high-resolution data.

While large-scale supervised training equips the model with broad erase capabilities, removal quality can still vary across stochastic samples for the same input. Some outputs achieve complete and clean removal, while others may leave target fragments or residual effects, or alter neighboring content. We therefore further optimize the supervised model with reinforcement learning (RL) using DiffusionNFT~\citep{zheng2026diffusionnft}. Region-aware rewards encourage complete removal, plausible reconstruction, and preservation of non-target content. At inference, \emph{TraceFusion} estimates the effective edit region from the selection mask and source--output residuals, and selectively restores the source image outside this region to improve fidelity. Together, RL and TraceFusion yield \textbf{LensErase}, combining reliable removal with high-fidelity preservation.

We also introduce \textbf{LensErase-Bench}, a benchmark of 500 removal cases spanning natural images, posters, watermarks, and visual text. Its reference-free visual evaluation assesses removal correctness, reconstruction quality, content preservation and visual cleanliness, complementing reference-based metrics when multiple plausible reconstructions exist. We evaluate LensErase against 12 existing methods on LensErase-Bench and six public benchmarks. LensErase achieves substantial gains under both reference-based objective metrics and reference-free VLM evaluation, demonstrating strong and consistent performance across diverse erase scenarios. We aim for LensErase to serve as a general-purpose erase model for diverse visual-generation and editing systems.

\vspace{-5pt}
\section{Data}
\vspace{-5pt}
\label{sec:data}
\subsection{Pre-Training Data: LensErase-100M}
\label{sec:pre-training-data}
\begin{wrapfigure}{r}{0.52\textwidth}
\vspace{-10pt}
    \centering
    \setlength{\abovecaptionskip}{4pt}
    \begin{minipage}[b]{0.49\linewidth}
        \centering
        \includegraphics[width=\linewidth,trim=2bp 10bp 0bp 10bp,clip]{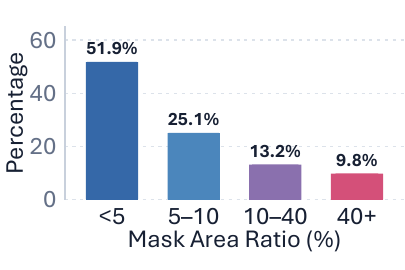}
    \end{minipage}\hfill
    \begin{minipage}[b]{0.49\linewidth}
        \centering
        \includegraphics[width=\linewidth,trim=2bp 10bp 0bp 10bp,clip]{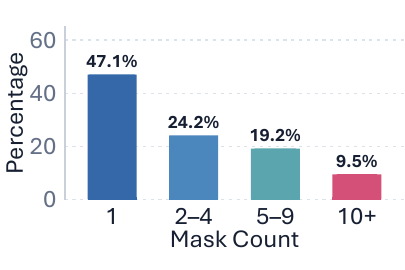}
    \end{minipage}
    \caption{\textbf{Left}: Distribution of mask area ratios.
    \textbf{Right}: Distribution of mask counts.}
    \vspace{-3mm}
    \label{fig:pretrain-data-summary}
\end{wrapfigure}

\vspace{-3pt}
We construct \textbf{LensErase-100M}, a large-scale pre-training dataset comprising 101.07M data triplets, where each triplet is represented as $(I_{\text{input}}, I_{\text{output}}, M)$, denoting the input image, erased target image, and corresponding erase mask, respectively. LensErase-100M covers five major data categories: \textit{General Data} (77.6M), \textit{Watermark Data} (9.99M), \textit{Large-Mask Data} (4.02M), \textit{Visual Text Data} (1.44M), and \textit{Poster Data} (8.02M). Figure~\ref{fig:pretrain-data-summary} shows the distributions of mask area ratios and mask counts in LensErase-100M. Below, we briefly introduce the construction pipeline for each data category, with additional details and filtering criteria provided in Appendix~\ref{app:data-construction-filtering}.

\textbf{General Data.}
We construct general object-removal pairs by adding an object to a real image and reversing the pair. We observe that general-purpose editing models perform addition more reliably than removal, which often leaves target remnants, changes unrelated backgrounds, or produces blurry fills and artifacts. Using the original real image as the clean target grounds supervision in the real-image distribution and encourages realistic outputs. The construction pipeline is as follows.
\begin{enumerate}[leftmargin=*, label=\textbf{\arabic*.}, itemsep=0pt, parsep=0pt, topsep=0pt]
    \item \textit{Object Taxonomy.} We define a hierarchical taxonomy with 15 coarse categories (e.g., Animal, Furniture, and Infrastructure), each further divided into multiple fine-grained object classes.

    \item \textit{Instruction Generation.} We first build an image-caption pool $\mathcal{D}={(I,T)}$. For each sample, we sample three distinct coarse categories. Given $(I,T)$ and the three sampled categories, GPT-4.1~\citep{openai2025gpt41} is instructed to select an object that is absent from $I$, belongs to one of these categories, and can be naturally inserted into the scene. It outputs the corresponding coarse category, object class, and an insertion instruction $P$ (e.g., ``add a dog on the grass'').

    \item \textit{Dynamic Balancing.} Sampling is dynamically adjusted according to accumulated category/object statistics. Over-represented categories or object classes are excluded from subsequent sampling to maintain data diversity and balance.

    \item \textit{Object Insertion.} Given the original image $I$ and insertion instruction $P$, we use an off-the-shelf image editing model $G_{\mathrm{add}}$ to synthesize an image with the target object inserted. We use the inserted image as the input to the erase model, $I_{\text{input}}=G_{\mathrm{add}}(I,P)$, and the original image as the desired erase output, $I_{\text{output}}=I$.

    \item \textit{Coarse-to-Fine Mask Generation.} We threshold low-frequency DCT and DINOv3~\citep{simeoni2025dinov3} feature differences between $I_{\text{input}}$ and $I_{\text{output}}$ to obtain complementary binary change maps. Their union defines a high-recall change region, while the interior of their overlap provides reliable positive-point candidates. Negative points are sampled outside the union. SAM2~\citep{ravi2024sam2} uses these prompts to produce the final object-selection mask $M$.
\end{enumerate}

\textbf{Watermark Data.} We construct watermark-removal data through the following pipeline.
\begin{enumerate}[leftmargin=*, label=\textbf{\arabic*.}, itemsep=0pt, parsep=0pt, topsep=0pt]
    \item \textit{Watermark Collection.} We collect a diverse set of commonly seen watermark patterns in real-world images, including logo watermarks, text watermarks, repeated watermarks, diagonal cross-image watermarks, stripe-style watermarks, and small corner marks.

    \item \textit{Watermark Synthesis.} For each clean image, we randomly select a watermark asset or text. We then synthesize the watermark with randomized font, color, position, scale, rotation, opacity, stroke, and overall layout to cover diverse real-world watermark appearances.

    \item \textit{Triplet Construction.} Since the watermark is rendered programmatically, we do not estimate the mask from image differences. We treat the watermarked image as the erase-model input, the clean image as the erase target, and the rendered region as the mask, i.e., $I_{\text{input}}$ is the synthesized watermarked image, $I_{\text{output}}$ is the original clean image, and $M$ is the watermark mask.
\end{enumerate}

\textbf{Large-Mask Data.}
We observe that large-mask samples are relatively under-represented in the General Data, since current general-purpose editing models struggle to insert very large objects naturally into a scene through prompting alone without occluding existing objects. We therefore construct an additional large-mask subset using a controllable copy-paste pipeline.
\begin{enumerate}[leftmargin=*, label=\textbf{\arabic*.}, itemsep=0pt, parsep=0pt, topsep=0pt]
    \item \textit{Foreground Generation.} We define a foreground-object taxonomy containing 25 coarse categories and approximately 2.5K fine-grained classes. We randomly select a fine-grained class and generate a matching object on a white background using an image generation model. We then apply BiRefNet~\citep{zheng2024birefnet} to extract the foreground object and its alpha mask.
    
    \item \textit{Copy-Paste.} We sample a clean background image $I$ and apply YOLOE~\citep{wang2025yoloe} to detect existing objects in the scene. The generated foreground is then scaled to occupy a relatively large image region and pasted preferentially into areas that do not overlap with detected objects.
    
    \item \textit{Triplet Construction.} We use the composite image after copy-paste as the erase-model input $I_{\text{input}}$, the original background image as the clean target $I_{\text{output}}$, and the transformed foreground alpha mask as the erase mask $M$.
\end{enumerate}

\textbf{Visual Text Data.}
We construct visual-text removal data through the following pipeline.
\begin{enumerate}[leftmargin=*, label=\textbf{\arabic*.}, itemsep=0pt, parsep=0pt, topsep=0pt]
    \item \textit{Text Detection.} Given an image $I$ containing visual text, we first apply PaddleOCR~\citep{cui2025paddleocr30technicalreport} to detect text regions and their bounding boxes.

    \item \textit{Text Editing.} We use an image editing model with a prompt that instructs it to erase text while preserving its carriers and other scene content, producing a teacher-edited image $\tilde{I}$.

    \item \textit{Candidate Removal Identification.} We run PaddleOCR again on $\tilde{I}$ and compare the OCR results before and after editing. Text regions detected in $I$ but no longer matched in $\tilde{I}$ are treated as candidate removal regions. We further use SSIM to exclude candidates that exhibit insufficient structural change despite the OCR mismatch.

    \item \textit{Poisson Blending.} We randomly select a subset of eligible text groups and blend the corresponding regions from $\tilde{I}$ into the original image $I$, preserving unselected text and surrounding content. Poisson blending~\citep{perez2023poisson} reduces color seams introduced by the editing model.

    \item \textit{Triplet Construction.} We use the original image with text as the erase-model input, i.e., $I_{\text{input}}=I$. The Poisson-blended image is used as the target $I_{\text{output}}$, and the union of the selected text regions, with a small dilation, forms the mask $M$. The resulting triplets are accepted as valid removal pairs after the quality filtering detailed in Appendix~\ref{app:scene-text-pipeline}.
\end{enumerate}

\textbf{Poster Data.}
We construct poster-element removal data by modeling a poster as a structured set of layers with explicit z-order.
\begin{enumerate}[leftmargin=*, label=\textbf{\arabic*.}, itemsep=0pt, parsep=0pt, topsep=0pt]
    \item \textit{Canvas and Layout Sampling.} We first sample a canvas from either a natural image or a design background, and randomly choose a canvas aspect ratio and a layout template.

    \item \textit{Element Generation.} We populate each poster with approximately 4--9 elements. The element types include title, subtitle, paragraph, label, price, date, URL, basic shape, decorative strip, logo, QR code, and chart. For each element, we randomly sample its content and attributes, including font, language, color, opacity, and other style parameters.

    \item \textit{Structured Layer Composition.} All elements are organized as a structured layer set with explicit z-order, which determines their spatial arrangement and mutual occlusion. Under a fixed background, layout, and layer configuration, we render the complete poster.

    \item \textit{Element Deletion.} We randomly remove a subset of elements while keeping the same background, layout, and remaining layer states unchanged, and render the edited poster after deletion.

    \item \textit{Triplet Construction.} We use the complete poster as the erase-model input $I_{\text{input}}$, the poster after deleting the selected elements as the target image $I_{\text{output}}$, and the visible regions of the removed elements as the mask $M$. The mask is determined according to the z-order of the layered poster, so that only the actually visible pixels of the removed elements are included.
\end{enumerate}

\vspace{-2mm}
\subsection{Supervised Fine-Tuning (SFT) and Reinforcement Learning (RL) Data}
\vspace{-1mm}
\label{sec:sft-data}

\textbf{SFT Data.} We follow the pretraining pipelines in Section~\ref{sec:pre-training-data}, prioritizing high-resolution images. All candidate triplets undergo GPT-5.6 Sol~\citep{openai2026gpt56sol} screening for mask validity, edit correctness, non-target preservation, and reconstruction quality (Appendix~\ref{app:sft_data}). We do not include Large-Mask Data in SFT, as these examples are synthetically generated and are mainly used to improve the coverage of large-mask cases during pre-training.
For Poster Data, instead of using the synthetic generation pipeline adopted during pre-training, we collect high-quality poster examples designed by professional designers, providing more realistic layouts, typography, and visual compositions. The SFT set contains 140K examples: 80K General Data, 20K Watermark Data, 20K Visual Text Data, and 20K Poster Data.

\textbf{RL Data.} We curate 19K image--mask tasks using source images disjoint from SFT. General Data contains 8K difficulty-filtered Hard/Extreme cases and 5K targeted supplements. Manual analysis of LensErase-SFT failures guides collection of new images with similar challenges. For annotated General Data tasks, GPT-5.6 Sol~\citep{openai2026gpt56sol} localizes objects, including those with associated effects. SAM2 converts their bounding boxes into masks. GPT then assesses difficulty from the validated source--mask pairs without observing erase-model outputs. Poster, Watermark, and Visual Text contribute 2K tasks each through high-resolution layered removal, controlled synthesis, and screened text removal with text-centered cropping, respectively. The model receives only $I_{\text{input}}$ and $M$. General Data requires no paired target. Poster and Watermark use deterministic clean references, whereas Visual Text uses verified weak references for rewards. Appendix~\ref{app:rl-data-pipeline} details the pipelines.

\vspace{-2mm}
\subsection{Evaluation Dataset: LensErase-Bench}
\vspace{-1mm}
\label{sec:evaluation-dataset}
\label{sec:lenserase_bench}
\textbf{LensErase-Bench.} Most existing object-removal benchmarks evaluate outputs against paired reference images obtained through synthesis or real-world capture. Synthesized references may contain artifacts, while separately captured references may differ in lighting or scene details. Since object removal also admits multiple valid outcomes, agreement with a single reference may not fully reflect removal quality. To complement reference-based metrics, we introduce \textit{LensErase-Bench}, a benchmark of 500 carefully curated removal cases evaluated without paired ground-truth target images. A VLM evaluates each output against the source image and selection mask. The Poster subset uses layered designs from Crello~\citep{yamaguchi2021canvasvae}. We also include image--mask pairs selected from existing segmentation datasets~\citep{qi2023highquality,deng2023recurrent,xie2022pyramid}. We normalize the selected images to a shorter side above 512 pixels and a longer side below 2048 pixels while preserving aspect ratio. We carefully curate the benchmark to provide balanced coverage across multiple dimensions, including application scenarios, mask area ratio, image resolution, selected-target count, and erase difficulty. These factors are jointly considered to ensure that LensErase-Bench is both diverse and representative of practical applications. Each evaluation sample consists of an input image $I_{\text{input}}$ and an erase mask $M$. Appendix~\ref{app:benchmark_construction} details the selection constraints and distributions.

\textbf{Evaluation Metrics.} We use GPT-5.6 Sol~\citep{openai2026gpt56sol} as the VLM judge to evaluate each output against the input image and selection mask along four dimensions: \emph{Removal}, the completeness of removing selected targets and their associated effects; \emph{Inpainting}, plausible completion consistent with the surrounding context; \emph{Preservation}, retention of non-target content; and \emph{Cleanliness}, absence of artifacts and unrelated additions. Each dimension receives an integer rating from $0$ to $4$, rescaled to $0$--$100$ for reporting. \emph{Overall} weights these dimensions by $40\%$, $20\%$, $20\%$, and $20\%$, respectively. We additionally report the \emph{High-Fidelity Removal Rate} (\emph{HiFi Removal}, \%), the percentage of all valid cases that achieve complete removal while meeting the quality requirements for inpainting, preservation, and visual cleanliness. Specifically, a case is successful when Removal $=4$ and Inpainting, Preservation, and Cleanliness are all $\geq3$ on the original $0$--$4$ scale. Mean scores are computed over all valid cases with equal weight per image. Detailed scoring rules and the evaluation protocol are provided in Appendix~\ref{app:benchmark_protocol}.

\vspace{-6pt}
\section{Method}
\label{sec:method}
\vspace{-6pt}

Given a source image $I_{\text{input}}$ and a binary selection mask $M$, LensErase produces an output image $\hat{I}_{\text{output}}$ that removes the selected objects and their associated visual effects, while preserving the remaining scene and plausibly completing the removed regions. Figure~\ref{fig:lens_erase_overview} shows the architecture.

\vspace{-1mm}
\subsection{Pretraining: Adapting Generation to Erasure}
\vspace{-1mm}
\label{sec:mask_conditioned_architecture}
\label{sec:supervised_training}

\begin{figure}[t]
  \centering
  \includegraphics[width=0.99\linewidth]{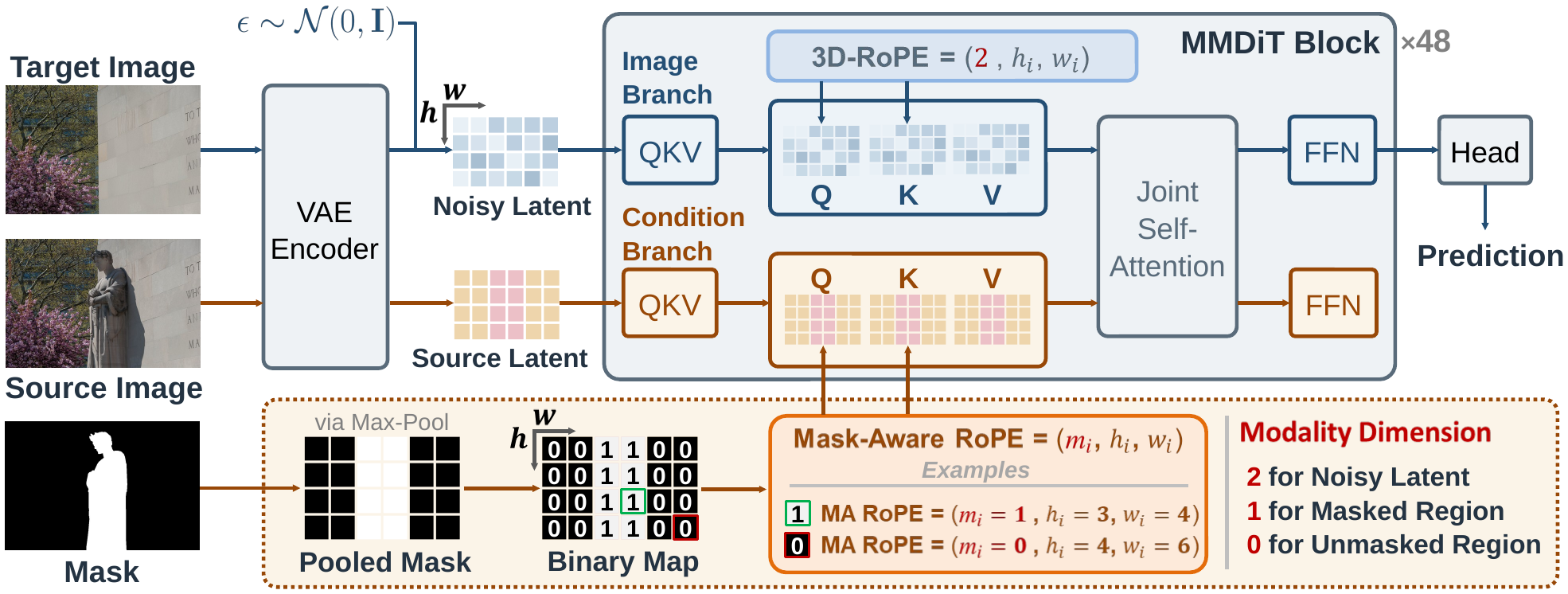}
  \vspace{-8pt}
\caption{\textbf{Architecture of LensErase.} LensErase follows a two-branch MMDiT architecture, where the image branch processes the noisy target latent and the condition branch processes the source latent. The proposed \textit{Mask-Aware RoPE} incorporates the pooled binary mask into the condition branch by encoding masked and unmasked source regions along an additional modality dimension.}

  \label{fig:lens_erase_overview}
  \vspace{-10pt}
\end{figure}

\textbf{Architecture and Initialization.}
We initialize LensErase from Lens~\citep{chen2026lens}, a text-to-image model whose MMDiT backbone contains a condition branch for the text modality and an image branch for the visual modality. LensErase follows the same backbone architecture, but both branches are initialized from the pretrained image branch of Lens. This is because, in the erase task, the condition input, including the mask $M$ and source image features, is also visual. Figure~\ref{fig:lens_erase_overview} illustrates the architecture of LensErase: the image branch processes the noisy target latent, while the condition branch processes the source latent. The erase mask $M$, with 1 denoting the masked region and 0 the unmasked region, is incorporated into the condition branch through the proposed \textit{Mask-Aware RoPE} with negligible computational overhead.

We retain the complete source image, including the selected region. The selected object's appearance and location help identify associated shadows and reflections, while semi-transparent overlays retain useful information about the underlying background. Masking out the selected region would therefore discard information useful for both effect removal and reconstruction.

\textbf{Mask-Aware RoPE.}
We apply RoPE~\citep{su2024roformer} to both noisy target tokens and source tokens. To encode the erase mask in the source tokens, we extend conventional 2D spatial RoPE to a 3D formulation by introducing an additional modality dimension. Specifically, we first downsample the binary mask $M$ to the source-token grid via max pooling, yielding $m_i\in\{0,1\}$ for each source token. We then define the positional coordinate for source tokens as
\begin{equation}
\mathbf{p}^{\mathrm{cond}}_i=(m_i,h_i,w_i),
\label{eq:mask_rope_positions}
\end{equation}
where $(h_i,w_i)$ are spatial coordinates and $m_i=1$ ($m_i=0$) denotes a masked (unmasked) source token. This directly encodes the mask into the positional representation without introducing additional mask tokens or embedding layers.
To represent the noisy target tokens within the same 3D RoPE formulation, we assign them a proxy modality coordinate of $2$:
\begin{equation}
\mathbf{p}^{\mathrm{img}}_i=(2,h_i,w_i).
\end{equation}

\textbf{Region-Weighted Flow Matching.}
We pretrain LensErase on LensErase-100M using a region-weighted flow-matching objective. Since object removal often affects only a small image region, a uniform objective can be dominated by unchanged background. Let $E$ denote the VAE encoder, with target latent $z^\star=E(I_{\text{output}})$ and source latent $z_{\text{src}}=E(I_{\text{input}})$. Following flow matching~\citep{lipman2022flow}, we construct $z_t=(1-t)z^\star+t\epsilon$ with target velocity $v^\star=\epsilon-z^\star$, where $\epsilon\sim\mathcal{N}(0,\mathbf{I})$. Conditioned on $z_{\text{src}}$ and the erase mask $M$, the model predicts the velocity $v_\theta(z_t,t,z_{\text{src}};M)$. To emphasize the intended edit, we assign larger loss weights to tokens within the masked region:
\begin{equation}
\mathcal{L}_{\mathrm{FM}}
=\mathbb{E}_{I_{\text{input}},I_{\text{output}},M,t,\epsilon}\!\left[
\frac{\sum_{i=1}^{N} \omega_i
\lVert v_{\theta,i}(z_t,t,z_{\text{src}};M)-v_i^\star\rVert_2^2}
{\sum_{i=1}^{N}\omega_i}
\right].
\label{eq:weighted_removal_fm}
\end{equation}
Here, $i$ indexes the $N$ target tokens and $\omega_i=1+(\lambda-1)\tilde{M}_i$, where $\tilde{M}$ is the dilated mask downsampled to the token grid and $\lambda=5$. The normalization reduces sensitivity of the overall loss scale to mask size. Further pretraining details and notation are provided in Appendices~\ref{app:pretraining_design} and~\ref{app:notation}.

\textbf{Supervised Fine-Tuning (SFT).}
After pretraining, we perform SFT on the curated SFT set using LoRA~\citep{hu2021lora}, while retaining the same region-weighted flow-matching objective. Training details are provided in Appendix~\ref{app:supervised_sampling}.

\vspace{-5pt}
\subsection{RL: Contract-Guided Erasure}
\label{sec:removal_rl}
\vspace{-5pt}

The SFT model already produces strong removals, but its success can vary across stochastic samples for the same image--mask input. We therefore use RL to increase the likelihood of complete removal while preserving non-target content. Figure~\ref{fig:rl_reward_pipeline} in Appendix~\ref{app:removal_reward} summarizes the reward pipeline.

\textbf{1. Constructing the Erase Contract.}
For each General Data source image $I_{\text{input}}$ and selection mask $M$, we first construct a source-grounded, candidate-independent erase contract that specifies the intended edit. GPT-5.6 Sol~\citep{openai2026gpt56sol} examines the source image, mask, and mask-centered context views to produce $\mathcal{K}=(\mathcal{T},\mathcal{E},\mathcal{O})$. The removal targets $\mathcal{T}$ specify the selected entities and key visual components whose presence would indicate incomplete removal. The associated effects $\mathcal{E}$ describe target-dependent effects, such as shadows or reflections, that should also disappear. The protected objects $\mathcal{O}$ identify nearby independent content that should remain unchanged. Thus, $\mathcal{T}$ and $\mathcal{E}$ specify what should disappear, while $\mathcal{O}$ specifies what should be preserved. The same contract is shared across all rollouts for a given image--mask condition, ensuring consistent reward evaluation.

\textbf{2. Policy Rollouts and Reward Computation.}
The SFT-initialized policy generates multiple stochastic candidates per image--mask pair, which are evaluated against the shared erase contract. GPT-5.6 Sol provides four semantic scores: \emph{Removal} assesses whether targets in $\mathcal{T}$ and effects in $\mathcal{E}$ are fully removed; \emph{Reconstruction} evaluates the plausibility and contextual consistency of synthesized content; \emph{Preservation} measures retention of protected objects in $\mathcal{O}$ and other non-target content; and \emph{Cleanliness} captures residual fragments, unwanted additions, and artifacts. We additionally compute \emph{Outside Fidelity} as source--candidate PSNR over expected unchanged regions. Reference-guided adaptations for other domains are detailed in Appendix~\ref{app:removal_reward}.

\textbf{3. Removal-Aware Reward Gating.}
A weighted average can assign a high reward to an edit that performs well on reconstruction or preservation but still exhibits clear removal failures. We therefore first aggregate the five reward components, namely \emph{Removal}, \emph{Reconstruction}, \emph{Preservation}, \emph{Cleanliness}, and \emph{Outside Fidelity}, into a base reward:
\begin{equation}
    b=\sum_{p\in\mathcal{P}}\lambda_p P_p,
    \qquad
    \sum_{p\in\mathcal{P}}\lambda_p=1,
\end{equation}
where
$\mathcal{P}=\{\mathrm{Removal},\mathrm{Reconstruction},\mathrm{Preservation},\mathrm{Cleanliness},\mathrm{OutsideFidelity}\}$,
and $P_p\in[0,1]$ denotes the normalized score of reward component $p$. We suppress the candidate index $j$. Component and subcriterion weights are renormalized over applicable terms, as detailed in Appendix~\ref{app:removal_reward}. The \emph{Removal} component is further decomposed into multiple subcriteria. Each zero-valued applicable Removal subcriterion incurs an additional penalty factor of $0.5$. Specifically, let
\begin{equation}
    n_0=\sum_{q\in\mathcal{Q}_{\mathrm{Removal}}}\mathbbm{1}[s_q=0],
\end{equation}
where $\mathcal{Q}_{\mathrm{Removal}}$ denotes the set of applicable Removal subcriteria and $s_q\in[0,1]$ is the normalized score of subcriterion $q$. The final reward is
\begin{equation}
    r=0.5^{\,n_0}b.
    \label{eq:removal_gate}
\end{equation}
This progressively penalizes candidates with multiple removal failures, reducing the extent to which strong performance on other criteria can compensate for them.

\textbf{4. Optimization.}
We optimize the SFT-initialized LensErase model with DiffusionNFT~\citep{zheng2026diffusionnft}. Implementation details are provided in Appendix~\ref{app:nft_implementation}.

\vspace{-3pt}
\subsection{Post-Processing with TraceFusion}
\label{sec:tracefuse}
\vspace{-3pt}

Although LensErase primarily modifies the selected target and associated effects, VAE reconstruction may introduce small unintended changes in otherwise unchanged regions. TraceFusion estimates the effective edit region from source--output residuals, preserves generated content there, and restores the source image elsewhere. This improves the fidelity of unchanged content while retaining intended edits that may extend beyond the input mask, such as removed shadows or reflections.

We compute pixel-wise residuals between the generated and source images after compensating for reconstruction bias in unchanged regions. On the 0--255 scale, pixels above 4 form candidate change components, of which only those containing sufficient pixels above 10 are retained to suppress small or fragmented artifacts. This allows the estimated edit region to extend beyond the input mask to associated effects such as shadows or reflections. Detailed criteria are provided in Appendix~\ref{app:tracefuse_parameters}.

We then combine the detected residual support with a dilated input mask to form the primary edit support. Because changes in dark regions can produce weak residuals in the original intensity space, we additionally apply the same detector after a shared $\gamma=0.75$ transform to both images. This gamma branch only supplements the primary support: newly detected components are retained only when they connect to the primary edit support and predominantly lie in dark regions of both images.

Finally, we feather the boundary of the resulting edit support to obtain a blending weight $\alpha\in[0,1]$. The generated image is preserved inside the estimated edit region, while the source image is restored elsewhere. Blending is performed in linear RGB:
\begin{equation}
  \hat{I}_{\text{output}}^{\mathrm{TF}}
  =
  \mathcal{C}^{-1}\!\left[
  \alpha\odot\mathcal{C}(\hat{I}_{\text{output}})
  +(1-\alpha)\odot\mathcal{C}(I_{\text{input}})
  \right],
  \label{eq:tracefusion}
\end{equation}
where $\mathcal{C}$ converts sRGB values to linear RGB.

\vspace{-5pt}
\section{Experiments}
\vspace{-5pt}
\label{sec:experiments}

\begin{table}[!t]
  \centering
  \setlength{\abovecaptionskip}{2.5pt}
  \setlength{\belowcaptionskip}{1.5pt}
  \begingroup
  \definecolor{LensRankFirst}{HTML}{F5C6C9}
  \definecolor{LensRankSecond}{HTML}{F9DCB9}
  \definecolor{LensRankThird}{HTML}{FBF1C3}
  \definecolor{LensTableRule}{HTML}{555B62}
  \definecolor{LensTableDivider}{HTML}{BDC3C9}
  \setlength{\fboxsep}{0.8pt}
  \caption{\textbf{Comparison with state-of-the-art object removal methods on six public benchmarks.}
\textit{For each benchmark, we report the three baselines with the lowest mean rank across the eight metrics together with LensErase}. Complete results for all 12 baselines are provided in Tables~\ref{tab:results-ober-test}-\ref{tab:results-text-erase} of the Appendix. Cell backgrounds mark the \colorbox{LensRankFirst}{\textbf{best}}, \colorbox{LensRankSecond}{second-best}, and \colorbox{LensRankThird}{third-best} scores among these four methods.}
    \label{tab:results-public}
  \fontsize{7.7}{9.1}\selectfont
  \setlength{\tabcolsep}{1.4pt}
  \renewcommand{\arraystretch}{0.98}
  \arrayrulecolor{LensTableRule}
  \begin{tabular}{@{}>{\raggedright\arraybackslash}m{\dimexpr\linewidth-276.9pt\relax}
    *{2}{>{\columncolor{white}[0pt][0pt]\centering\arraybackslash}m{31.5pt}}
    >{\columncolor{white}[0pt][0pt]\centering\arraybackslash}m{34pt}
    *{5}{>{\columncolor{white}[0pt][0pt]\centering\arraybackslash}m{31.5pt}}@{}}
    \toprule[0.6pt]
    \textbf{Method} & \mbox{\textbf{PSNR}\,$\uparrow$} & \mbox{\textbf{FID}\,$\downarrow$}
    & \mbox{\textbf{CMMD}\,$\downarrow$} & \mbox{\textbf{LPIPS}\,$\downarrow$}
    & \mbox{\textbf{SSIM}\,$\uparrow$} & \mbox{\textbf{DISTS}\,$\downarrow$}
    & \mbox{\textbf{CLIP-I}\,$\uparrow$} & \mbox{\textbf{DINO}\,$\uparrow$} \\
    \midrule[0.35pt]
    \multicolumn{9}{c}{\textbf{OBER-Test} \citep{zhao2026precise}} \\
    \addlinespace[0.8pt]
    ObjectClear \citep{zhao2026precise} & \cellcolor{LensRankSecond}33.05 & \cellcolor{LensRankSecond}19.16 & \cellcolor{LensRankSecond}0.0132 & \cellcolor{LensRankSecond}0.0338 & \cellcolor{LensRankSecond}0.9353 & \cellcolor{LensRankSecond}0.0217 & \cellcolor{LensRankSecond}0.9894 & \cellcolor{LensRankFirst}\textbf{0.9839} \\
    OSOR \citep{zhou2026osor} & \cellcolor{LensRankThird}30.14 & 27.82 & \cellcolor{LensRankThird}0.0674 & \cellcolor{LensRankThird}0.0524 & \cellcolor{LensRankThird}0.8748 & \cellcolor{LensRankThird}0.0377 & 0.9827 & \cellcolor{LensRankThird}0.9693 \\
    OmniPaint \citep{yu2025omnipaint} & 28.79 & \cellcolor{LensRankThird}23.68 & 0.0829 & 0.0549 & 0.8575 & 0.0427 & \cellcolor{LensRankThird}0.9836 & \cellcolor{LensRankSecond}0.9803 \\
    \textbf{LensErase} & \cellcolor{LensRankFirst}\textbf{34.96} & \cellcolor{LensRankFirst}\textbf{16.97} & \cellcolor{LensRankFirst}\textbf{0.0073} & \cellcolor{LensRankFirst}\textbf{0.0256} & \cellcolor{LensRankFirst}\textbf{0.9543} & \cellcolor{LensRankFirst}\textbf{0.0176} & \cellcolor{LensRankFirst}\textbf{0.9930} & \cellcolor{LensRankFirst}\textbf{0.9839} \\
    \arrayrulecolor{LensTableDivider}
    \specialrule{0.25pt}{1.6pt}{1.9pt}
    \arrayrulecolor{LensTableRule}
    \multicolumn{9}{c}{\textbf{RORD-Val-343} \citep{sagong2022rord}} \\
    \addlinespace[0.8pt]
    ObjectClear \citep{zhao2026precise} & \cellcolor{LensRankSecond}26.21 & \cellcolor{LensRankSecond}25.42 & 0.0511 & 0.1163 & \cellcolor{LensRankSecond}0.8577 & 0.0657 & \cellcolor{LensRankSecond}0.9704 & \cellcolor{LensRankSecond}0.9566 \\
    OSOR \citep{zhou2026osor} & \cellcolor{LensRankThird}25.67 & \cellcolor{LensRankThird}26.33 & \cellcolor{LensRankThird}0.0390 & \cellcolor{LensRankSecond}0.1037 & \cellcolor{LensRankThird}0.8138 & \cellcolor{LensRankSecond}0.0542 & 0.9663 & 0.9475 \\
    OmniPaint \citep{yu2025omnipaint} & 23.64 & 27.45 & \cellcolor{LensRankSecond}0.0334 & \cellcolor{LensRankThird}0.1119 & 0.7823 & \cellcolor{LensRankThird}0.0623 & \cellcolor{LensRankThird}0.9672 & \cellcolor{LensRankThird}0.9541 \\
    \textbf{LensErase} & \cellcolor{LensRankFirst}\textbf{28.23} & \cellcolor{LensRankFirst}\textbf{20.53} & \cellcolor{LensRankFirst}\textbf{0.0272} & \cellcolor{LensRankFirst}\textbf{0.0601} & \cellcolor{LensRankFirst}\textbf{0.9053} & \cellcolor{LensRankFirst}\textbf{0.0357} & \cellcolor{LensRankFirst}\textbf{0.9715} & \cellcolor{LensRankFirst}\textbf{0.9568} \\
    \arrayrulecolor{LensTableDivider}
    \specialrule{0.25pt}{1.6pt}{1.9pt}
    \arrayrulecolor{LensTableRule}
    \multicolumn{9}{c}{\textbf{RemovalBench} \citep{wei2025omnieraser}} \\
    \addlinespace[0.8pt]
    OSOR \citep{zhou2026osor} & \cellcolor{LensRankThird}25.80 & \cellcolor{LensRankThird}43.89 & 0.0275 & \cellcolor{LensRankSecond}0.1935 & \cellcolor{LensRankThird}0.7656 & \cellcolor{LensRankSecond}0.0882 & \cellcolor{LensRankThird}0.9838 & 0.9597 \\
    OmniPaint \citep{yu2025omnipaint} & 24.57 & \cellcolor{LensRankSecond}43.25 & \cellcolor{LensRankThird}0.0265 & \cellcolor{LensRankThird}0.2103 & 0.7535 & \cellcolor{LensRankThird}0.0998 & 0.9821 & \cellcolor{LensRankThird}0.9663 \\
    ObjectClear \citep{zhao2026precise} & \cellcolor{LensRankSecond}25.99 & 46.61 & \cellcolor{LensRankSecond}0.0248 & 0.2853 & \cellcolor{LensRankSecond}0.7794 & 0.1428 & \cellcolor{LensRankSecond}0.9853 & \cellcolor{LensRankSecond}0.9718 \\
    \textbf{LensErase} & \cellcolor{LensRankFirst}\textbf{28.16} & \cellcolor{LensRankFirst}\textbf{33.67} & \cellcolor{LensRankFirst}\textbf{0.0116} & \cellcolor{LensRankFirst}\textbf{0.1125} & \cellcolor{LensRankFirst}\textbf{0.8306} & \cellcolor{LensRankFirst}\textbf{0.0586} & \cellcolor{LensRankFirst}\textbf{0.9874} & \cellcolor{LensRankFirst}\textbf{0.9737} \\
    \arrayrulecolor{LensTableDivider}
    \specialrule{0.25pt}{1.6pt}{1.9pt}
    \arrayrulecolor{LensTableRule}
    \multicolumn{9}{c}{\textbf{OmniPaint-Bench (captured subset)} \citep{yu2025omnipaint}} \\
    \addlinespace[0.8pt]
    OSOR \citep{zhou2026osor} & \cellcolor{LensRankSecond}26.02 & \cellcolor{LensRankSecond}37.52 & \cellcolor{LensRankSecond}0.0687 & \cellcolor{LensRankSecond}0.1336 & \cellcolor{LensRankSecond}0.8421 & \cellcolor{LensRankSecond}0.0566 & \cellcolor{LensRankSecond}0.9617 & \cellcolor{LensRankSecond}0.9506 \\
    OmniPaint \citep{yu2025omnipaint} & 23.42 & 40.65 & \cellcolor{LensRankThird}0.0743 & \cellcolor{LensRankThird}0.1484 & \cellcolor{LensRankThird}0.8193 & \cellcolor{LensRankThird}0.0644 & \cellcolor{LensRankThird}0.9562 & 0.9423 \\
    GeoRemover \citep{zhu2026georemover} & \cellcolor{LensRankThird}24.87 & \cellcolor{LensRankThird}38.54 & 0.1376 & 0.1575 & 0.8173 & 0.0704 & 0.9524 & \cellcolor{LensRankThird}0.9494 \\
    \textbf{LensErase} & \cellcolor{LensRankFirst}\textbf{26.22} & \cellcolor{LensRankFirst}\textbf{33.29} & \cellcolor{LensRankFirst}\textbf{0.0245} & \cellcolor{LensRankFirst}\textbf{0.1257} & \cellcolor{LensRankFirst}\textbf{0.8521} & \cellcolor{LensRankFirst}\textbf{0.0544} & \cellcolor{LensRankFirst}\textbf{0.9660} & \cellcolor{LensRankFirst}\textbf{0.9595} \\
    \arrayrulecolor{LensTableDivider}
    \specialrule{0.25pt}{1.6pt}{1.9pt}
    \arrayrulecolor{LensTableRule}
    \multicolumn{9}{c}{\textbf{AnimeEraseBench} \citep{zhou2026osor}} \\
    \addlinespace[0.8pt]
    OSOR \citep{zhou2026osor} & \cellcolor{LensRankSecond}27.96 & \cellcolor{LensRankSecond}24.45 & \cellcolor{LensRankFirst}\textbf{0.0094} & \cellcolor{LensRankSecond}0.0967 & \cellcolor{LensRankSecond}0.8870 & \cellcolor{LensRankSecond}0.0475 & \cellcolor{LensRankThird}0.9900 & 0.9751 \\
    OmniPaint \citep{yu2025omnipaint} & \cellcolor{LensRankThird}25.58 & \cellcolor{LensRankThird}27.76 & \cellcolor{LensRankThird}0.0138 & \cellcolor{LensRankThird}0.1195 & \cellcolor{LensRankThird}0.8581 & \cellcolor{LensRankThird}0.0636 & 0.9874 & \cellcolor{LensRankThird}0.9764 \\
    ObjectClear \citep{zhao2026precise} & 25.42 & 36.76 & \cellcolor{LensRankSecond}0.0121 & 0.1791 & 0.8417 & 0.0978 & \cellcolor{LensRankSecond}0.9914 & \cellcolor{LensRankFirst}\textbf{0.9821} \\
    \textbf{LensErase} & \cellcolor{LensRankFirst}\textbf{30.50} & \cellcolor{LensRankFirst}\textbf{17.20} & \cellcolor{LensRankFirst}\textbf{0.0094} & \cellcolor{LensRankFirst}\textbf{0.0527} & \cellcolor{LensRankFirst}\textbf{0.9398} & \cellcolor{LensRankFirst}\textbf{0.0159} & \cellcolor{LensRankFirst}\textbf{0.9920} & \cellcolor{LensRankSecond}0.9798 \\
    \arrayrulecolor{LensTableDivider}
    \specialrule{0.25pt}{1.6pt}{1.9pt}
    \arrayrulecolor{LensTableRule}
    \multicolumn{9}{c}{\textbf{TextEraseBench} \citep{zhou2026osor}} \\
    \addlinespace[0.8pt]
    OSOR \citep{zhou2026osor} & \cellcolor{LensRankSecond}32.04 & \cellcolor{LensRankSecond}18.52 & \cellcolor{LensRankSecond}0.0093 & \cellcolor{LensRankSecond}0.0728 & \cellcolor{LensRankSecond}0.9126 & \cellcolor{LensRankSecond}0.0412 & \cellcolor{LensRankSecond}0.9843 & \cellcolor{LensRankSecond}0.9771 \\
    OmniEraser \citep{wei2025omnieraser} & \cellcolor{LensRankThird}30.70 & \cellcolor{LensRankThird}21.97 & 0.0195 & \cellcolor{LensRankThird}0.0763 & \cellcolor{LensRankThird}0.9005 & \cellcolor{LensRankThird}0.0435 & 0.9812 & \cellcolor{LensRankThird}0.9758 \\
    RORem-mixed \citep{li2025rorem} & 29.50 & 24.83 & \cellcolor{LensRankThird}0.0178 & 0.1034 & 0.8292 & 0.0584 & \cellcolor{LensRankThird}0.9815 & 0.9744 \\
    \textbf{LensErase} & \cellcolor{LensRankFirst}\textbf{36.88} & \cellcolor{LensRankFirst}\textbf{13.51} & \cellcolor{LensRankFirst}\textbf{0.0091} & \cellcolor{LensRankFirst}\textbf{0.0339} & \cellcolor{LensRankFirst}\textbf{0.9654} & \cellcolor{LensRankFirst}\textbf{0.0135} & \cellcolor{LensRankFirst}\textbf{0.9906} & \cellcolor{LensRankFirst}\textbf{0.9822} \\
    \addlinespace[0.8pt]
    \bottomrule[0.6pt]
  \end{tabular}
  \arrayrulecolor{black}
  \endgroup
  \vspace{-11pt}
\end{table}
\FloatBarrier

\textbf{Benchmarks.}
We evaluate on seven benchmarks, including six public benchmarks: OBER-Test~\citep{zhao2026precise}, RORD-Val-343~\citep{sagong2022rord}, RemovalBench~\citep{wei2025omnieraser}, OmniPaint-Bench~\citep{yu2025omnipaint}, AnimeEraseBench, and TextEraseBench~\citep{zhou2026osor}, together with our proposed LensErase-Bench (Section~\ref{sec:lenserase_bench}). For OmniPaint-Bench, we use only its 300 captured cases, excluding the RORD-derived subset to avoid overlap with RORD-Val-343.

\textbf{Baselines.}
We compare against 12 publicly released methods: OSOR~\citep{zhou2026osor}, ObjectClear~\citep{zhao2026precise}, DORS~\citep{tang2026dors}, PANDORA~\citep{vo2026pandora}, GeoRemover~\citep{zhu2026georemover}, SmartEraser~\citep{jiang2025smarteraser}, RORem-mixed~\citep{li2025rorem}, EntityErasure~\citep{zhu2025entityerasure}, Attentive Eraser~\citep{sun2025attentive}, DesignEdit~\citep{jia2025designedit}, OmniEraser~\citep{wei2025omnieraser}, and OmniPaint~\citep{yu2025omnipaint}.
Related methods are reviewed in Appendix~\ref{sec:related_work}, and detailed baseline configurations are provided in Appendix~\ref{app:baseline_configurations}.

\textbf{Implementation Details.}
Pretraining takes approximately two weeks on 96 NVIDIA A100 GPUs. SFT and RL are each conducted on 16 GPUs, taking approximately three and five days, respectively. Additional implementation details are provided in Appendix~\ref{app:method_details}.

\textbf{Metrics.}
For the six public benchmarks with ground-truth target images, we report PSNR, SSIM, LPIPS, DISTS, CLIP-I, DINO, FID, and CMMD. For our LensErase-Bench, we adopt the evaluation metrics described in Section~\ref{sec:evaluation-dataset}.

\textbf{State-of-the-Art Performance Across Six Public Benchmarks.}
As shown in Table~\ref{tab:results-public}, LensErase achieves state-of-the-art performance across six public benchmarks. For clarity, we report only the three baselines with the lowest mean rank across the eight metrics per benchmark. Full comparisons with all 12 baselines are provided in Tables~\ref{tab:results-ober-test}--\ref{tab:results-text-erase} of the Appendix.
\begin{table}[!t]
  \centering
  \setlength{\abovecaptionskip}{2.5pt}
  \setlength{\belowcaptionskip}{1.5pt}
  \begingroup
  \definecolor{LensVLMFirst}{HTML}{F5C6C9}
  \definecolor{LensVLMSecond}{HTML}{F9DCB9}
  \definecolor{LensVLMThird}{HTML}{FBF1C3}
  \setlength{\fboxsep}{0.8pt}
  \definecolor{LensVLMTableRule}{HTML}{555B62}
  \caption{\textbf{Comparison on LensErase-Bench.}
We report the six representative baselines together with LensErase. Complete results for all 12 baselines are provided in Table~\ref{tab:vlm-full} of the Appendix.
}
    \label{tab:results-vlm}
  \fontsize{7.7}{9.1}\selectfont
  \setlength{\tabcolsep}{1.4pt}
  \renewcommand{\arraystretch}{0.98}
  \arrayrulecolor{LensVLMTableRule}
  \begin{tabular*}{\linewidth}{@{\extracolsep{\fill}}l *{6}{>{\columncolor{white}[0pt][0pt]}c}@{}}
    \toprule[0.6pt]
    \textbf{Method} & \mbox{\textbf{Removal}\,$\uparrow$} & \mbox{\textbf{Inpainting}\,$\uparrow$}
    & \mbox{\textbf{Preservation}\,$\uparrow$} & \mbox{\textbf{Cleanliness}\,$\uparrow$}
    & \mbox{\textbf{Overall}\,$\uparrow$} & \mbox{\textbf{HiFi Removal (\%)}\,$\uparrow$} \\
    \midrule[0.35pt]
    OSOR \citep{zhou2026osor} & \cellcolor{LensVLMThird}73.85 & 60.45 & 75.55 & 73.55 & 71.45 & \cellcolor{LensVLMThird}32.60 \\
    OmniPaint \citep{yu2025omnipaint} & 69.20 & \cellcolor{LensVLMSecond}66.15 & \cellcolor{LensVLMSecond}80.20 & \cellcolor{LensVLMSecond}83.20 & \cellcolor{LensVLMSecond}73.59 & \cellcolor{LensVLMSecond}42.80 \\
    ObjectClear \citep{zhao2026precise} & \cellcolor{LensVLMSecond}73.95 & \cellcolor{LensVLMThird}61.45 & 73.80 & 75.20 & \cellcolor{LensVLMThird}71.67 & 28.40 \\
    GeoRemover \citep{zhu2026georemover} & 54.30 & 49.85 & 77.05 & \cellcolor{LensVLMThird}78.00 & 62.70 & 25.60 \\
    OmniEraser \citep{wei2025omnieraser} & 55.65 & 47.95 & \cellcolor{LensVLMThird}77.55 & 57.65 & 58.89 & 22.60 \\
    RORem-mixed \citep{li2025rorem} & \cellcolor{LensVLMSecond}73.95 & 59.70 & 74.05 & 67.10 & 69.75 & 27.40 \\
    \textbf{LensErase} & \cellcolor{LensVLMFirst}\textbf{95.15} & \cellcolor{LensVLMFirst}\textbf{92.40} & \cellcolor{LensVLMFirst}\textbf{96.35} & \cellcolor{LensVLMFirst}\textbf{97.05} & \cellcolor{LensVLMFirst}\textbf{95.22} & \cellcolor{LensVLMFirst}\textbf{82.80} \\
    \addlinespace[0.8pt]
    \bottomrule[0.6pt]
  \end{tabular*}
  \arrayrulecolor{black}
  \endgroup
  \vspace{-8pt}
\end{table}
\FloatBarrier

\textbf{State-of-the-Art Performance on LensErase-Bench.}
We further compare LensErase with 12 publicly released baselines on the proposed LensErase-Bench, which contains more challenging and diverse removal cases. As shown in Table~\ref{tab:results-vlm}, LensErase achieves the best overall performance. For clarity, we report six representative baselines together with LensErase in the main table. Complete comparisons with all 12 baselines are provided in Table~\ref{tab:vlm-full} of the Appendix.

\begin{table}[!t]
  \centering
  \setlength{\abovecaptionskip}{2.5pt}
  \setlength{\belowcaptionskip}{1.5pt}
  \caption{\textbf{Stage-wise ablation of LensErase.}
We report performance after pre-training, SFT, RL, and TraceFusion. Results on LensErase-Bench are shown with all evaluation metrics, while results on public benchmarks report the average PSNR and LPIPS across the six public benchmarks.}
    \label{tab:ablation_vlm}
  \begingroup
  \fontsize{7.7}{9.1}\selectfont
  \setlength{\tabcolsep}{1.8pt}
  \renewcommand{\arraystretch}{0.98}
  \begin{tabular*}{\linewidth}{@{\extracolsep{\fill}}l*{6}{c}@{\hspace{6pt}}*{2}{c}@{}}
    \toprule
    & \multicolumn{6}{c}{\textbf{LensErase-Bench}} & \multicolumn{2}{c}{\textbf{Public Benchmarks}} \\
    \cmidrule(lr){2-7}\cmidrule(l){8-9}
    \raisebox{\dimexpr(\arraystretch\ht\strutbox+\arraystretch\dp\strutbox+\aboverulesep+\belowrulesep+\cmidrulewidth)/2\relax}[0pt][0pt]{\textbf{Model}} & \mbox{\textbf{Removal}\,$\uparrow$} & \mbox{\textbf{Inpainting}\,$\uparrow$} & \mbox{\textbf{Preservation}\,$\uparrow$} & \mbox{\textbf{Cleanliness}\,$\uparrow$} & \mbox{\textbf{Overall}\,$\uparrow$} & \mbox{\textbf{HiFi Removal}\,$\uparrow$} & \mbox{\textbf{PSNR}\,$\uparrow$} & \mbox{\textbf{LPIPS}\,$\downarrow$} \\
    \midrule
        Pre-Training & 82.35 & 84.65 & 91.05 & 92.15 & 86.51 & 65.80 & 28.03 & 0.0855 \\
    +SFT & 88.65 & 87.15 & 92.05 & 93.05 & 89.91 & 70.80 & 28.14 & 0.0826 \\
    ~+RL & 94.95 & 92.65 & 93.35 & 96.15 & 94.41 & 81.60 & 28.70 & 0.0816 \\
    ~~+TraceFusion & 95.15 & 92.40 & 96.35 & 97.05 & 95.22 & 82.80 & 30.82 & 0.0684 \\
                            \bottomrule
  \end{tabular*}
  \vspace{-8pt}
  \endgroup
\end{table}
\FloatBarrier

\begin{table}[!t]
  \centering
  \setlength{\abovecaptionskip}{2.5pt}
  \setlength{\belowcaptionskip}{1.5pt}
  \caption{\textbf{Comparison of mask injection strategies.}
Mask-Aware RoPE is compared with channel- and token-concatenation baselines, all evaluated at the pre-training stage. Both baselines obtain mask features via pixel unshuffle of $16\times16$ patches. \textit{Channel Concat} concatenates source and mask features along channels and fuses them with an MLP before condition projection. \textit{Token Concat} places source tokens alongside noisy tokens in the image branch and projects mask features to the condition branch. Both use positional indices independent of mask values.
}
    \label{tab:pretrain_ablation_vlm}
  \label{tab:structure_ablation}
  \begingroup
  \fontsize{7.7}{9.1}\selectfont
  \setlength{\tabcolsep}{1.2pt}
  \renewcommand{\arraystretch}{0.98}
  \begin{tabular*}{\linewidth}{@{\extracolsep{\fill}}l*{6}{c}@{\hspace{6pt}}*{2}{c}@{}}
    \toprule
    & \multicolumn{6}{c}{\textbf{LensErase-Bench}} & \multicolumn{2}{c}{\textbf{Public Benchmarks}} \\
    \cmidrule(lr){2-7}\cmidrule(l){8-9}
    Model & \mbox{\textbf{Removal}\,$\uparrow$} & \mbox{\textbf{Inpainting}\,$\uparrow$} & \mbox{\textbf{Preservation}\,$\uparrow$} & \mbox{\textbf{Cleanliness}\,$\uparrow$} & \mbox{\textbf{Overall}\,$\uparrow$} & \mbox{\textbf{HiFi Removal}\,$\uparrow$} & \mbox{\textbf{PSNR}\,$\uparrow$} & \mbox{\textbf{LPIPS}\,$\downarrow$} \\
    \midrule
    \textbf{Mask-Aware RoPE} & 82.35 & 84.65 & 91.05 & 92.15 & 86.51 & 65.80 & 28.03 & 0.0855 \\
    \midrule
    Channel Concat & 81.50 & 84.55 & 82.90 & 91.15 & 84.32 & 58.80 & 27.84 & 0.0869 \\
    Token Concat & 77.50 & 76.85 & 84.90 & 87.15 & 80.78 & 51.40 & 27.41 & 0.0965 \\
                            \bottomrule
  \end{tabular*}
  \vspace{-10pt}
  \endgroup
\end{table}

\textbf{Stage-wise Performance.}
Table~\ref{tab:ablation_vlm} shows that HiFi Removal increases from 65.80\% after pretraining to 70.80\% after SFT and 81.60\% after RL. TraceFusion further improves mean PSNR from 28.70 to 30.82\,dB and LPIPS from 0.0816 to 0.0684, while reaching 82.80\% HiFi Removal. Appendix~\ref{sec:reward_ablation} reports reward-component ablations.

\textbf{Effective Design for Mask Injection.} Section~\ref{sec:mask_conditioned_architecture} introduces our Mask-Aware RoPE for injecting mask information. In Table~\ref{tab:structure_ablation}, we compare it with two variants, \emph{Channel Concat} and \emph{Token Concat}. Appendix~\ref{app:pretraining_design} provides further pretraining ablations.

\textbf{Visualizations.}
Appendix Figures~\ref{fig:qualitative_commercial_01}--\ref{fig:qualitative_open_07} compare LensErase with closed-source commercial models (mask-based GPT-Image~2.5-Sunburst \citep{openai2026gptimage25} and Flux2-Erase \citep{blackforestlabs2026fluxerase}) and open-source baselines. Appendix~\ref{app:mask_variants} shows further examples using hand-drawn and SAM masks.

\clearpage

\bibliography{references}

@inproceedings{zhou2026osor,
  title={OSOR: One-Step Diffusion Inpainting for Effect-Aware Object Removal},
  author={Zhou, Qinming and Sun, Chenxi and Kong, Deyang and He, Junhao and Tang, Xiangheng and Yu, Peike and Wu, Haotian and Cao, Leilei and Zhang, Linfeng},
  booktitle={European Conference on Computer Vision},
  pages={367--384},
  year={2026},
  organization={Springer}
}

@inproceedings{zhao2026precise,
  title={Precise object and effect removal with adaptive target-aware attention},
  author={Zhao, Jixin and Wang, Zhouxia and Yang, Peiqing and Zhou, Shangchen},
  booktitle={Proceedings of the IEEE/CVF Conference on Computer Vision and Pattern Recognition},
  pages={19370--19379},
  year={2026}
}

@article{tang2026dors,
  title={DORS: Dynamic Attention Routing for Diffusion-based Object Removal in Dense Scenes},
  author={Tang, Haitong and Liu, Haipeng and Wang, Yang},
  journal={arXiv preprint arXiv:2607.16656},
  year={2026}
}

@article{vo2026pandora,
  title={PANDORA: Pixel-wise Attention Dissolution and Latent Guidance for Zero-Shot Object Removal},
  author={Vo, Dinh-Khoi and Nguyen, Van-Loc and Nguyen, Tam V and Tran, Minh-Triet and Le, Trung-Nghia},
  journal={arXiv preprint arXiv:2603.27555},
  year={2026}
}

@inproceedings{
zhu2026georemover,
title={GeoRemover: Removing Objects and Their Causal Visual Artifacts},
author={Zixin Zhu and Haoxiang Li and Xuelu Feng and He Wu and Chunming Qiao and Junsong Yuan},
booktitle={The Thirty-ninth Annual Conference on Neural Information Processing Systems},
year={2025},
url={https://openreview.net/forum?id=RnfyqrkOxD}
}

@inproceedings{jiang2025smarteraser,
  title={Smarteraser: Remove anything from images using masked-region guidance},
  author={Jiang, Longtao and Wang, Zhendong and Bao, Jianmin and Zhou, Wengang and Chen, Dongdong and Shi, Lei and Chen, Dong and Li, Houqiang},
  booktitle={2025 IEEE/CVF Conference on Computer Vision and Pattern Recognition (CVPR)},
  pages={24452--24462},
  year={2025},
  organization={IEEE}
}

@inproceedings{li2025rorem,
  title={Rorem: Training a robust object remover with human-in-the-loop},
  author={Li, Ruibin and Yang, Tao and Guo, Song and Zhang, Lei},
  booktitle={2025 IEEE/CVF Conference on Computer Vision and Pattern Recognition (CVPR)},
  pages={14024--14035},
  year={2025},
  organization={IEEE}
}

@inproceedings{zhu2025entityerasure,
  title={Entityerasure: Erasing entity cleanly via amodal entity segmentation and completion},
  author={Zhu, Yixing and Zhang, Qing and Wang, Yitong and Nie, Yongwei and Zheng, Wei-Shi},
  booktitle={2025 IEEE/CVF Conference on Computer Vision and Pattern Recognition (CVPR)},
  pages={28274--28283},
  year={2025},
  organization={IEEE}
}

@inproceedings{sun2025attentive,
  title={Attentive eraser: Unleashing diffusion model’s object removal potential via self-attention redirection guidance},
  author={Sun, Wenhao and Dong, Xue-Mei and Cui, Benlei and Tang, Jingqun},
  booktitle={Proceedings of the AAAI Conference on Artificial Intelligence},
  volume={39},
  number={19},
  pages={20734--20742},
  year={2025}
}

@inproceedings{jia2025designedit,
  title={DesignEdit: unify spatial-aware image editing via training-free inpainting with a multi-layered latent diffusion framework},
  author={Jia, Yueru and Cheng, Aosong and Yuan, Yuhui and Wang, Chuke and Li, Ji and Jia, Huizhu and Zhang, Shanghang},
  booktitle={Proceedings of the AAAI Conference on Artificial Intelligence},
  volume={39},
  number={4},
  pages={3958--3966},
  year={2025}
}

@article{wei2025omnieraser,
  title={Omnieraser: Remove objects and their effects in images with paired video-frame data},
  author={Wei, Runpu and Yin, Zijin and Zhang, Shuo and Zhou, Lanxiang and Wang, Xueyi and Ban, Chao and Cao, Tianwei and Sun, Hao and He, Zhongjiang and Liang, Kongming and others},
  journal={arXiv preprint arXiv:2501.07397},
  year={2025}
}

@inproceedings{yu2025omnipaint,
  title={Omnipaint: Mastering object-oriented editing via disentangled insertion-removal inpainting},
  author={Yu, Yongsheng and Zeng, Ziyun and Zheng, Haitian and Luo, Jiebo},
  booktitle={2025 IEEE/CVF International Conference on Computer Vision (ICCV)},
  pages={17324--17334},
  year={2025},
  organization={IEEE}
}

@article{ekin2024clipaway,
  title={Clipaway: Harmonizing focused embeddings for removing objects via diffusion models},
  author={Ekin, Yi{\u{g}}it and Yildirim, Ahmet B and Caglar, Erdem E and Erdem, Aykut and Erdem, Erkut and Dundar, Aysegul},
  journal={Advances in Neural Information Processing Systems},
  volume={37},
  pages={17572--17601},
  year={2024}
}

@inproceedings{zhuang2024task,
  title={A task is worth one word: Learning with task prompts for high-quality versatile image inpainting},
  author={Zhuang, Junhao and Zeng, Yanhong and Liu, Wenran and Yuan, Chun and Chen, Kai},
  booktitle={European Conference on Computer Vision},
  pages={195--211},
  year={2024},
  organization={Springer}
}

@inproceedings{chen2024freecompose,
  title={Freecompose: Generic zero-shot image composition with diffusion prior},
  author={Chen, Zhekai and Wang, Wen and Yang, Zhen and Yuan, Zeqing and Chen, Hao and Shen, Chunhua},
  booktitle={European Conference on Computer Vision},
  pages={70--87},
  year={2024},
  organization={Springer}
}

@inproceedings{ju2024brushnet,
  title={Brushnet: A plug-and-play image inpainting model with decomposed dual-branch diffusion},
  author={Ju, Xuan and Liu, Xian and Wang, Xintao and Bian, Yuxuan and Shan, Ying and Xu, Qiang},
  booktitle={European Conference on Computer Vision},
  pages={150--168},
  year={2024},
  organization={Springer}
}

@inproceedings{winter2024objectdrop,
  title={Objectdrop: Bootstrapping counterfactuals for photorealistic object removal and insertion},
  author={Winter, Daniel and Cohen, Matan and Fruchter, Shlomi and Pritch, Yael and Rav-Acha, Alex and Hoshen, Yedid},
  booktitle={European Conference on Computer Vision},
  pages={112--129},
  year={2024},
  organization={Springer}
}

@inproceedings{li2024magiceraser,
  title={Magiceraser: Erasing any objects via semantics-aware control},
  author={Li, Fan and Zhang, Zixiao and Huang, Yi and Liu, Jianzhuang and Pei, Renjing and Shao, Bin and Xu, Songcen},
  booktitle={European Conference on Computer Vision},
  pages={215--231},
  year={2024},
  organization={Springer}
}

@inproceedings{chen2024anydoor,
  title={Anydoor: Zero-shot object-level image customization},
  author={Chen, Xi and Huang, Lianghua and Liu, Yu and Shen, Yujun and Zhao, Deli and Zhao, Hengshuang},
  booktitle={2024 IEEE/CVF Conference on Computer Vision and Pattern Recognition (CVPR)},
  pages={6593--6602},
  year={2024},
  organization={IEEE}
}

@article{yildirim2023inst,
  title={Inst-inpaint: Instructing to remove objects with diffusion models},
  author={Yildirim, Ahmet Burak and Baday, Vedat and Erdem, Erkut and Erdem, Aykut and Dundar, Aysegul},
  journal={arXiv preprint arXiv:2304.03246},
  year={2023}
}

@article{yang2023magicremover,
  title={Magicremover: Tuning-free text-guided image inpainting with diffusion models},
  author={Yang, Siyuan and Zhang, Lu and Ma, Liqian and Liu, Yu and Fu, JingJing and He, You},
  journal={arXiv preprint arXiv:2310.02848},
  year={2023}
}

@article{avrahami2023blended,
  title={Blended latent diffusion},
  author={Avrahami, Omri and Fried, Ohad and Lischinski, Dani},
  journal={ACM transactions on graphics (TOG)},
  volume={42},
  number={4},
  pages={1--11},
  year={2023},
  publisher={ACM New York, NY, USA}
}

@inproceedings{lugmayr2022repaint,
  title={Repaint: Inpainting using denoising diffusion probabilistic models},
  author={Lugmayr, Andreas and Danelljan, Martin and Romero, Andres and Yu, Fisher and Timofte, Radu and Van Gool, Luc},
  booktitle={2022 IEEE/CVF conference on computer vision and pattern recognition (CVPR)},
  pages={11451--11461},
  year={2022},
  organization={IEEE}
}

@inproceedings{avrahami2022blended,
  title={Blended diffusion for text-driven editing of natural images},
  author={Avrahami, Omri and Lischinski, Dani and Fried, Ohad},
  booktitle={2022 IEEE/CVF Conference on Computer Vision and Pattern Recognition (CVPR)},
  pages={18187--18197},
  year={2022},
  organization={IEEE}
}

@inproceedings{saharia2022palette,
  title={Palette: Image-to-image diffusion models},
  author={Saharia, Chitwan and Chan, William and Chang, Huiwen and Lee, Chris and Ho, Jonathan and Salimans, Tim and Fleet, David and Norouzi, Mohammad},
  booktitle={ACM SIGGRAPH 2022 conference proceedings},
  pages={1--10},
  year={2022}
}

@inproceedings{bertalmio2000image,
  title={Image inpainting},
  author={Bertalmio, Marcelo and Sapiro, Guillermo and Caselles, Vincent and Ballester, Coloma},
  booktitle={Proceedings of the 27th annual conference on Computer graphics and interactive techniques},
  pages={417--424},
  year={2000}
}

@article{criminisi2004region,
  title={Region filling and object removal by exemplar-based image inpainting},
  author={Criminisi, Antonio and P{\'e}rez, Patrick and Toyama, Kentaro},
  journal={IEEE Transactions on image processing},
  volume={13},
  number={9},
  pages={1200--1212},
  year={2004},
  publisher={IEEE}
}

@article{barnes2009patchmatch,
  title={PatchMatch: A randomized correspondence algorithm for structural image editing},
  author={Barnes, Connelly and Shechtman, Eli and Finkelstein, Adam and Goldman, Dan B},
  journal={ACM Trans. Graph.},
  volume={28},
  number={3},
  pages={24},
  year={2009}
}

@inproceedings{pathak2016context,
  title={Context encoders: Feature learning by inpainting},
  author={Pathak, Deepak and Krahenbuhl, Philipp and Donahue, Jeff and Darrell, Trevor and Efros, Alexei A},
  booktitle={Proceedings of the IEEE conference on computer vision and pattern recognition},
  pages={2536--2544},
  year={2016}
}

@article{iizuka2017globally,
  title={Globally and locally consistent image completion},
  author={Iizuka, Satoshi and Simo-Serra, Edgar and Ishikawa, Hiroshi},
  journal={ACM Transactions on Graphics (ToG)},
  volume={36},
  number={4},
  pages={1--14},
  year={2017},
  publisher={ACM New York, NY, USA}
}

@inproceedings{liu2018image,
  title={Image inpainting for irregular holes using partial convolutions},
  author={Liu, Guilin and Reda, Fitsum A and Shih, Kevin J and Wang, Ting-Chun and Tao, Andrew and Catanzaro, Bryan},
  booktitle={European conference on computer vision},
  pages={89--105},
  year={2018},
  organization={Springer}
}

@inproceedings{yu2018generative,
  title={Generative image inpainting with contextual attention},
  author={Yu, Jiahui and Lin, Zhe and Yang, Jimei and Shen, Xiaohui and Lu, Xin and Huang, Thomas S},
  booktitle={2018 IEEE/CVF conference on computer vision and pattern recognition},
  pages={5505--5514},
  year={2018},
  organization={IEEE}
}

@inproceedings{yu2019free,
  title={Free-form image inpainting with gated convolution},
  author={Yu, Jiahui and Lin, Zhe and Yang, Jimei and Shen, Xiaohui and Lu, Xin and Huang, Thomas S},
  booktitle={Proceedings of the IEEE/CVF international conference on computer vision},
  pages={4471--4480},
  year={2019}
}

@inproceedings{nazeri2019edgeconnect,
  title={Edgeconnect: Structure guided image inpainting using edge prediction},
  author={Nazeri, Kamyar and Ng, Eric and Joseph, Tony and Qureshi, Faisal and Ebrahimi, Mehran},
  booktitle={2019 IEEE/CVF International Conference on Computer Vision Workshop (ICCVW)},
  pages={3265--3274},
  year={2019},
  organization={IEEE}
}

@inproceedings{suvorov2022resolution,
  title={Resolution-robust large mask inpainting with fourier convolutions},
  author={Suvorov, Roman and Logacheva, Elizaveta and Mashikhin, Anton and Remizova, Anastasia and Ashukha, Arsenii and Silvestrov, Aleksei and Kong, Naejin and Goka, Harshith and Park, Kiwoong and Lempitsky, Victor},
  booktitle={2022 IEEE/CVF Winter Conference on Applications of Computer Vision (WACV)},
  pages={3172--3182},
  year={2022},
  organization={IEEE}
}

@inproceedings{li2022mat,
  title={Mat: Mask-aware transformer for large hole image inpainting},
  author={Li, Wenbo and Lin, Zhe and Zhou, Kun and Qi, Lu and Wang, Yi and Jia, Jiaya},
  booktitle={2022 IEEE/CVF Conference on Computer Vision and Pattern Recognition (CVPR)},
  pages={10748--10758},
  year={2022},
  organization={IEEE}
}

@inproceedings{dong2022incremental,
  title={Incremental transformer structure enhanced image inpainting with masking positional encoding},
  author={Dong, Qiaole and Cao, Chenjie and Fu, Yanwei},
  booktitle={2022 IEEE/CVF Conference on Computer Vision and Pattern Recognition (CVPR)},
  pages={11348--11358},
  year={2022},
  organization={IEEE}
}

@article{cao2023zits++,
  title={Zits++: Image inpainting by improving the incremental transformer on structural priors},
  author={Cao, Chenjie and Dong, Qiaole and Fu, Yanwei},
  journal={IEEE transactions on pattern analysis and machine intelligence},
  volume={45},
  number={10},
  pages={12667--12684},
  year={2023},
  publisher={IEEE}
}

@inproceedings{liu2025erase,
  title={Erase diffusion: Empowering object removal through calibrating diffusion pathways},
  author={Liu, Yi and Zhou, Hao and Cui, Benlei and Shang, Wenxiang and Lin, Ran},
  booktitle={2025 IEEE/CVF Conference on Computer Vision and Pattern Recognition (CVPR)},
  pages={2418--2427},
  year={2025},
  organization={IEEE}
}

@article{jo2025eraselora,
  title={EraseLoRA: MLLM-Driven Foreground Exclusion and Background Subtype Aggregation for Dataset-Free Object Removal},
  author={Jo, Sanghyun and Lee, Donghwan and Jung, Eunji and Oh, Seong Je and Kim, Kyungsu},
  journal={arXiv preprint arXiv:2512.21545},
  year={2025}
}

@inproceedings{liu2026raa,
  title={RAA: Achieving Interactive Remove/Add Anything via Fully Synthetic Data},
  author={Liu, Delong and Hou, Haotian and Hou, Zhaohui and Han, Shihao and Huang, Zhiyuan and Zhan, Mingjie and Su, Fei and Zhao, Zhicheng},
  booktitle={Proceedings of the AAAI Conference on Artificial Intelligence},
  volume={40},
  number={9},
  pages={7123--7131},
  year={2026}
}

@inproceedings{chadebec2025lbm,
  title={Lbm: Latent bridge matching for fast image-to-image translation},
  author={Chadebec, Cl{\'e}ment and Tasar, Onur and Sreetharan, Sanjeev and Aubin, Benjamin},
  booktitle={2025 IEEE/CVF International Conference on Computer Vision (ICCV)},
  pages={29086--29098},
  year={2025},
  organization={IEEE}
}

@article{xiu2026prederase,
  title={PredErase: Training-Free Object-and-Effect Removal with Predictive Latent Guidance},
  author={Xiu, Waikit and Lu, Qiang and Chen, Junbiao and Li, Xiying},
  journal={arXiv preprint arXiv:2609.00956},
  year={2026}
}

@article{vo2025cpam,
  title={CPAM: Context-Preserving Adaptive Manipulation for Zero-Shot Real Image Editing},
  author={Vo, Dinh-Khoi and Do, Thanh-Toan and Nguyen, Tam V and Tran, Minh-Triet and Le, Trung-Nghia},
  journal={arXiv preprint arXiv:2506.18438},
  year={2025}
}

@article{zhang2026clickremoval,
  title={ClickRemoval: An Interactive Open-Source Tool for Object Removal in Diffusion Models},
  author={Zhang, Ledun and Ji, Yatu and Zhuang, Xufei and Yao, Xinying},
  journal={arXiv preprint arXiv:2605.14461},
  year={2026}
}

@inproceedings{sagong2022rord,
  title={RORD: A Real-world Object Removal Dataset},
  author={Sagong, Min-Cheol and Yeo, Yoon-Jae and Jung, Seung-Won and Ko, Sung-Jea},
  booktitle={Proceedings of the British Machine Vision Conference (BMVC)},
  year={2022}
}

@inproceedings{kuprashevich2026nohumansrequired,
  title={Nohumansrequired: Autonomous high-quality image editing triplet mining},
  author={Kuprashevich, Maksim and Alekseenko, Grigorii and Tolstykh, Irina and Fedorov, Georgii and Suleimanov, Bulat and Dokholyan, Vladimir and Gordeev, Aleksandr},
  booktitle={2026 IEEE/CVF Winter Conference on Applications of Computer Vision (WACV)},
  pages={6059--6068},
  year={2026},
  organization={IEEE}
}

@article{ho2020denoising,
  title={Denoising diffusion probabilistic models},
  author={Ho, Jonathan and Jain, Ajay and Abbeel, Pieter},
  journal={Advances in neural information processing systems},
  volume={33},
  pages={6840--6851},
  year={2020}
}

@article{song2020denoising,
  title={Denoising diffusion implicit models},
  author={Song, Jiaming and Meng, Chenlin and Ermon, Stefano},
  journal={arXiv preprint arXiv:2010.02502},
  year={2020}
}

@inproceedings{rombach2022high,
  title={High-resolution image synthesis with latent diffusion models},
  author={Rombach, Robin and Blattmann, Andreas and Lorenz, Dominik and Esser, Patrick and Ommer, Bj{\"o}rn},
  booktitle={2022 IEEE/CVF conference on computer vision and pattern recognition (CVPR)},
  pages={10674--10685},
  year={2022},
  organization={ieee}
}

@inproceedings{podell2024sdxl,
  title={Sdxl: Improving latent diffusion models for high-resolution image synthesis},
  author={Podell, Dustin and English, Zion and Lacey, Kyle and Blattmann, Andreas and Dockhorn, Tim and M{\"u}ller, Jonas and Penna, Joe and Rombach, Robin},
  booktitle={International Conference on Learning Representations},
  volume={2024},
  pages={1862--1874},
  year={2024}
}

@incollection{perez2023poisson,
  title={Poisson image editing},
  author={P{\'e}rez, Patrick and Gangnet, Michel and Blake, Andrew},
  booktitle={Seminal Graphics Papers: Pushing the Boundaries, Volume 2},
  pages={577--582},
  year={2023}
}

@inproceedings{black2024training,
  title={Training diffusion models with reinforcement learning},
  author={Black, Kevin and Janner, Michael and Du, Yilun and Kostrikov, Ilya and Levine, Sergey},
  booktitle={International Conference on Learning Representations},
  volume={2024},
  pages={4965--4987},
  year={2024}
}

@article{fan2023dpok,
  title={Dpok: Reinforcement learning for fine-tuning text-to-image diffusion models},
  author={Fan, Ying and Watkins, Olivia and Du, Yuqing and Liu, Hao and Ryu, Moonkyung and Boutilier, Craig and Abbeel, Pieter and Ghavamzadeh, Mohammad and Lee, Kangwook and Lee, Kimin},
  journal={Advances in neural information processing systems},
  volume={36},
  pages={79858--79885},
  year={2023}
}

@article{prabhudesai2023aligning,
  title={Aligning text-to-image diffusion models with reward backpropagation},
  author={Prabhudesai, Mihir and Goyal, Anirudh and Pathak, Deepak and Fragkiadaki, Katerina},
  year={2023}
}

@inproceedings{clark2024directly,
  title={Directly fine-tuning diffusion models on differentiable rewards},
  author={Clark, Kevin and Vicol, Paul and Swersky, Kevin and Fleet, David},
  booktitle={International Conference on Learning Representations},
  volume={2024},
  pages={4793--4822},
  year={2024}
}

@inproceedings{
liu2026flow,
title={Flow-{GRPO}: Training Flow Matching Models via Online {RL}},
author={Jie Liu and Gongye Liu and Jiajun Liang and Yangguang Li and Jiaheng Liu and Xintao Wang and Pengfei Wan and Di ZHANG and Wanli Ouyang},
booktitle={The Thirty-ninth Annual Conference on Neural Information Processing Systems},
year={2025},
url={https://openreview.net/forum?id=oCBKGw5HNf}
}

@inproceedings{zheng2026diffusionnft,
  title={Diffusionnft: Online diffusion reinforcement with forward process},
  author={Zheng, Kaiwen and Chen, Huayu and Ye, Haotian and Wang, Haoxiang and Zhang, Qinsheng and Jiang, Kai and Su, Hang and Ermon, Stefano and Zhu, Jun and Liu, Ming-Yu},
  booktitle={International Conference on Learning Representations},
  volume={2026},
  pages={134129--134150},
  year={2026}
}

@article{zhang2023magicbrush,
  title={Magicbrush: A manually annotated dataset for instruction-guided image editing},
  author={Zhang, Kai and Mo, Lingbo and Chen, Wenhu and Sun, Huan and Su, Yu},
  journal={Advances in neural information processing systems},
  volume={36},
  pages={31428--31449},
  year={2023}
}

@article{wang2004image,
  title={Image quality assessment: from error visibility to structural similarity},
  author={Wang, Zhou and Bovik, Alan C and Sheikh, Hamid R and Simoncelli, Eero P},
  journal={IEEE transactions on image processing},
  volume={13},
  number={4},
  pages={600--612},
  year={2004},
  publisher={IEEE}
}

@article{chen2026lens,
  title={Lens: Rethinking Training Efficiency for Foundational Text-to-Image Models},
  author={Chen, Dong and Wei, Fangyun and Wan, Ziyu and Chen, Dongdong and Zhang, Jiawei and Zhao, Jinjing and Zhang, Sirui and Yue, Yang and Liang, Zhiyang and Guo, Baining and others},
  journal={arXiv preprint arXiv:2605.21573},
  year={2026}
}

@article{hu2021lora,
  title={Lora: Low-rank adaptation of large language models},
  author={Hu, Edward J and Shen, Yelong and Wallis, Phillip and Allen-Zhu, Zeyuan and Li, Yuanzhi and Wang, Shean and Wang, Lu and Chen, Weizhu},
  journal={arXiv preprint arXiv:2106.09685},
  year={2021}
}

@article{ravi2024sam2,
  title={SAM 2: Segment Anything in Images and Videos},
  author={Ravi, Nikhila and Gabeur, Valentin and Hu, Yuan-Ting and Hu, Ronghang and Ryali, Chaitanya and Ma, Tengyu and Khedr, Haitham and R{\"a}dle, Roman and Rolland, Chloe and Gustafson, Laura and Mintun, Eric and Pan, Junting and Alwala, Kalyan Vasudev and Carion, Nicolas and Wu, Chao-Yuan and Girshick, Ross and Doll{\'a}r, Piotr and Feichtenhofer, Christoph},
  journal={arXiv preprint arXiv:2408.00714},
  url={https://arxiv.org/abs/2408.00714},
  year={2024}
}

@article{zheng2024birefnet,
  title={Bilateral Reference for High-Resolution Dichotomous Image Segmentation},
  author={Zheng, Peng and Gao, Dehong and Fan, Deng-Ping and Liu, Li and Laaksonen, Jorma and Ouyang, Wanli and Sebe, Nicu},
  journal={CAAI Artificial Intelligence Research},
  volume = {3},
  pages = {9150038},
  year={2024}
}

@inproceedings{jayasumana2024rethinking,
  title={Rethinking fid: Towards a better evaluation metric for image generation},
  author={Jayasumana, Sadeep and Ramalingam, Srikumar and Veit, Andreas and Glasner, Daniel and Chakrabarti, Ayan and Kumar, Sanjiv},
  booktitle={2024 IEEE/CVF Conference on Computer Vision and Pattern Recognition (CVPR)},
  pages={9307--9315},
  year={2024},
  organization={IEEE}
}

@article{su2024roformer,
  title={Roformer: Enhanced transformer with rotary position embedding},
  author={Su, Jianlin and Ahmed, Murtadha and Lu, Yu and Pan, Shengfeng and Bo, Wen and Liu, Yunfeng},
  journal={Neurocomputing},
  volume={568},
  pages={127063},
  year={2024},
  publisher={Elsevier}
}

@article{lipman2022flow,
  title={Flow matching for generative modeling},
  author={Lipman, Yaron and Chen, Ricky TQ and Ben-Hamu, Heli and Nickel, Maximilian and Le, Matt},
  journal={arXiv preprint arXiv:2210.02747},
  year={2022}
}

@article{simeoni2025dinov3,
  title={Dinov3},
  author={Sim{\'e}oni, Oriane and Vo, Huy V and Seitzer, Maximilian and Baldassarre, Federico and Oquab, Maxime and Jose, Cijo and Khalidov, Vasil and Szafraniec, Marc and Yi, Seungeun and Ramamonjisoa, Micha{\"e}l and others},
  journal={arXiv preprint arXiv:2508.10104},
  year={2025}
}

@inproceedings{wang2025yoloe,
  title={YOLOE: Real-Time Seeing Anything},
  author={Wang, Ao and Liu, Lihao and Chen, Hui and Lin, Zijia and Han, Jungong and Ding, Guiguang},
  booktitle={2025 IEEE/CVF International Conference on Computer Vision (ICCV)},
  pages={24591--24602},
  year={2025},
  organization={IEEE}
}

@misc{cui2025paddleocr30technicalreport,
  title={PaddleOCR 3.0 Technical Report},
  author={Cheng Cui and Ting Sun and Manhui Lin and Tingquan Gao and Yubo Zhang and Jiaxuan Liu and Xueqing Wang and Zelun Zhang and Changda Zhou and Hongen Liu and Yue Zhang and Wenyu Lv and Kui Huang and Yichao Zhang and Jing Zhang and Jun Zhang and Yi Liu and Dianhai Yu and Yanjun Ma},
  year={2025},
  eprint={2507.05595},
  archivePrefix={arXiv},
  primaryClass={cs.CV},
  url={https://arxiv.org/abs/2507.05595}
}

@misc{openai2025gpt41,
  title        = {{Introducing GPT-4.1 in the API}},
  author       = {{OpenAI}},
  year         = {2025},
  howpublished = {\url{https://openai.com/index/gpt-4-1/}}
}

@misc{openai2026gpt56sol,
  title        = {{GPT-5.6 Sol}},
  author       = {{OpenAI}},
  year         = {2026},
  howpublished = {\url{https://developers.openai.com/api/docs/models/gpt-5.6-sol}},
  note         = {Accessed: 2026-09-01}
}

@inproceedings{yamaguchi2021canvasvae,
  title={Canvasvae: Learning to generate vector graphic documents},
  author={Yamaguchi, Kota},
  booktitle={2021 IEEE/CVF International Conference on Computer Vision (ICCV)},
  pages={5461--5469},
  year={2021},
  organization={IEEE}
}

@inproceedings{qi2023highquality,
  title     = {High-Quality Entity Segmentation},
  author    = {Qi, Lu and Kuen, Jason and Shen, Tiancheng and Gu, Jiuxiang and Guo, Weidong and Jia, Jiaya and Lin, Zhe and Yang, Ming-Hsuan},
  booktitle = {Proceedings of the IEEE/CVF International Conference on Computer Vision},
  year      = {2023}
}

@inproceedings{deng2023recurrent,
  title     = {Recurrent Multi-scale Transformer for High-Resolution Salient Object Detection},
  author    = {Deng, Xinhao and Zhang, Pingping and Liu, Wei and Lu, Huchuan},
  booktitle = {Proceedings of the 31st ACM International Conference on Multimedia},
  pages     = {7413--7423},
  year      = {2023},
  doi       = {10.1145/3581783.3611983}
}

@inproceedings{xie2022pyramid,
  title     = {Pyramid Grafting Network for One-Stage High Resolution Saliency Detection},
  author    = {Xie, Chenxi and Xia, Changqun and Ma, Mingcan and Zhao, Zhirui and Chen, Xiaowu and Li, Jia},
  booktitle = {Proceedings of the IEEE/CVF Conference on Computer Vision and Pattern Recognition},
  year      = {2022}
}

@misc{openai2026gptimage25,
  author       = {{OpenAI}},
  title        = {{GPT-Image-2.5 Sunburst}},
  year         = {2026},
  howpublished = {\url{https://developers.openai.com/api/docs/models/gpt-image-2.5-sunburst}},
  note         = {Accessed: 2026-09-21}
}

@article{lan2026freestyle,
  title={FreeStyle: Free Control of Style-Content Dual-Reference Generation from Community LoRA Mining},
  author={Lan, Jinghong and Cheng, Wei and Chen, Yunuo and Ye, Ziqi and Xing, Peng and Fang, Yixiao and Wang, Rui and Yang, Yufeng and Zhang, Xuanyang and Zeng, Xianfang and others},
  journal={arXiv preprint arXiv:2606.20506},
  year={2026}
}

@article{fu2025imontage,
  title={imontage: Unified, versatile, highly dynamic many-to-many image generation},
  author={Fu, Zhoujie and Zeng, Xianfang and Lan, Jinghong and Liao, Xinyao and Chen, Cheng and Chen, Junyi and Wei, Jiacheng and Cheng, Wei and Liu, Shiyu and Chen, Yunuo and others},
  journal={arXiv preprint arXiv:2511.20635},
  year={2025}
}

@misc{blackforestlabs2026fluxerase,
  author       = {{Black Forest Labs}},
  title        = {{FLUX Erase}},
  year         = {2026},
  howpublished = {\url{https://docs.bfl.ai/flux_tools/flux_erase}},
  note         = {Accessed: 2026-09-21}
}

@article{zhao2026vibeedit,
  title={VibeEdit: Image Editing with Canvas Instructions},
  author={Zhao, Jinjing and Wei, Fangyun and Wang, Yitong and Wu, Xiuyu and Chen, Yunuo and Yue, Yang and Zhang, Sirui and Wang, Wenbo and Zhang, Hongyang and Chen, Dong and others},
  journal={arXiv preprint arXiv:2610.12229},
  year={2026}
}
\bibliographystyle{unsrtnat}

\clearpage
\appendix

\section{Related Work}
\label{sec:related_work}

\subsection{Image Inpainting and Object Removal}

Classical image inpainting propagates structure from hole boundaries or retrieves compatible patches from visible regions~\citep{bertalmio2000image,criminisi2004region,barnes2009patchmatch}. Learning-based methods introduce semantic completion, global--local adversarial supervision, partial or gated convolutions, contextual feature transfer, and explicit structural priors~\citep{pathak2016context,iizuka2017globally,liu2018image,yu2018generative,yu2019free,nazeri2019edgeconnect}. Fourier-based models and mask- or structure-aware Transformers further expand the effective context for completing large and irregular missing regions~\citep{suvorov2022resolution,li2022mat,dong2022incremental,cao2023zits++}.

Diffusion models provide strong generative priors for reconstructing missing content~\citep{ho2020denoising,song2020denoising,rombach2022high,podell2024sdxl}. Inpainting methods adapt these priors through constrained resampling, spatial blending, conditional image translation, task prompting, or dedicated conditioning branches~\citep{lugmayr2022repaint,avrahami2022blended,avrahami2023blended,saharia2022palette,zhuang2024task,ju2024brushnet}. Object-removal methods further introduce language or semantic control~\citep{yildirim2023inst,ekin2024clipaway,li2024magiceraser}, masked-region guidance~\citep{jiang2025smarteraser}, amodal completion~\citep{zhu2025entityerasure}, and calibrated diffusion pathways~\citep{liu2025erase}. A complementary line avoids task-specific training by redirecting attention or modifying latent trajectories~\citep{yang2023magicremover,sun2025attentive,vo2026pandora,tang2026dors}. ClickRemoval provides interactive click-based removal through attention manipulation~\citep{zhang2026clickremoval}, while EraseLoRA combines multimodal scene interpretation with test-time optimization for dataset-free removal~\citep{jo2025eraselora}.

\subsection{Data and Models for Image Editing}

Editing supervision ranges from manually annotated source--instruction--target triplets in MagicBrush~\citep{zhang2023magicbrush} to fully synthetic pairs in RAA~\citep{liu2026raa} and automated triplet mining in NoHumansRequired~\citep{kuprashevich2026nohumansrequired}. RORem introduces human-in-the-loop refinement to improve removal supervision~\citep{li2025rorem}. Broader image-editing systems study image composition, object customization, and spatial control~\citep{chen2024freecompose,chen2024anydoor,jia2025designedit}. VibeEdit specifies edits through on-image canvas instructions~\citep{zhao2026vibeedit}. Reference-guided generation includes style--content control in FreeStyle~\citep{lan2026freestyle} and many-to-many image generation in iMontage~\citep{fu2025imontage}. CPAM emphasizes context preservation during real-image editing~\citep{vo2025cpam}, while LBM formulates image translation through latent bridge matching~\citep{chadebec2025lbm}.

LensErase targets a unified source--mask interface spanning natural objects, visual text, translucent watermarks, and layered designs. Its coverage is supported by large-scale, domain-specific data construction, while visual initialization and mask-aware conditioning adapt a pretrained text-to-image prior to erasing. Supervised fine-tuning and reward-based post-training further refine removal quality and content preservation.

\subsection{Object Effects and Background Preservation}

Object erasing often requires removing associated effects, such as shadows and reflections, that may extend beyond the selection mask. ObjectDrop explicitly models such effects through counterfactual supervision~\citep{winter2024objectdrop}. OmniEraser mines paired video frames~\citep{wei2025omnieraser}, OmniPaint couples removal with insertion~\citep{yu2025omnipaint}, and GeoRemover separates geometry removal from appearance rendering~\citep{zhu2026georemover}. ObjectClear reuses target-aware attention to localize associated effects and guide inference-time fusion~\citep{zhao2026precise}, whereas OSOR combines alpha-region prediction with single-step, mask-robust generation~\citep{zhou2026osor}. PredErase provides a training-free alternative based on predictive latent guidance~\citep{xiu2026prederase}.

Spatial diffusion blending and gradient-domain editing provide established mechanisms for combining edited content with preserved surroundings~\citep{avrahami2022blended,perez2023poisson}. LensErase instead retains the intact source image as a visual condition, preserving information about the selected target, its associated effects, and content visible through translucent overlays. At inference, TraceFusion estimates the effective edit support from the selection mask and source--output residuals, then combines generated content with the original image outside this support.

\subsection{Reward-Based Post-Training}

Reward optimization adapts generative models to criteria that paired reconstruction objectives may not fully capture. DDPO and DPOK formulate denoising as a sequential decision process and optimize terminal rewards with policy gradients, with DPOK additionally regularizing updates toward a pretrained reference~\citep{black2024training,fan2023dpok}. For differentiable rewards, AlignProp and DRaFT backpropagate through the sampling trajectory~\citep{prabhudesai2023aligning,clark2024directly}. Flow-GRPO extends online policy optimization to flow-matching models through stochastic sampling~\citep{liu2026flow}. DiffusionNFT~\citep{zheng2026diffusionnft} instead learns from scored final rollouts using a forward-process flow-matching objective, with relative rewards defining positive and negative implicit objectives without backpropagating through the sampling trajectory.

LensErase adopts DiffusionNFT with removal-specific rewards that combine semantic assessment of the intended edit with deterministic fidelity measures. The reward separately evaluates removal, region reconstruction, semantic preservation, visual cleanliness, and fidelity outside the edit region, encouraging successful erasure while protecting unrelated content.

\FloatBarrier
\section{Data Construction and Filtering Details}
\label{app:data-construction-filtering}

All supervised data pipelines produce the aligned triplet $(I_{\text{input}}, I_{\text{output}}, M)$. Language is used for data construction and screening, but is never provided as an input to the student model.

\begin{figure*}[t]
    \centering
    \includegraphics[width=\textwidth]{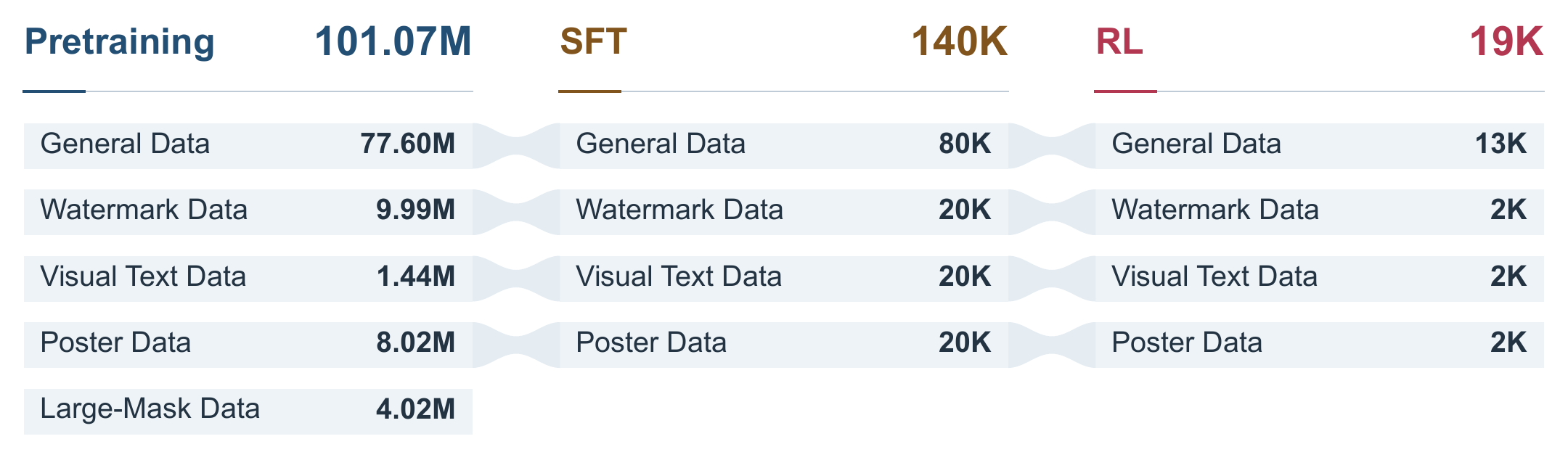}
    \caption{\textbf{Data composition across training stages.} Dataset sizes and category counts for pretraining, SFT, and RL. Bands connect shared categories across stages.}
    \label{fig:data-overview}
\end{figure*}

Figure~\ref{fig:data-overview} summarizes the composition of the 101.07M pretraining examples, 140K SFT examples, and 19K RL tasks. Pretraining and SFT use aligned input--target--mask triplets, whereas RL uses curated image--mask tasks with domain-specific references for reward computation. The following sections detail the construction and filtering procedures for pretraining, SFT, and RL data.

\subsection{Pretraining Data Pipelines}
\label{app:specialized-data-pipelines}

\subsubsection{General Data: addition-to-removal counterfactuals}
\label{sec:addition-to-removal}

Directly asking a generative teacher to remove an object makes the clean target a hallucinated inpainting. We instead begin with a real image $I_{\text{output}}$ and ask an editing teacher $G_{\mathrm{add}}$ to add an object that was absent from the original scene:
\begin{equation}
    I_{\text{input}}=G_{\mathrm{add}}(I_{\text{output}},P),
    \qquad
    M=\operatorname{RecoverMask}(I_{\text{input}},I_{\text{output}}).
    \label{eq:inverse-addition-pair}
\end{equation}
The instruction $P$ is used only by the teacher. Reversing the pair therefore provides removal supervision whose target is the untouched real image instead of a generated background.

\paragraph{Scene-aware proposals.}
Candidate additions are organized into 15 categories: human, animal, apparel and accessories, vehicles, furniture, household items, electronic devices, food and beverages, containers and tableware, sports and leisure, plants, buildings, infrastructure, tools and hardware, and natural elements. Each category contains a fine-grained object catalog. For every image, we sample three distinct coarse categories and provide them, the image, and its caption to GPT-4.1~\citep{openai2025gpt41}. The model proposes one scene-compatible object absent from the image and belonging to one of the sampled categories, returning its coarse category, object class, and insertion instruction. Restricting the proposal to one object reduces ambiguous additions, while mask recovery can still handle occasional multi-instance outputs.

Sampling is dynamically adjusted using accumulated category and object statistics. Over-represented categories or object classes are excluded from subsequent sampling, while proposals remain constrained by scene compatibility.

\paragraph{Addition teacher.}
We instantiate $G_{\mathrm{add}}$ with an off-the-shelf image editing model. It receives the real image and a short ``Add \ldots'' instruction and produces only $I_{\text{input}}$, while the original is preserved exactly as $I_{\text{output}}$. We retain only spatially aligned pairs. Because the teacher does not provide an accurate object mask, we recover it from the aligned before/after pair.

\paragraph{DCT--DINOv3--SAM2 mask recovery.}

A raw RGB difference is sensitive to texture noise and incidental appearance changes. We instead form two complementary maps. First, we retain only low-frequency DCT coefficients in each color channel, reconstruct both images, and compute their absolute difference $D_{\mathrm{DCT}}$. This suppresses high-frequency generation noise while retaining object-scale pixel changes. Second, we compare aligned DINOv3~\citep{simeoni2025dinov3} dense features using cosine distance,
\begin{equation}
    D_{\mathrm{DINO}}(i,j)=1-
    \frac{f_{\text{input}}(i,j)^\top f_{\text{output}}(i,j)}
    {\lVert f_{\text{input}}(i,j)\rVert_2
     \lVert f_{\text{output}}(i,j)\rVert_2}.
    \label{eq:dino-difference}
\end{equation}
Here $f_{\text{input}}$ and $f_{\text{output}}$ denote the dense features of $I_{\text{input}}$ and $I_{\text{output}}$, respectively. Thresholding and morphological opening/closing give binary maps $B_{\mathrm{DCT}}$ and $B_{\mathrm{DINO}}$. When both are nonempty, we use
\begin{equation}
    B_{\mathrm{coarse}}=B_{\mathrm{DCT}}\cup B_{\mathrm{DINO}},
    \qquad
    B_{\mathrm{pos}}=B_{\mathrm{DCT}}\cap B_{\mathrm{DINO}}.
    \label{eq:difference-fusion}
\end{equation}
The union maximizes recall for the possible edit extent, while the intersection provides conservative positive-point candidates.  Cross-signal IoU is retained as a filtering indicator.

\paragraph{SAM2-based mask refinement.}
To recover object boundaries from the coarse change maps, we apply SAM2~\citep{ravi2024sam2} to the object-added image $I_{\text{input}}$ using sparse point prompts. Positive candidates are preferentially selected from the interior of $B_{\mathrm{pos}}$ using a distance transform, while negative candidates are sought outside $B_{\mathrm{coarse}}$, preferably in a nearby exterior band. Farthest-point sampling selects up to eight positive and four negative points per image to obtain spatially distributed prompts.

The sampled positive points are then grouped by the 8-connected components of $B_{\mathrm{pos}}$. Each component receiving a positive point is processed separately with its own positive prompts and the shared negative prompts. We retain the highest-scoring SAM2 prediction from each group and take their union as the final mask $M$. The coarse change maps guide prompt placement but are not directly merged into the final mask.

\paragraph{Object-focused masks and associated effects.}
Our mask recovery combines low-frequency DCT differences for photometric change detection, DINOv3 feature differences for semantic localization, and SAM2 for object-boundary refinement. The DCT branch can capture spatially extended appearance changes, such as cast shadows, while DINOv3 provides complementary semantic cues for localizing the added object. An associated effect may therefore enter the high-recall region $B_{\mathrm{coarse}}$ through the DCT branch even when the shared positive region $B_{\mathrm{pos}}$ remains concentrated on the object body. SAM2 can then produce an object-focused selection mask without requiring it to cover the entire detected change region.

Because $I_{\text{output}}$ is the original image before insertion, the full input--target pair still contains the corresponding associated-effect changes, including those outside $M$. Such pairs provide supervision for removing the selected object together with its associated effects, without requiring every effect pixel to be explicitly included in the selection mask.

\paragraph{General Data pair filtering.}
We screen General Data for unreliable localization, ineffective additions, and unintended scene changes. The remaining pairs are assessed using the following measures:

\begin{itemize}
    \item \textbf{Cross-signal agreement.}
    The intersection-over-union between $B_{\mathrm{DCT}}$ and $B_{\mathrm{DINO}}$ measures agreement between photometric and semantic change localization. Pairs flagged for low agreement are rejected to reduce supervision based on inconsistent estimates of the edited region.

    \item \textbf{Mask coverage.}
    The recorded mask-to-image area ratio identifies excessively broad selections. We reject masks with excessive coverage to limit cases in which the recovered selection extends over a large fraction of the scene.

    \item \textbf{Within-mask change.}
    We compute input--target PSNR within $M$ and reject pairs with unusually high similarity. Such cases exhibit insufficient change in the selected region, as can occur when an insertion fails or the mask includes substantial unchanged background.

    \item \textbf{Outside-mask consistency.}
    Input--target PSNR over the complement of $M$ measures aggregate changes outside the selection. Low values trigger rejection, targeting outpainting-induced shifts, global appearance drift, and broad background regeneration. This check bounds aggregate discrepancy without requiring every exterior pixel to remain identical.

    \item \textbf{Feature-level neighborhood change.}
    We compare SAM2 features in an exterior band separated from $M$ by a guard margin to reduce contamination from receptive fields overlapping the inserted object. One minus the mean cosine similarity of the least-similar patches provides a score for suspected replacement or damage to neighboring content.

    \item \textbf{Pixel-level neighborhood change.}
    A complementary score combines the fraction of changed pixels in an adjacent exterior ring with the strength of the largest RGB differences. It measures the spatial extent and intensity of nearby appearance changes, providing pixel-level evidence for the replacement check.

\end{itemize}

The feature- and pixel-level neighborhood scores form a joint replacement criterion: this criterion triggers only when both scores indicate substantial change. The other rejection criteria are checked separately.

\subsubsection{Alpha-supervised watermark synthesis}
\label{app:watermark-pipeline}

Watermark-removal pairs are generated by compositing rendered overlays onto clean images, providing known watermark support without image-difference-based mask recovery.

\paragraph{Assets and typography.}
We combine a pool of approximately 2K logo assets, in raster or SVG format, with template-based multi-lingual watermark text. The text pool includes copyright and ownership notices, preview and sample labels, usage restrictions, photographic attributions, website addresses, and identifiers. Templates are instantiated with sampled company names, domains, years, and numeric IDs. Font faces and sizes are randomized.

\paragraph{Layout and appearance.}
The renderer samples six synthesis modes: individual logos, standalone text, tiled logos or text, large diagonal text, stripe patterns with optional embedded text, and corner logos. Mode-specific randomization controls position, scale, orientation, opacity, color, and repetition spacing. Additional appearance variations include text outlines or offset shadows, slight logo blur, and translucent backdrops for corner marks. These combinations produce both localized overlays and repeated patterns extending across the image.

\paragraph{Triplet construction.}
Let $(W,\alpha)$ denote the rendered watermark layer in image coordinates, with $\alpha\in[0,1]$ incorporating the sampled opacity and any rendered decorations. Alpha compositing gives
\begin{equation}
    I_{\text{input}}=\alpha W+(1-\alpha)I_{\text{output}},
    \qquad M=\mathbf{1}[\alpha>0].
    \label{eq:watermark-composite}
\end{equation}
The clean image is retained as $I_{\text{output}}$, while $M$ is derived directly from the rendered alpha support, including disconnected glyphs, repeated elements, semi-transparent marks, and any rendered outlines, shadows, or backdrops.

\subsubsection{Generated large-foreground compositing}
\label{app:large-object-pipeline}

Large removal regions are relatively under-represented in General Data. We complement this branch by generating isolated foregrounds and compositing them onto real images. Separating foreground generation from placement allows us to vary object appearance, location, and visible area while retaining the pre-compositing image as the clean removal target.

\paragraph{Foreground diversity.}
We sample from a catalog of 25 coarse categories and approximately 2.5K object descriptions, spanning people, animals, vehicles, furniture, plants, and everyday objects. Generation prompts combine the sampled object description with viewpoint or pose descriptors, including frontal, side, three-quarter, and top-down views, using varied studio-photography templates that encourage isolated objects on white backgrounds.
BiRefNet~\citep{zheng2024birefnet} predicts the foreground alpha matte. We discard degenerate extractions with negligible or nearly frame-filling foreground support, then apply random in-plane rotation and crop to the foreground bounding box. Viewpoint variation and in-plane transformations diversify both object appearance and the geometry of the resulting masks.

\paragraph{Background-aware placement.}
To vary image position while limiting interference with existing content, we apply YOLOE~\citep{wang2025yoloe} to estimate foreground occupancy in each background image. Candidate anchors are sampled across the central image region, with preference given to locations whose neighborhoods contain less occupied area. Placement aligns the foreground's alpha-weighted centroid with the selected anchor, positioning irregular objects according to their foreground mass rather than the center of their bounding box.

\paragraph{Visible-area and occlusion control.}
We sample a desired mask-area ratio uniformly from 20\%--60\% and adjust foreground scale to approach this ratio. Coverage is measured from the foreground actually visible after placement and clipping to the image. We constrain occlusion at two levels: the covered fraction of each sufficiently large detected background instance, and the covered fraction of the union of all detected instances. Placements causing excessive occlusion are retried or discarded. These constraints allow partial overlap while limiting severe loss of visible context. The realized mask area is checked again after the final compositing steps, and samples outside the designated large-mask range are discarded.

\paragraph{Appearance adjustment and triplet construction.}
We partially adjust foreground color statistics toward those of the local background and feather the alpha boundary to reduce appearance and boundary discontinuities. Let $F$ and $\alpha$ denote the adjusted foreground and its alpha matte in final image coordinates, with $\alpha\in[0,1]$. We construct
\begin{equation}
    I_{\text{input}}=\operatorname{Composite}(I_{\text{output}},F,\alpha),
    \qquad M=\mathbf{1}[\alpha>0.5].
    \label{eq:large-mask-composite}
\end{equation}
The pre-compositing image remains $I_{\text{output}}$, so the background behind the inserted foreground is known rather than hallucinated by a removal teacher. The selection mask follows the transformed foreground silhouette, providing object-shaped, large-area removal supervision with a fully specified target.

\subsubsection{OCR-verified partial visual text removal}
\label{app:scene-text-pipeline}

A full-image text-removal teacher may leave residual lettering or alter unrelated scene content. Rather than adopting its output directly as the training target, we use it to propose local reconstructions and construct selection-specific removal pairs through OCR-guided region selection and local blending.

\paragraph{Text localization and teacher editing.}
Given the original text-containing image $I_{\text{input}}$, we apply PaddleOCR~\citep{cui2025paddleocr30technicalreport} with the PP-OCRv6 medium detector and recognizer to obtain text polygons. An image editing teacher receives the source image and a global instruction to erase text while preserving objects, signs, boards, and surfaces, producing a teacher-edited image $\tilde{I}$. We then run PaddleOCR on $\tilde{I}$ to obtain post-edit text detections. We use a higher OCR-confidence threshold on the source image to prioritize reliable text localization and a lower threshold after editing to increase sensitivity to weak residual text. The two OCR passes provide the source locations and residual-text evidence used by the eligibility assessment below.

\paragraph{Grouped selection and mask construction.}
We group overlapping or nearby text polygons in the source image into joint selection units. A nonempty subset of eligible groups is randomly selected for removal, with eligibility determined by the OCR-based criterion described below. Group-level selection couples the treatment of neighboring or overlapping text regions, while subset sampling varies the removal scope without requiring a separately prompted teacher edit.
The union of the selected text polygons, with a small dilation, defines the binary condition mask $M$. This retains the selected regions' spatial geometry without replacing each group with its enclosing rectangle.

\paragraph{Region-restricted target construction.}
Starting from $I_{\text{input}}$, we blend the selected groups' teacher-edited regions back into the source through arbitrary-mask Poisson blending~\citep{perez2023poisson}. Blending is performed group by group using the corresponding dilated polygon masks to reduce color and exposure seams from direct patch replacement. The resulting composite is $I_{\text{output}}$: the source image supplies unselected text and surrounding content, while the teacher supplies candidate reconstructions for the selected removal regions. This local reuse of teacher edits constructs partial-removal targets without adopting the teacher's full-image appearance changes. Removal effectiveness and blending quality are assessed using the criteria below.

\paragraph{Visual Text Data pair filtering.}
We assess text-removal effectiveness separately from artifacts introduced by local paste-back. Failed construction records and missing, undecodable, or empty masks are excluded. Let $\tilde{I}$ denote the teacher-edited image before paste-back. The filtering criteria are:
\begin{itemize}
    \item \textbf{OCR-based removal eligibility.}
    Original text detections are matched to post-edit detections by bounding-box overlap. A group is selectable only when none of its members has a post-edit match. Candidates without a selectable group are discarded. This prevents regions with detected residual text from entering the partial-removal target.

    \item \textbf{Structural change in selected text.}
    We compute RGB SSIM~\citep{wang2004image} between $I_{\text{input}}$ and $\tilde{I}$ within each selected group's dilated polygon support. High similarity indicates insufficient structural change despite the absence of an OCR match. We reject candidates with a high maximum SSIM across groups, so ineffective removal in one group cannot be averaged away by successful edits elsewhere.

    \item \textbf{Boundary gradient discrepancy.}
    From the paste-back residual $R=I_{\text{output}}-\tilde{I}$, we compute per-channel Sobel gradient magnitudes and take their maximum across RGB channels at each pixel. For each selected group, we average this magnitude in an inner boundary band. A high maximum group mean triggers rejection to screen for boundary leakage introduced by blending.

    \item \textbf{Regional gradient discrepancy.}
    Using the same residual-gradient map, we compute the 90th-percentile magnitude over each group's full selected region.
    A high maximum across groups triggers rejection to screen for strong internal discrepancies relative to the teacher reconstruction, complementing the boundary-focused check.
\end{itemize}
The two gradient criteria are applied independently. They measure discrepancies introduced by paste-back rather than penalizing image gradients that are also present in the teacher reconstruction.

\subsubsection{Layer-aware poster rendering}
\label{app:poster-pipeline}

We construct poster-removal pairs from explicit color--alpha layers with known z-order. The same rendered design defines both the complete poster and a selected-layer removal target, allowing visual diversity without independently regenerating the two images.

\paragraph{Canvas and layout sampling.}
Background images are fitted to canvases with varied resolutions and square, portrait, landscape, and widescreen aspect ratios. Layout templates guide initial element placement through header--body--footer, centered, grid, sidebar, full-bleed, and diagonal arrangements, alongside unrestricted layouts. Sample plans balance canvas types, layout templates, element types, and element and removal counts, with an initial budget of 4--9 elements. Element sizes are also varied to encourage coverage of different removal extents.

\paragraph{Element and typographic diversity.}
Elements span text, geometric shapes, decorative graphics, synthetic logos and icons, and presentation-style components such as bullet lists, tables, and chart placeholders. Text pools distinguish titles, subtitles, paragraphs, labels, prices, dates, URLs, and mixed-script content, with multi-lingual examples. Fonts are sampled according to typographic role and, for CJK strings, character coverage. Font size and line wrapping are adapted to the allocated region. We further vary color, opacity, alignment, filled versus outlined shapes, and optional drop shadows or outer glows.

\paragraph{Layer-consistent pair construction.}
Each element is rendered independently and placed in z-order. Layout constraints prevent overlap between text-element bodies, limit excessive occlusion and overlap depth, and check text contrast against the content beneath it. Retained text receives additional protection against severe occlusion, including checks at the character-region level. The planned removal set is refined by excluding candidates overlapped by higher layers designated for retention. We then reuse the identical background and rendered layers to produce $I_{\text{input}}$ with all accepted elements and $I_{\text{output}}$ with the selected layers omitted. The content, appearance, position, and order of the remaining layers are held fixed.

\paragraph{Body-focused mask construction.}
For mask construction, we use the element-body layers before shadow or glow is added. A front-to-back traversal assigns each pixel to the topmost body layer with nonzero alpha. Pixels assigned to selected layers are unioned and slightly dilated to obtain $M$. This defines a body-focused selection instead of a bounding-box mask. The paired images nevertheless remove the full selected layers, including their rendered effects, so supervision for effect removal can extend beyond $M$ without requiring an exhaustive effect mask.

\subsection{SFT Data Curation and Quality Screening}
\label{app:sft_data}

We curate SFT examples from General, Watermark, Visual Text, and Poster Data, emphasizing high-resolution candidates. Construction follows the domain-specific procedures summarized in Section~\ref{sec:sft-data}, including professionally designed posters rather than the procedural layouts used for pretraining. GPT-5.6 Sol~\citep{openai2026gpt56sol} examines each aligned triplet $(I_{\text{input}},I_{\text{output}},M)$ using the source image, a mask-overlay visualization, and the target image. When the selected region is difficult to inspect at full-image scale, aligned mask-centered crops of the source and target supplement the full-image views to reveal fine residuals and reconstruction artifacts.

\begin{itemize}
    \item \textbf{Mask validity.}
    The selection must identify one or more intended targets without substantial leakage into unrelated regions. Coarse boundaries and limited inclusion of nearby background or target-associated effects are permitted when the erase request remains clear.

    \item \textbf{Edit correctness and preservation.}
    The source--target pair must realize the intended removal completely while preserving unrelated content. Target remnants, unintended deletions, and spatial misalignment lead to rejection.

    \item \textbf{Reconstruction quality.}
    The revealed region must have plausible geometry and texture without conspicuous editing artifacts. We reject seams, smearing, ghosting, abnormal blur, and implausible local reconstruction.
\end{itemize}

For General Data only, we additionally require natural photographs or credible photorealistic renderings, excluding illustrations, documents, and poster-like graphics. A triplet is accepted only when its mask is valid, all pair-quality criteria pass, and the overall judgment has medium or high confidence without contradictory failure evidence.
Preservation does not require pixel-wise equality outside the intended edit. Screening focuses on correctly realizing the erase request without imposing additional scene-level constraints unrelated to it.

\subsection{RL Data Pipeline}
\label{app:rl-data-pipeline}

\paragraph{Task definition and composition.}
The dataset contains 19K tasks: 13K General Data examples and 2K examples each from Poster Data, Watermark Data, and Visual Text Data. General Data comprises 8K difficulty-filtered tasks and 5K supplementary tasks. Each task conditions LensErase only on the source image ($I_{\mathrm{input}}$) and selection mask ($M$). General Data requires no paired target, whereas Poster Data and Watermark Data retain deterministic clean references and Visual Text Data retains verified weak references. Available reference images and structured annotations support data curation and reward evaluation but are not additional inputs to the erase model. During RL, domains are sampled at a ratio of General:Poster:Watermark:Visual Text $= 6{:}2{:}1{:}1$, independently of their dataset cardinalities; the corresponding reward ablation protocol is reported in Section~\ref{sec:reward_ablation}.

\paragraph{General Data: failure-mode-guided collection and effect mining.}
The 5K supplementary tasks augment the difficulty-filtered subset through failure-mode-guided collection and targeted associated-effect mining. During extensive internal evaluation of LensErase-SFT, we manually analyze recurring failures and identify challenging scenarios. We then collect images presenting similar erasure challenges.

For associated-effect mining, GPT-5.6 Sol~\citep{openai2026gpt56sol} examines candidate images and selects those containing visibly attributable object-associated effects. These effects are already present in the images and are not synthetically added. They include cast shadows, reflections, water ripples and wakes, splashes, smoke or steam, emitted light, refraction or caustics, contact traces, and motion trails when their causal relation to a selected object is visually supported.

\paragraph{General Data: object localization and mask construction.}
For newly annotated tasks, GPT-5.6 Sol identifies the target objects and returns their physical-object bounding boxes. For effect-focused tasks, these boxes localize the objects responsible for the effects rather than the effects themselves. Source images and bounding-box overlays are used to verify object localization and, where present, the visibility and causal attribution of associated effects. We then provide the image and verified object boxes to SAM~2.1~\citep{ravi2024sam2} to segment the object bodies, taking the union of masks when multiple objects are selected. Further VLM checks assess object-body completeness, instance identity, the natural-image domain, and leakage into non-target content.

The mask is an object-selection cue, not an exhaustive annotation of all pixels that should change. We do not automatically dilate or close it to absorb shadows or reflections. Accurately and compactly captured effects may be included, but do not justify covering large unrelated regions. Only after mask construction and validation do we analyze the finalized source--mask task and assess its difficulty using the procedure below. Thus, difficulty is evaluated for the actual selection, not for the source image or proposed bounding boxes alone.

\paragraph{General Data: quality filtering and difficulty screening.}
We separate source--mask validity from task difficulty. After mask construction, we exclude invalid conditions, including unidentifiable targets and degenerate masks, and suppress exact and near-duplicate source--mask pairs. The strict pool additionally excludes ambiguous or highly irregular selections, including masks with substantial overlap with independent non-target objects. Thus, strict filtering establishes usable erase conditions, while task difficulty is assessed separately.

We then apply a two-stage offline analysis using GPT-5.6 Sol~\citep{openai2026gpt56sol}. First, the VLM establishes a structured understanding of the selected targets, their constituent parts, associated effects and effect bounding boxes, and nearby independent content to preserve. Second, it estimates the intrinsic difficulty of the same source--mask task without observing an erase-model output. Five integer scores $s_k\in\{0,\ldots,4\}$ assess, in order, removal scope and associated effects; reconstruction and fine-detail recovery; occlusion and object interactions; preservation and locality risk; and the risk of introducing unwanted entities. With respective weights $(w_1,\ldots,w_5)=(0.25,0.30,0.20,0.20,0.05)$, overall difficulty is
\begin{equation}
    D=10\sum_{k=1}^{5}w_k\frac{s_k}{4}.
    \label{eq:rl-task-difficulty}
\end{equation}
Tasks are categorized as Easy ($D\leq2$), Moderate ($2<D\leq4$), Challenging ($4<D\leq6$), Hard ($6<D\leq8$), or Extreme ($8<D\leq10$). High ratings must reflect multiple substantive challenges, not mask area alone. From the strict pool, we retain 6,991 Hard and 1,009 Extreme tasks, yielding the 8K difficulty-filtered subset.
This score is used for task selection, not as the training reward.

Effects outside the selection remain explicitly described in the source understanding for reward evaluation. Final masks are stored as single-channel binary PNGs with values 0 and 255 at the source-image resolution, and the associated understanding is bound to hashes of the finalized source and mask.

\paragraph{Poster Data: high-resolution layered removal.}
The RL Poster Data subset contains 2,000 unique high-quality poster examples designed by professional designers. We rank candidate designs using their native canvas area, side lengths, and layer count, and construct the subset from an ordered pool of 2,500 candidates, replacing ineligible designs until 2,000 valid tasks are obtained. The source canvases are not downscaled during pair construction. Each design contributes one task that removes a single visible non-background layer, which may contain text, graphics, icons, color blocks, or an image. The selected layer may lie behind other foreground layers. Rendering the same layer stack with and without the selected layer gives the source image and an exact removal reference ($I_{\mathrm{ref}}$), following the layer-aware construction principle described in Appendix~\ref{app:poster-pipeline}.

We derive the mask from the final rendered RGB difference,
\begin{equation}
M(x)
=
\mathbf{1}\!\left[
\exists\, c \in \{R,G,B\} :
I_{\mathrm{input}}(x,c)
\neq
I_{\mathrm{ref}}(x,c)
\right],
\label{eq:poster_diff_mask}
\end{equation}
rather than directly using the selected layer's alpha support. This excludes occluded or otherwise unchanged pixels from the selection. Unselected foreground content that remains visible over the selected layer's spatial extent is part of the correct reference, not evidence of incomplete removal. Candidate selection encourages diversity in removal area, element type, and layer depth, while prioritizing high resolution without enforcing equal counts across these attributes.

\paragraph{Watermark Data: controlled synthesis and mask validation.}
For the RL watermark subset, we start with a curated pool of 25,000 high-resolution natural images. We first retain images whose original width and height are both at least 1,440 pixels, yielding 24,688 eligible backgrounds. From this eligible pool, we select 2,000 distinct images and use each as the clean background for one watermark-removal task, producing 2,000 tasks in total. Before watermark synthesis, we resize each selected image as needed to limit its long side to 2,048 pixels while preserving its aspect ratio. The upstream asset library combines logo resources with procedurally generated stamps, signatures, patterns, machine-readable codes, and multilingual text. Asset provenance describes the candidate library; the assets actually used by accepted samples are recorded in the sample manifest. Following the synthesis formulation in Appendix~\ref{app:watermark-pipeline}, we composite sampled assets onto the clean canvas while varying layout, color, opacity, scale, and rotation, retaining the pre-compositing image as the exact clean reference.

Precise change regions are computed from the final source--reference RGB differences. Half of the tasks use precise masks, while the other half use user-style coarse local selections, such as rectangular or rounded regions, that cover all actual changes without substituting an indiscriminate whole-image box. We validate image--reference--mask dimensions, mask binarity, and change coverage. The final subset enforces marginal quotas over six dimensions: ten asset families with 200 tasks each; four layouts (corner, central, diagonal band, and tiled) with 500 each; synthesized watermark-area bins of $0.1$--$1\%$, $1$--$5\%$, $5$--$15\%$, and $15$--$45\%$ with 500 each; four opacity bins with 500 each; five requested color families with 400 each; and precise versus coarse masks with 1,000 each. These are marginal quotas, not a requirement to balance every joint attribute combination.

\paragraph{Visual Text Data: curated weak references.}
The 2,000 Visual Text tasks use the OCR-verified partial-removal procedure in Appendix~\ref{app:scene-text-pipeline}. Pre- and post-edit OCR identify candidate removals, and selected teacher-edited regions are blended back into the source. RL curation further applies VLM screening and text-centered cropping. When selected text occupies only a small part of the image, we crop around the target text, applying the same crop to the source image, selection mask, and weak reference. This increases the selected text's relative extent in the resulting task while maintaining alignment. Selection also considers effective image resolution and mask coverage. The target remains a verified weak reference, since the hidden text-free appearance is not observed directly. Reference-guided rewards therefore assess removal and reconstruction without requiring every plausible completion to match the teacher reconstruction exactly, as detailed in Appendix~\ref{app:removal_reward}.

\paragraph{Interface to reward evaluation.}
Offline source understanding supports task curation and validation; reward evaluation follows the candidate-independent contract construction in Section~\ref{sec:removal_rl} and Appendix~\ref{app:removal_reward}. For General Data, the erase contract and aligned detail-view geometry are defined after the training geometry transform and fixed across candidates for the same condition. Poster Data, Watermark Data, and Visual Text Data use their domain-specific references as described in that appendix. These reward-side specifications do not change the model's source-image--mask conditioning interface. Training resolution buckets and RL optimization are described in Appendices~\ref{app:supervised_sampling} and~\ref{app:nft_implementation}, respectively.

\FloatBarrier
\section{LensErase-Bench: Construction and Evaluation}
\label{app:benchmark_construction}

LensErase-Bench contains 500 cases spanning four erase domains: 200 General cases and 100 each for Poster, Watermark, and Visual Text. The benchmark is curated to cover diverse mask scales, target counts, image resolutions, and removal difficulty. Evaluation follows the reference-free protocol in Appendix~\ref{app:benchmark_protocol}.

\subsection{Selection and Coverage}
\label{app:benchmark_selection}

\paragraph{Mask-area coverage.}
We stratify each domain by mask area to ensure that the benchmark covers both highly localized edits and large-region removal. Table~\ref{tab:bench_area_quotas} reports the resulting domain-specific distribution. General images are evenly distributed across five mask-area ranges, while Poster, Watermark, and Visual Text use bins adapted to their characteristic edit scales.

\begin{table}[!htbp]
  \centering
  \caption{Domain-specific mask-area distribution in LensErase-Bench.}
  \label{tab:bench_area_quotas}
  \begingroup
  \fontsize{9}{11}\selectfont
  \setlength{\tabcolsep}{3pt}
  \renewcommand{\arraystretch}{1.10}
  \begin{tabular*}{\linewidth}{@{\extracolsep{\fill}}lcc@{}}
    \toprule
    Domain & Mask-area bins (\%) & Cases per bin \\
    \midrule
    General & $<1,\ [1,5),\ [5,15),\ [15,30),\ \geq30$ & 40 / 40 / 40 / 40 / 40 \\
    Poster & $<5,\ [5,15),\ \geq15$ & 35 / 33 / 32 \\
    Watermark & $<1,\ [1,5),\ [5,15),\ \geq15$ & 25 / 25 / 25 / 25 \\
    Visual Text & $<0.5,\ [0.5,1),\ [1,2),\ \geq2$ & 25 / 25 / 25 / 25 \\
    \bottomrule
  \end{tabular*}
  \endgroup
\end{table}

\paragraph{Attribute coverage.}
Beyond mask area, we curate complementary attributes that characterize practical erase requests. The General subset is evenly split between cases with and without associated visual effects, and between single-target and multi-target removal. Poster cases are approximately balanced across graphic, image, and text layers, as well as back, middle, and front layer depths. Watermark cases evenly cover ten asset families, four spatial layouts, four opacity ranges, and precise versus user-style coarse masks. Visual Text includes diverse languages, layouts, and background types.

\paragraph{Difficulty and target count.}
GPT-5.6 Sol~\citep{openai2026gpt56sol} rates the filling difficulty of each case on a 0--4 scale. Table~\ref{tab:bench_difficulty_targets} reports the distribution of filling difficulty together with the number of selected semantic targets. Across the full benchmark, 59.2\% of cases contain a single selected target and 40.8\% contain multiple targets, with Watermark providing particularly dense multi-target cases.

\begin{table}[!htbp]
  \centering
  \caption{Distribution of filling difficulty and number of selected targets in LensErase-Bench.}
  \label{tab:bench_difficulty_targets}
  \begingroup
  \fontsize{9}{11}\selectfont
  \setlength{\tabcolsep}{3pt}
  \renewcommand{\arraystretch}{1.10}
  \begin{tabular*}{\linewidth}{@{\extracolsep{\fill}}l*{7}{c}@{}}
    \toprule
    & \multicolumn{4}{c}{Filling difficulty} & \multicolumn{3}{c}{Selected targets} \\
    \cmidrule(lr){2-5}\cmidrule(l){6-8}
    Domain & 0--1 & 2 & 3 & 4 & 1 & 2--3 & $\geq4$ \\
    \midrule
    General & 9 & 50 & 87 & 54 & 100 & 87 & 13 \\
    Poster & 34 & 25 & 25 & 16 & 93 & 3 & 4 \\
    Watermark & 0 & 9 & 64 & 27 & 40 & 16 & 44 \\
    Visual Text & 25 & 45 & 30 & 0 & 63 & 25 & 12 \\
    \bottomrule
  \end{tabular*}
  \endgroup
\end{table}

\paragraph{Resolution coverage.}
All cases preserve their original aspect ratio and are normalized without cropping or padding. The benchmark spans four long-side resolution ranges, $\leq1024$, 1025--1280, 1281--1536, and 1537--2047 pixels. Each range contains 125 cases: 50 General cases and 25 from each of Poster, Watermark, and Visual Text. This produces balanced resolution coverage across domains while retaining the original selection geometry.
\FloatBarrier

\subsection{Evaluation Protocol}
\label{app:benchmark_protocol}

LensErase-Bench uses the four evaluation dimensions defined in Section~\ref{sec:lenserase_bench}: Removal, Inpainting, Preservation, and Cleanliness. GPT-5.6 Sol evaluates each candidate using a shared rubric across all four domains, with domain-specific interpretation only where the semantics of the selected target differ.

\paragraph{Evaluation inputs.}
For each case, the judge receives four aligned full-image views: the source image, the source image with the selection overlaid, the candidate output, and the binary selection mask.
No clean reference is provided.

\paragraph{Scoring rubric.}
Each case receives an integer score from 0 to 4 for each dimension, where 4 denotes full satisfaction of the criterion, 3 a minor defect, 2 a clear localized failure, 1 a major failure, and 0 a fundamentally unsuccessful result. Removal measures whether all selected targets and associated effects disappear; Inpainting measures the plausibility and local consistency of newly revealed content; Preservation measures fidelity of non-target content; and Cleanliness measures artifacts and unrelated additions. Multi-target cases are judged jointly, so a missed selected instance remains a Removal failure even if the remaining targets are successfully erased. To avoid double counting, a single visual defect is attributed to the most directly affected dimension.

\paragraph{Scores and aggregation.}
Let $r_i^{\mathrm{rem}}, r_i^{\mathrm{inp}}, r_i^{\mathrm{pre}}, r_i^{\mathrm{cln}} \in \{0,1,2,3,4\}$ denote the Removal, Inpainting, Preservation, and Cleanliness ratings for case $i$. Each dimension is mapped to $0$--$100$ by multiplying its rating by $25$. The case-level Overall Score is
\begin{equation}
  o_i = 25\left[0.4r_i^{\mathrm{rem}} + 0.2\left(r_i^{\mathrm{inp}} + r_i^{\mathrm{pre}} + r_i^{\mathrm{cln}}\right)\right].
  \label{eq:benchmark_overall}
\end{equation}
For each method, we report the mean of each dimension and Overall across its $N$ valid judged cases, with equal weight assigned to every case. Here, $N=500$ for all methods except PANDORA, for which $N=499$ because one output was consistently rejected by the judge's safety filter; see Table~\ref{tab:vlm-full}.

\paragraph{HiFi Removal.}
We additionally report the percentage of cases that achieve complete removal while maintaining high editing quality:
\begin{equation}
  \mathrm{HiFiRemoval} = \frac{100}{N}\sum_{i=1}^{N}
  \mathbbm{1}\!\left[r_i^{\mathrm{rem}}=4\ \land\ \min\{r_i^{\mathrm{inp}},r_i^{\mathrm{pre}},r_i^{\mathrm{cln}}\}\geq3\right],
  \label{eq:benchmark_remove_success}
\end{equation}
where $\mathbbm{1}$ is the indicator function. Overall summarizes average quality across the four dimensions, whereas HiFi Removal measures the fraction of cases that satisfy all removal and editing-quality requirements simultaneously.

\FloatBarrier

\FloatBarrier
\section{Pretraining Design and Ablations}
\label{app:pretraining_design}

This section provides additional implementation details and ablations for the initialization, mask conditioning, and two-stage adaptation strategy introduced in Section~\ref{sec:mask_conditioned_architecture}.

\subsection{Visual Initialization and Mask Conditioning}
\label{app:pretraining_initialization}

Both branches of LensErase process visual inputs. We retain the pretrained Lens image branch and initialize the condition branch from the same image-branch weights, while discarding the pretrained text-specific weights. The source image remains intact, preserving visual evidence of the selected target and its associated effects. Mask-Aware RoPE injects the selection mask by assigning different modality coordinates to masked and unmasked source tokens, as defined in Equation~\ref{eq:mask_rope_positions}.

\paragraph{Architecture and positional encoding.}
The image and condition streams operate on spatial token grids with effective stride 16 relative to the input image. The backbone contains 48 MMDiT blocks, with 24 attention heads per block and 64 channels per head. The three-dimensional RoPE allocates these channels across the modality, height, and width axes as $(8,28,28)$. Both streams use the centered spatial grid of the pretrained positional encoder. The binary selection mask is adaptively max-pooled to the source-token grid and thresholded at $0.5$. The resulting value selects between the two condition-stream modality indices in Equation~\ref{eq:mask_rope_positions}. The mask is therefore encoded only through RoPE and is not embedded into the latent channels. RoPE is applied to queries and keys in every block, while values are unchanged. The condition stream is initialized from the image stream's input projection and corresponding attention, feed-forward, normalization, and modulation parameters.

\paragraph{Alternative mask representations.}
For the two feature-based alternatives in Table~\ref{tab:structure_ablation}, we form mask features with $\operatorname{PixelUnshuffle}_{16}(M)$, yielding a 256-dimensional vector aligned with each source-image token. \emph{Channel Concat} concatenates this vector with the corresponding source feature and fuses them with a two-layer MLP before the condition projection. \emph{Token Concat} places source and noisy-image tokens together in the image branch and projects the mask features into the condition branch. Both alternatives use positional indices independent of mask values.

\paragraph{Text-branch initialization control.}
This control retains the pretrained text-Transformer weights for the condition branch while keeping the same visual inputs, branch routing, and Mask-Aware RoPE as LensErase. Its visual input projection is still initialized from the Lens image branch, and no text prompt is used. This isolates the effect of the pretrained prior used to initialize the condition Transformer.

\subsection{Region-Weighted Objective Details}
\label{app:supervised_objective}

Training follows the flow-matching path and region-weighted objective in Section~\ref{sec:mask_conditioned_architecture}. In implementation, the per-token squared velocity error is averaged over channels. Equation~\ref{eq:weighted_removal_fm} omits this fixed normalization factor for clarity. The conditioning mask uses the original selection, whereas the loss support uses a separate $31\times31$ pixel-space dilation, followed by adaptive max-pooling to the token grid and thresholding at $0.5$.

\subsection{Two-Stage Pretraining Schedule}
\label{sec:freeze_unfreeze_training}

\begin{figure}[!htb]
  \centering
  \includegraphics[width=\linewidth]{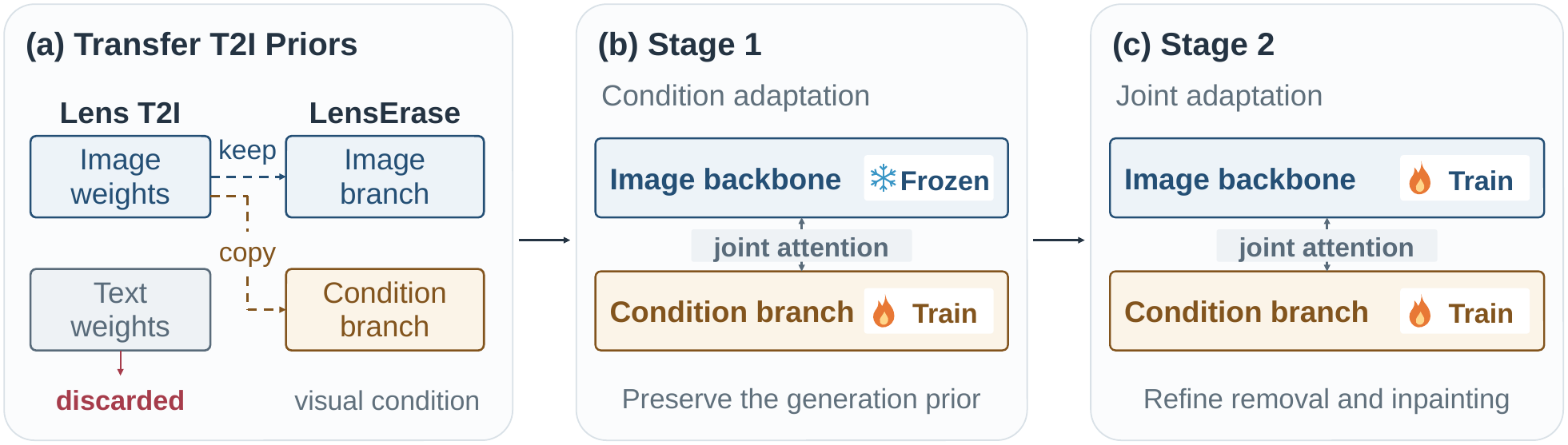}
  \caption{\textbf{Initialization and two-stage pretraining.} Both visual branches are initialized from the pretrained Lens image branch. Stage 1 adapts the condition branch while freezing the image-specific backbone. Stage 2 jointly adapts both branches for removal and reconstruction.}
  \label{fig:pretraining_design}
\end{figure}

To adapt the newly repurposed visual condition stream without immediately disrupting the pretrained image-generation prior, we use a two-stage schedule. Stage 1 trains the condition stream while freezing the image-specific backbone. Stage 2 then unfreezes the image backbone and jointly optimizes both streams using the same region-weighted flow-matching objective (Figure~\ref{fig:pretraining_design}).

In Stage 1, all image-specific parameters are frozen, including the image input projection, image-side attention projections, query/key normalization, modulation, layer normalization, and feed-forward layers. Condition-side parameters remain trainable, while the shared timestep embeddings and output normalization and projection are not frozen. Stage 2 jointly optimizes both streams. Pretraining uses nominal resolution buckets of 512, 768, 1024, 1280, and 1440, each with nine aspect ratios spanning $1{:}2$ to $2{:}1$.

\subsection{Initialization and Training Ablations}
\label{app:pretraining_ablations}

Table~\ref{tab:pretraining_design_ablation} compares alternative conditioning and initialization designs together with different pretraining schedules. The shared reference is the Stage 2 checkpoint obtained with Mask-Aware RoPE, image-branch initialization, and the two-stage schedule. \emph{Stage 1} denotes the endpoint of the initial frozen phase. \emph{No freeze} optimizes both streams from the outset, whereas \emph{Always freeze} continues the Stage 1 checkpoint for the second training phase while keeping the image-specific backbone frozen.

\begin{table}[!htbp]
  \centering
  \caption{\textbf{Pretraining design ablations.} The shared reference uses Mask-Aware RoPE, image-branch initialization, and two-stage freeze--unfreeze training. All rows use raw outputs. LensErase-Bench metrics follow Section~\ref{sec:lenserase_bench}; PSNR and LPIPS are equal-weight means over the six public benchmarks.}
  \label{tab:pretraining_design_ablation}
  \begingroup
  \fontsize{7.5}{8.9}\selectfont
  \setlength{\tabcolsep}{1.6pt}
  \renewcommand{\arraystretch}{1.08}
  \begin{tabular*}{\linewidth}{@{\extracolsep{\fill}}l*{6}{c}@{\hspace{6pt}}*{2}{c}@{}}
    \toprule
    & \multicolumn{6}{c}{\textbf{LensErase-Bench}} & \multicolumn{2}{c}{\textbf{Public Benchmarks}} \\
    \cmidrule(lr){2-7}\cmidrule(l){8-9}
    \raisebox{\dimexpr(\arraystretch\ht\strutbox+\arraystretch\dp\strutbox+\aboverulesep+\belowrulesep+\cmidrulewidth)/2\relax}[0pt][0pt]{\textbf{Model}} & \mbox{\textbf{Removal}\,$\uparrow$} & \mbox{\textbf{Inpainting}\,$\uparrow$} & \mbox{\textbf{Preservation}\,$\uparrow$} & \mbox{\textbf{Cleanliness}\,$\uparrow$} & \mbox{\textbf{Overall}\,$\uparrow$} & \mbox{\textbf{HiFi Removal}\,$\uparrow$} & \mbox{\textbf{PSNR}\,$\uparrow$} & \mbox{\textbf{LPIPS}\,$\downarrow$} \\
    \midrule
    \begin{tabular}[c]{@{}l@{}}\textbf{Mask-Aware RoPE}\\\textbf{(Stage 2)}\end{tabular} & 82.35 & 84.65 & 91.05 & 92.15 & 86.51 & 65.80 & 28.03 & 0.0855 \\
    \midrule
    \multicolumn{9}{@{}l}{\textit{Conditioning and initialization}} \\
    \addlinespace[1pt]
    Channel Concat & 81.50 & 84.55 & 82.90 & 91.15 & 84.32 & 58.80 & 27.84 & 0.0869 \\
    Token Concat & 77.50 & 76.85 & 84.90 & 87.15 & 80.78 & 51.40 & 27.41 & 0.0965 \\
    Text initialization & 89.50 & 1.30 & 0.05 & 1.75 & 36.42 & 0.00 & 10.71 & 0.7420 \\
    \addlinespace[3pt]
    \multicolumn{9}{@{}l}{\textit{Training schedule}} \\
    \addlinespace[1pt]
    Stage 1 & 79.10 & 78.95 & 89.40 & 90.05 & 83.32 & 60.40 & 26.74 & 0.1089 \\
    No freeze & 75.35 & 76.05 & 89.75 & 89.85 & 81.27 & 55.80 & 26.76 & 0.1020 \\
    Always freeze & 80.75 & 80.25 & 88.00 & 88.20 & 83.59 & 58.60 & 27.66 & 0.0864 \\
    \bottomrule
  \end{tabular*}
  \endgroup
\end{table}

\paragraph{Visual conditioning and initialization.}
Mask-Aware RoPE achieves 65.80\% HiFi Removal, compared with 58.80\% for Channel Concat and 51.40\% for Token Concat.
Text-branch initialization retains a high Removal score (89.50) but collapses Inpainting to 1.30 and HiFi Removal to 0.00\%, indicating that, in this setup, visual initialization is substantially more suitable for the condition stream than text-branch initialization.

\paragraph{Two-stage adaptation.}
The freeze--unfreeze schedule achieves 65.80\% HiFi Removal, compared with 55.80\% when both streams are optimized from the outset and 58.60\% when the image backbone remains frozen throughout. This supports first adapting the visual condition stream while preserving the pretrained image prior, followed by joint adaptation for removal and reconstruction.

\FloatBarrier

\FloatBarrier
\section{Post-Training and Inference Details}
\label{app:method_details}

\subsection{Supervised Fine-Tuning}
\label{app:supervised_sampling}

Appendix~\ref{app:pretraining_design} describes the initialization and pretraining schedule. SFT uses rank-64 LoRA with scaling parameter 64 and zero dropout, while keeping the VAE and backbone weights fixed. The adapters are applied to the joint query/key/value and attention-output projections of both streams. We retain the same image--mask conditioning interface and region-weighted flow-matching objective as in pretraining.
SFT is performed for 5,000 optimization steps using a learning rate of \(2 \times 10^{-5}\) with a cosine schedule.
SFT and RL use nominal resolution buckets of 512, 768, 1024, 1280, 1536, and 1792, with nine aspect ratios spanning $1{:}2$ to $2{:}1$.

\subsection{Contract-Guided Reward Construction}
\label{app:removal_reward}

\begin{figure}[!htb]
  \centering
  \includegraphics[width=\linewidth]{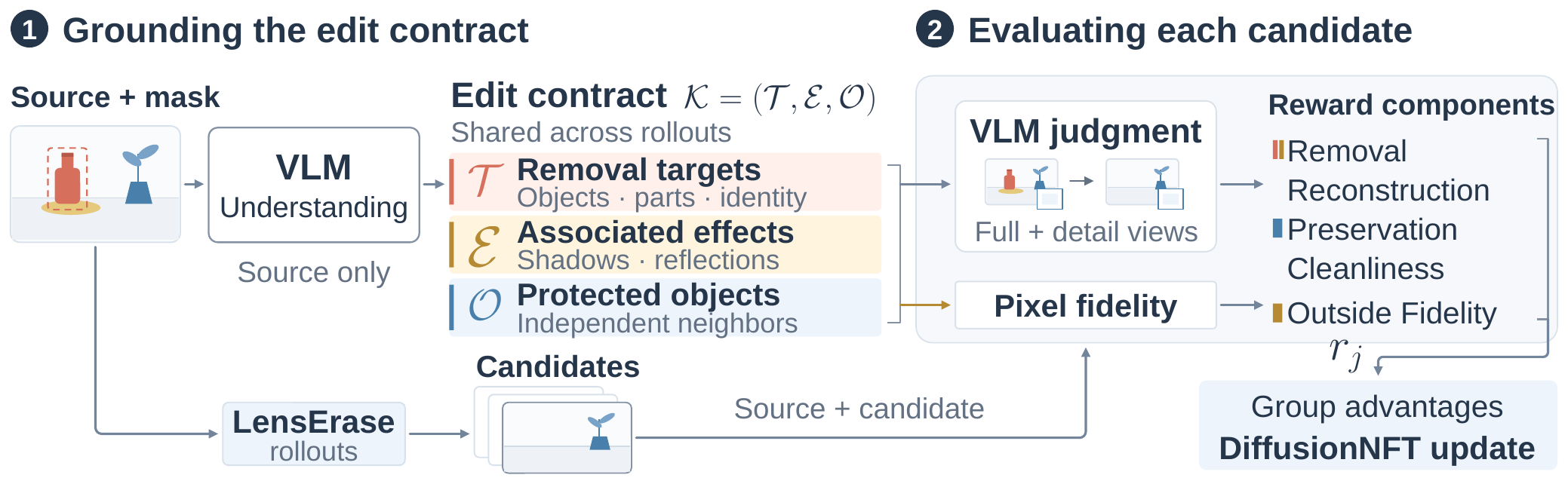}
  \caption{\textbf{Contract-guided reward construction for General Data.} Source analysis establishes an erase contract shared across rollouts. Each candidate is evaluated against this fixed contract using semantic judgments and pixel fidelity. The five reward components are combined and passed through the removal-aware gate in Equation~\ref{eq:removal_gate}. The resulting rewards define group advantages for DiffusionNFT.}
  \label{fig:rl_reward_pipeline}
\end{figure}

Following Section~\ref{sec:removal_rl}, reward construction for General Data first establishes a candidate-independent erase contract, then evaluates each candidate against this contract and applies the removal-aware reward gate (Figure~\ref{fig:rl_reward_pipeline}). GPT-5.6 Sol  serves as the VLM evaluator. Reference-guided adaptations for Poster, Watermark, and Visual Text are described below.

\paragraph{Establishing the erase contract.}
Before observing any candidate, the VLM analyzes the source image, selection mask, and mask-centered context to construct an erase contract $\mathcal{K}=(\mathcal{T},\mathcal{E},\mathcal{O})$. Here, $\mathcal{T}$ denotes the removal targets, $\mathcal{E}$ their clearly attributable visual effects, and $\mathcal{O}$ nearby independent content that should be preserved. The same contract and detail-view geometry are shared by all candidates of the same image--mask condition.

\paragraph{Evaluating candidate edits.}
Each candidate is evaluated against the fixed contract using the source image, selection mask, candidate output, and aligned detail views. Removal assesses whether $\mathcal{T}$ and $\mathcal{E}$ have been fully removed; Reconstruction measures the plausibility and local consistency of newly revealed content; Preservation assesses $\mathcal{O}$ and other non-target content; and Cleanliness captures residual fragments, unwanted additions, and technical artifacts. Outside Fidelity is computed separately from source--candidate PSNR over regions expected to remain unchanged.

\paragraph{Reward aggregation.}
VLM criteria are scored on a 0--4 scale and normalized to $[0,1]$, while computed fidelity scores are already normalized. For candidate $j$, let $s_{j,p,k}$ denote a normalized subscore and let $\mathcal{A}_{j,p}$ contain the enabled, applicable subcriteria for component $p$. We aggregate them as
\begin{equation}
  P_{j,p}=\sum_{k\in\mathcal{A}_{j,p}}\widetilde{w}_{j,p,k}s_{j,p,k},
  \label{eq:reward_parent_aggregation}
\end{equation}
where the weights are renormalized over applicable subcriteria. If an entire component is inapplicable, the remaining component weights are likewise renormalized. With the resulting weights $\widetilde{\lambda}_{j,p}$, the base reward is $b_j=\sum_p\widetilde{\lambda}_{j,p}P_{j,p}$. The final reward follows the removal-aware gate in Equation~\ref{eq:removal_gate}.

\begin{table}[htbp]
  \centering
  \caption{Reward weights before applicability renormalization and removal-aware gating. The last column summarizes the General Data criteria. Poster, Watermark, and Visual Text share component weights, with reference-specific subcriteria described below.}
  \label{tab:reward_weights}
  \small
  \setlength{\tabcolsep}{4pt}
  \begin{tabular}{@{}>{\raggedright\arraybackslash}p{0.18\linewidth}
    c c >{\raggedright\arraybackslash}p{0.48\linewidth}@{}}
    \toprule
    Component & General Data & \shortstack{Poster /\\Watermark /\\Visual Text}
    & General Data criteria \\
    \midrule
    Removal & 0.30 & 0.20 &
    Target body absence (0.35), parts/identity absence (0.30),
    associated-effect removal (0.35) \\
    Reconstruction & 0.15 & 0.20 &
    Semantic plausibility, structural continuity, and appearance/material
    consistency \\
    Preservation & 0.15 & 0.15 &
    Independent-object preservation (0.60),
    background/scene-structure preservation (0.40) \\
    Outside Fidelity & 0.30 & 0.25 & Outside-region PSNR score (1.00) \\
    Cleanliness & 0.10 & 0.20 &
    Absence of residual fragments, unwanted additions (objects, duplicates,
    text, logos, or watermarks), and technical artifacts \\
    \bottomrule
  \end{tabular}
\end{table}

\paragraph{Outside Fidelity.}
For General Data, the valid pixel region for candidate $j$ is
\begin{equation}
  \Omega_j=\mathcal{V}\setminus
  \left(M_{\mathrm{dil}}\cup B_{\mathcal{E}}
  \cup M_{\Delta,j}\right),
  \label{eq:reward_fidelity_region}
\end{equation}
where $\mathcal{V}$ excludes padding, $M_{\mathrm{dil}}$ is a mildly dilated selection, and $B_{\mathcal{E}}$ is the union of expanded source-stage effect boxes. $M_{\Delta,j}$ further excludes size- and strength-qualified source--candidate residual components, including a gamma-$0.75$ pass for weak changes in dark regions. These residual components may be spatially detached from the mask, as associated reflections or shadows can extend beyond it. If this supplemental exclusion would cover more than 45\% of the valid canvas, it is discarded while the selection and declared effect boxes are retained. Outside Fidelity is used only when at least 1,024 pixels and 8\% of the valid canvas remain. These exclusions define the support of the pixel-fidelity metric, while semantic Preservation is still evaluated over the full image.

For an applicable PSNR measurement $p_j$ in decibels, we map the interval $20$--$40$ dB to $[0,1]$ with a clipped smoothstep:
\begin{equation}
  u_j=\operatorname{clip}\!\left(\frac{p_j-20}{20},0,1\right),
  \qquad s^{\mathrm{abs}}_j=u_j^2(3-2u_j).
  \label{eq:absolute_psnr_reward}
\end{equation}
We combine this absolute quality anchor with relative differences among valid candidates of the same condition:
\begin{equation}
  s^{\mathrm{PSNR}}_j
  =\tfrac{1}{2}s^{\mathrm{abs}}_j+
    \tfrac{1}{2}\frac{p_j-p_{\min}}{p_{\max}-p_{\min}}.
  \label{eq:relative_psnr_reward}
\end{equation}
The relative term is used only when at least two valid measurements are available and $p_{\max}-p_{\min}>10^{-6}$ dB. Otherwise, the score remains $s^{\mathrm{abs}}_j$.

\paragraph{Reference-guided domains.}
Poster, Watermark, and Visual Text additionally use their clean or weak references. Their component weights are 0.20 for Removal, 0.20 for Reconstruction, 0.15 for Preservation, 0.25 for Outside Fidelity, and 0.20 for Cleanliness. Removal combines a deterministic reference-based score (0.60) with the VLM target-absence score (0.40). The deterministic score measures both progress toward the reference and suppression of the original source--reference residual over changed pixels. For Watermark, a spatial low-tail term additionally checks whether repeated overlay elements remain.

For Poster and Watermark, Reconstruction combines masked reference fidelity and visual recovery quality with weights 0.70 and 0.30. For Visual Text, whose weak reference may not represent a unique valid reconstruction, the corresponding weights are 0.40 and 0.60. Visual Text Preservation additionally balances retention of unselected text and scene content with the absence of newly introduced or replacement text. Cleanliness assesses residual fragments, unwanted additions, and technical artifacts in all three reference-guided domains. The PSNR blending in Equation~\ref{eq:relative_psnr_reward} applies to Outside Fidelity in every domain and to masked reference fidelity for Poster and Watermark; the inside text-reference score remains absolute.

\paragraph{Removal-aware reward gating.}
After component aggregation, the candidate-specific form of the removal-aware gate in Equation~\ref{eq:removal_gate} is
\[
  n_{0,j}=\sum_{k\in\mathcal{A}_{j,\mathrm{Removal}}}
  \mathbbm{1}[s_{j,\mathrm{Removal},k}=0],
  \qquad
  r_j=0.5^{\,n_{0,j}}b_j.
\]
Each zero-valued applicable Removal subcriterion therefore contributes a factor of $0.5$, so multiple removal failures compound without hard-rejecting a candidate solely because one subcriterion is zero. The same rule is applied across all data domains.

\subsubsection{Reward Prompt Excerpts}
\label{app:reward_prompts}

The following excerpts summarize the principal semantic instructions used by the training reward evaluator. They illustrate the edit semantics and domain-specific judgments underlying Section~\ref{sec:removal_rl}; serialization details and output schemas are omitted. LensErase-Bench uses the separate reference-free evaluation protocol in Section~\ref{sec:lenserase_bench}.

\begin{quote}
\small
\noindent\textbf{General Data: source analysis.}
The mask is treated as object-selection evidence rather than an exact deletion boundary. If most of a recognizable object's visible main body is selected, the complete object---including its parts, attachments, and visible extensions---is considered a removal target. Clearly attributable effects such as cast shadows, reflections, wakes, smoke, glow, or contact traces are included in the intended removal.

\medskip
\noindent\textbf{General Data: candidate assessment.}
All selected targets are assessed jointly, and a clear failure on one target cannot be compensated by successful removal of the others. Independent non-target objects are evaluated for both presence and structural integrity, including objects not explicitly listed in the source-stage contract.

\medskip
\noindent\textbf{Poster.}
Evaluation is guided by the selected layer semantics and the corresponding removed reference, not by generic inpainting expectations. Foreground content may therefore remain visible inside the selected region after a back or middle layer is removed.

\medskip
\noindent\textbf{Watermark.}
All selected repeated watermark instances belong to the removal target, while text, logos, symbols, signatures, QR codes, and graphics present in the clean reference are treated as protected content.

\medskip
\noindent\textbf{Visual Text.}
Because the weak reference may not be unique, a plausible clean reconstruction is not penalized solely for differing from it. Evaluation instead focuses on residual target text, reconstruction defects, newly introduced text, and damage to content that should remain.
\end{quote}

\subsection{DiffusionNFT Implementation}
\label{app:nft_implementation}

\paragraph{Group-relative advantages.}
For condition group $g$, let $\bar{r}_g$ and $\sigma_g$ denote the mean and standard deviation of its valid candidate rewards, and let $\sigma_{\mathcal{B}}$ denote the standard deviation over all valid rewards in the current distributed collection. We compute
\begin{equation}
  \rho_g=\min\!\left(1,\frac{\sigma_g}{0.03}\right),
  \qquad
  A_j=\operatorname{clip}\!\left(
    \rho_g\frac{r_j-\bar{r}_g}{\sigma_{\mathcal{B}}+10^{-4}},-5,5
  \right).
  \label{eq:removal_advantage}
\end{equation}
A group is used for training only if at least 12 of its 16 candidates receive valid rewards and the reward spread is nonzero. Collections with $\sigma_{\mathcal{B}}<10^{-3}$ are skipped. The factor $\rho_g$ suppresses updates from groups with insufficient within-condition reward variation.

\paragraph{DiffusionNFT objective.}
We convert the clipped advantage into $q_j=(1+A_j/A_{\max})/2$. Let $v_\theta$ denote the current policy, $v_{\mathrm{old}}$ the rollout policy, and $v_{\mathrm{ref}}$ the SFT reference. Following DiffusionNFT, we define
\begin{equation}
  v_\theta^\pm
  =v_{\mathrm{old}}\pm\beta\bigl(v_\theta-v_{\mathrm{old}}\bigr).
  \label{eq:nft_implicit_velocities}
\end{equation}
For a generated latent $\hat{z}_j$, we sample fresh Gaussian noise at each nonterminal scheduler time and form $z_{j,t}=(1-t)\hat{z}_j+t\epsilon$, followed by $\hat{z}_{j,0}^\pm=z_{j,t}-t v_\theta^\pm$.
Defining $\delta_j^\pm=\hat{z}_{j,0}^\pm-\hat{z}_j$, we use
\begin{equation}
  d_j^\pm=\operatorname{sg}\!\left[
    \max\!\left(\operatorname{mean}|\delta_j^\pm|,10^{-5}\right)
  \right],
  \qquad
  \ell_j^\pm=\frac{\operatorname{mean}[(\delta_j^\pm)^2]}{d_j^\pm},
  \label{eq:nft_adaptive_normalization}
\end{equation}
where $\operatorname{sg}$ denotes stop-gradient. The RL objective is
\begin{equation}
  \mathcal{L}_{\mathrm{RL}}
  =\mathbb{E}_{j,t,\epsilon}\!\left[
    \frac{A_{\max}}{\beta}
    \bigl(q_j\ell_j^+ +(1-q_j)\ell_j^-\bigr)
    +\lambda_{\mathrm{ref}}\,
    \operatorname{mean}\!\left[(v_\theta-v_{\mathrm{ref}})^2\right]
  \right].
  \label{eq:removal_nft}
\end{equation}
We use $\beta=0.5$, $A_{\max}=5$, and $\lambda_{\mathrm{ref}}=10^{-4}$. Gradients are propagated only through the current policy; sampled latents and the old and reference policies are treated as fixed targets. The RL reconstruction terms are averaged over the full latent and do not reuse the supervised mask weighting in Equation~\ref{eq:weighted_removal_fm}.

\paragraph{Policy updates.}
Each RL collection, corresponding to one policy update, samples conditions from a single data domain; different domains are not mixed within the same collection.
We merge the SFT adapters into the backbone and use the resulting model as the fixed RL base and SFT reference. Fresh rank-64 LoRA adapters, with scaling parameter 64 and zero dropout, are attached to the same attention modules used during SFT for the current and rollout policies. We optimize the current adapters with AdamW at a learning rate of $5\times10^{-5}$ and clip the gradient norm to 1.
We perform 100 successful policy updates in total.
After the $k$th successful update, the rollout policy is updated as
\begin{equation}
  \theta_{\mathrm{old}}
  \leftarrow
  \eta_k\theta_{\mathrm{old}}+(1-\eta_k)\theta,
  \qquad
  \eta_k=\min(0.001k,0.5).
\end{equation}
The SFT reference remains fixed throughout RL training.

\subsection{Inference and Sampling}
\label{app:inference_sampling}

All methods receive the same source images and masks and produce one output per case, with fixed per-case seeds.  LensErase uses a 20-step Euler sampler throughout, pads input dimensions to multiples of 16, and crops predictions back to their original size. All outputs are restored to the source dimensions for scoring.

\paragraph{Sampling.}
At inference, we initialize $z_1\sim\mathcal{N}(0,\mathbf{I})$ and integrate from $t=1$ to $t=0$ with deterministic Euler updates:
\begin{equation}
  z_{t+\Delta t}
  =z_t+\Delta t\,v_\theta(z_t,t,z_{\text{src}};M),
  \qquad \Delta t<0.
  \label{eq:removal_euler}
\end{equation}
Each step recomputes the velocity at the updated latent, while $z_{\text{src}}=E(I_{\text{input}})$ and $M$ remain fixed. The Euler update has no independent learned parameters. A resolution-dependent timestep shift changes the sequence of step sizes but leaves this update rule unchanged. We use 20 Euler steps throughout. The VAE decodes only the terminal latent.

\subsection{TraceFusion Parameters}
\label{app:tracefuse_parameters}

Let $M_{\mathrm{dil}}$ denote a dilation of the input selection mask. To suppress small reconstruction biases when estimating the effective edit region, we first Gaussian-smooth the signed residual between the generated output and source image and subtract its per-channel median outside $M_{\mathrm{dil}}$:
\begin{equation}
  R=G_\sigma*(\hat{I}_{\text{output}}-I_{\text{input}}),
  \qquad
  d(x)=\left\|R(x)-
  \operatorname{median}_{u\notin M_{\mathrm{dil}}}R(u)\right\|_\infty.
  \label{eq:tracefuse_residual}
\end{equation}
Here, $G_\sigma$ is a Gaussian kernel, $*$ denotes convolution, and the infinity norm takes the largest absolute response across color channels.

The detector in Section~\ref{sec:tracefuse} operates on sRGB values in the $0$--$255$ range. For an image of height $H$ and width $W$, we define the resolution scale $s=\min(H,W)/512$. The Gaussian standard deviation is $\max(0.5,5s)$ pixels with reflected boundary handling, and $M_{\mathrm{dil}}$ uses a dilation radius of $12s$ pixels. If $M_{\mathrm{dil}}$ covers the full image, the median offset in Equation~\ref{eq:tracefuse_residual} is set to zero.

We threshold $d(x)$ at 4 and 10 to obtain low- and high-confidence change regions. The low-threshold mask is opened with a disk of radius $s$, after which 8-connected components are extracted. A component is retained if it contains at least $48s^2$ pixels, including at least $8s^2$ pixels above the high threshold. Components that do not intersect $M_{\mathrm{dil}}$ additionally require at least 1\% of their pixels to exceed the high threshold. The retained components are closed with a disk of radius $3s$ and dilated by $5s$, yielding the residual support $S_{\mathrm{res}}$. The primary edit support is then
\begin{equation}
  S_0=M_{\mathrm{dil}}\cup S_{\mathrm{res}}.
\end{equation}

To recover weak changes in dark regions, we apply the same detector after a shared gamma transform,
\begin{equation}
  I_{\text{input},\gamma}
  =255\left(\frac{I_{\text{input}}}{255}\right)^{0.75},
\end{equation}
with the same transform applied to $\hat{I}_{\text{output}}$. Only changes not already covered by $S_0$ are considered. We extract 8-connected components from this supplemental support and retain a component only if it contains at least $8s^2$ pixels, intersects a dilation of $S_0$ by radius $s$, and lies predominantly in dark regions. Specifically, at least half of its pixels must have luma at most 96 in both the untransformed source and generated images, where luma is computed as $0.2126R+0.7152G+0.0722B$.
Let $S_\gamma$ denote the union of accepted supplemental components. The final fusion support is
\begin{equation}
  S=S_0\cup S_\gamma.
\end{equation}
All integer radii and area thresholds above are rounded and clamped to at least one.

Finally, we feather the boundary of $S$ using its normalized Euclidean distance transform. Let $d_S(x)$ be zero inside $S$ and the distance to $S$ outside it, with vertical and horizontal pixel spacings $1/H$ and $1/W$. The blending weight is
\begin{equation}
  \alpha(x)=\max\left(0,1-\frac{d_S(x)}{0.1}\right).
\end{equation}
If $S$ is empty, $\alpha=0$ everywhere. The source and generated images are converted from sRGB to linear RGB before blending and converted back to sRGB for the final output.

\textbf{We use the same TraceFusion parameters for all reported experiments.}

\subsection{Notation Summary}
\label{app:notation}

Table~\ref{tab:notation} summarizes the principal notation used in Section~\ref{sec:method} and Appendices~\ref{app:pretraining_design} and~\ref{app:method_details}.

\begin{table}[htbp]
  \centering
  \caption{Principal notation for LensErase.}
  \label{tab:notation}
  \begingroup
  \fontsize{9}{10.6}\selectfont
  \setlength{\tabcolsep}{5pt}
  \renewcommand{\arraystretch}{1.08}
  \begin{tabular}{@{}>{\raggedright\arraybackslash}p{0.27\linewidth}>{\raggedright\arraybackslash}p{\dimexpr0.73\linewidth-2\tabcolsep\relax}@{}}
    \toprule
    \textbf{Symbol} & \textbf{Definition} \\
    \midrule
    \multicolumn{2}{@{}l}{\textit{Images, masks, and positional coordinates}} \\
    $I_{\text{input}}$ & Source image containing the selected targets. \\
    $I_{\text{output}}$ & Paired erase target used for supervised training. \\
    $\hat{I}_{\text{output}}$, $\hat{I}_{\text{output}}^{\mathrm{TF}}$ & Generated image before and after TraceFusion, respectively. \\
    $M$, $m_i$ & Binary selection mask and its max-pooled value for source token $i$, with 1 denoting a selected region. \\
    $h_i$, $w_i$ & Height and width coordinates of token $i$. \\
    $\mathbf{p}^{\mathrm{cond}}_i$, $\mathbf{p}^{\mathrm{img}}_i$ & Source and noisy-target RoPE coordinates: $(m_i,h_i,w_i)$ and $(2,h_i,w_i)$, respectively. \\
    \midrule
    \multicolumn{2}{@{}l}{\textit{Latents and flow matching}} \\
    $E$ & VAE encoder. \\
    $z_{\text{src}}$, $z^\star$ & Source latent $E(I_{\text{input}})$ and clean target latent $E(I_{\text{output}})$. \\
    $t$, $\epsilon$, $\mathbf{I}$ & Noise level $t\in[0,1]$, Gaussian noise $\epsilon\sim\mathcal{N}(0,\mathbf{I})$, and identity covariance. \\
    $z_t$ & Noisy target latent; during supervised training, $z_t=(1-t)z^\star+t\epsilon$. \\
    $v_\theta$, $v^\star$ & Predicted velocity $v_\theta(z_t,t,z_{\text{src}};M)$ with model parameters $\theta$, and target velocity $v^\star=\epsilon-z^\star$. \\
    $N$, $i$ & Number of target tokens and token index, respectively. \\
    $\tilde{M}$ & Binary loss mask obtained by dilating $M$ and downsampling to the token grid. \\
    $\lambda$, $\omega_i$ & Masked-region loss multiplier and token weight $\omega_i=1+(\lambda-1)\tilde{M}_i$. \\
    \midrule
    \multicolumn{2}{@{}l}{\textit{Contract-guided rewards}} \\
    $\mathcal{K}=(\mathcal{T},\mathcal{E},\mathcal{O})$ & Erase contract containing removal targets, associated effects, and protected objects. \\
    $\mathcal{P}$, $P_p$, $\lambda_p$ & Reward-component set, normalized component score, and component weight. \\
    $\mathcal{A}_{j,p}$, $s_{j,p,k}$, $\widetilde{w}_{j,p,k}$ & Applicable subcriteria, normalized subscore, and renormalized subcriterion weight for candidate $j$ and component $p$. \\
    $b$, $n_0$, $r$ & Base reward, number of zero-valued Removal subcriteria, and gated reward $r=0.5^{n_0}b$. \\
    $j$ & Candidate index, suppressed in Section~\ref{sec:removal_rl} and explicit in quantities such as $P_{j,p}$, $\widetilde{\lambda}_{j,p}$, $b_j$, $n_{0,j}$, and $r_j$. \\
    \midrule
    \multicolumn{2}{@{}l}{\textit{TraceFusion}} \\
    $M_{\mathrm{dil}}$ & Dilated selection mask used by the residual detector. \\
    $R$, $d(x)$ & Smoothed signed source--output residual and its median-centered magnitude at pixel $x$, respectively. \\
    $S_{\mathrm{res}}$ & Residual support retained after component filtering and morphology. \\
    $S_0$ & Initial support: $M_{\mathrm{dil}}\cup S_{\mathrm{res}}$. \\
    $\gamma$, $S_\gamma$ & Shared power-transform exponent and the union of accepted dark-region supplemental components. \\
    $S$ & Final fusion support: $S_0\cup S_\gamma$. \\
    $d_S(x)$, $\alpha(x)$ & Distance to $S$ (zero inside), using normalized pixel spacings, and blending weight $\max(0,1-d_S(x)/0.1)$. \\
    $\mathcal{C}$, $\odot$ & sRGB-to-linear-RGB conversion and pixelwise multiplication. \\
    \bottomrule
  \end{tabular}
  \endgroup
\end{table}

\FloatBarrier

\FloatBarrier
\section{Complete Benchmark Results}
\label{app:full_benchmark_results}

\subsection{Evaluation Protocol}
\label{app:baseline_protocols}

All methods receive the same source images and masks and produce one output per case, with fixed per-case seeds. Clean targets are used only for scoring. Outputs are restored to the evaluation image dimensions.

\paragraph{CMMD (full-image resize).}
\label{app:cmmd_protocol}
For CMMD~\citep{jayasumana2024rethinking}, we bicubically resize full predictions and references to $336\times336$ without center cropping, to retain off-center removal regions and associated effects. Within each benchmark, all methods use the same samples, preprocessing, CLIP encoder, and MMD parameters.

\subsubsection{OmniPaint-Bench Mask Corrections}
\label{app:omnipaint_mask_corrections}

Among the 300 captured cases of OmniPaint-Bench~\citep{yu2025omnipaint}, we correct 32 masks whose orientation is misaligned with the corresponding input image. We rotate 29 masks $90^\circ$ counterclockwise and flip three horizontally; the other 268 masks are unchanged. These transformations preserve mask pixel values and foreground area without interpolation. All input images and ground-truth targets remain unchanged, and ground-truth targets are not used to generate or redraw the masks. All evaluated methods receive the same corrected masks.

\subsection{Baseline Configurations}
\label{app:baseline_configurations}

We use released checkpoints with the variants specified below. Code repositories and weights are linked for reproducibility.

\paragraph{ObjectClear.}
We use the released ObjectClear pipeline~\citep{zhao2026precise}\footnote{\url{https://github.com/zjx0101/ObjectClear}; \href{https://huggingface.co/jixin0101/ObjectClear/tree/c73af80888dbd519819d0a521b5c8d0f3cda6859}{released ObjectClear checkpoint}.} with attention-guided fusion.

\paragraph{SmartEraser.}
We run the released SmartEraser controller~\citep{jiang2025smarteraser}\footnote{\url{https://github.com/longtaojiang/SmartEraser}; the released \texttt{smarteraser-weights.zip} archive and CLIP ViT-L/14.}.

\paragraph{OmniPaint.}
We use OmniPaint's FLUX.1-dev removal LoRA and fixed prompt embeddings~\citep{yu2025omnipaint}\footnote{\url{https://github.com/yeates/OmniPaint}; \href{https://huggingface.co/yeates/OmniPaint/tree/76736dcbda6535146f5881ec5aa5f295a94ffd6a}{removal weights and prompt embeddings}.}.

\paragraph{OmniEraser.}
We evaluate OmniEraser-ControlNet~\citep{wei2025omnieraser}\footnote{\url{https://github.com/PRIS-CV/Omnieraser}; \href{https://huggingface.co/theSure/Omnieraser_Controlnet_version/tree/35f103690df0de9ce64bebe283c7809fe1d9bc3d}{OmniEraser-ControlNet checkpoint}.} using FLUX.1-dev, the Alimama inpainting ControlNet, and the released LoRA.

\paragraph{OSOR.}
We use OSOR-FLUX-Fill Phase~II~\citep{zhou2026osor}\footnote{\url{https://github.com/Zhouqm-Git/osor}; \href{https://huggingface.co/QinmingZhou/OSOR/tree/9288156766b1df44e5f50fb697876a6f2fd8c062}{\texttt{fluxfill\_phase2.pth} in the OSOR release}.}. Alpha-head latent blending is retained.

\paragraph{Attentive Eraser.}
We evaluate the SDXL+SIP configuration~\citep{sun2025attentive}\footnote{\url{https://github.com/Alibaba-VELLDEPTH/AttentiveEraser}, following the SDXL+SIP example.}.

\paragraph{PANDORA.}
We use the SDXL configuration of PANDORA~\citep{vo2026pandora}\footnote{\url{https://github.com/vdkhoi20/PANDORA}, with the released SDXL configuration.}.

\paragraph{DesignEdit.}
We use DesignEdit's SDXL object-removal mode~\citep{jia2025designedit}\footnote{\url{https://github.com/design-edit/DesignEdit}, using \texttt{run\_remove}.} with DDIM inversion and LayerFusion.

\paragraph{DORS.}
We use DORS with SDXL-Inpainting and SAM3~\citep{tang2026dors}\footnote{\url{https://github.com/httang1224/DORS}, with the released SDXL-Inpainting and SAM3 weights.}.

\paragraph{GeoRemover.}
We run both GeoRemover stages~\citep{zhu2026georemover}\footnote{\url{https://github.com/buxiangzhiren/GeoRemover}; \href{https://huggingface.co/buxiangzhiren/GeoRemover/tree/ef66b9ad63b9ad3a57fe7e9c39cb002b524e7ddc}{released stage-1 and stage-2 checkpoints}.} on FLUX.1-Fill-dev with Video Depth Anything Large.

\paragraph{EntityErasure.}
We use EntityErasure's full amodal segmentation and completion pipeline~\citep{zhu2025entityerasure}\footnote{\url{https://github.com/zyxunh/entity_erasure}; \href{https://huggingface.co/unhzyx/entity_erasure/tree/005659929e124bd6bb67c10bf95ddf7bbf63d7d7}{released segmentation and completion checkpoints}.}.

\paragraph{RORem-mixed.}
We use RORem's nine-channel mixed UNet~\citep{li2025rorem}\footnote{\url{https://github.com/leeruibin/RORem}; \href{https://drive.google.com/drive/folders/1G46Rs0-fZvoJ55OLQrC35dbRohFM917z}{authors' mixed-checkpoint release}.} with the released fixed prompts.

\subsection{Results on Public Benchmarks}
\label{app:complete_result_tables}

Tables~\ref{tab:results-ober-test}--\ref{tab:results-text-erase} expand Table~\ref{tab:results-public} with all twelve baselines on each benchmark, using the same evaluation protocol and eight metrics. Red, orange, and yellow cell backgrounds mark the best, second-best, and third-best values among all thirteen methods within each benchmark. Best values are also bold; ties at the displayed precision share a color.

\begingroup
\definecolor{LensAppendixFirst}{HTML}{F5C6C9} \definecolor{LensAppendixSecond}{HTML}{F9DCB9} \definecolor{LensAppendixThird}{HTML}{FBF1C3} \definecolor{LensAppendixRule}{HTML}{555B62} \definecolor{LensAppendixDivider}{HTML}{BDC3C9}

\begin{table}[!htbp]
  \centering
  \begingroup
  \caption{Complete comparison on OBER-Test~\citep{zhao2026precise}.}
  \label{tab:results-ober-test}
  \fontsize{8}{10.2}\selectfont
  \setlength{\tabcolsep}{1.7pt}
  \renewcommand{\arraystretch}{1.05}
  \arrayrulecolor{LensAppendixRule}
  \begin{tabular}{@{}>{\raggedright\arraybackslash}m{\dimexpr\linewidth-281.7pt\relax}
    *{2}{>{\columncolor{white}[0pt][0pt]\centering\arraybackslash}m{31.5pt}}
    >{\columncolor{white}[0pt][0pt]\centering\arraybackslash}m{34pt}
    *{5}{>{\columncolor{white}[0pt][0pt]\centering\arraybackslash}m{31.5pt}}@{}}
    \toprule[0.6pt]
    \textbf{Method} & \mbox{\textbf{PSNR}\,$\uparrow$} & \mbox{\textbf{FID}\,$\downarrow$}
    & \mbox{\textbf{CMMD}\,$\downarrow$} & \mbox{\textbf{LPIPS}\,$\downarrow$}
    & \mbox{\textbf{SSIM}\,$\uparrow$} & \mbox{\textbf{DISTS}\,$\downarrow$}
    & \mbox{\textbf{CLIP-I}\,$\uparrow$} & \mbox{\textbf{DINO}\,$\uparrow$} \\
    \midrule[0.35pt]
    \mbox{SmartEraser~\citep{jiang2025smarteraser}} & 26.67 & 94.58 & 0.4178 & 0.0762 & \cellcolor{LensAppendixSecond}0.9446 & 0.0792 & 0.9393 & 0.8423 \\
    \addlinespace[0.9pt]
    \mbox{RORem-mixed~\citep{li2025rorem}} & 25.93 & 61.32 & 0.1832 & 0.1122 & 0.8549 & 0.0836 & 0.9593 & 0.9186 \\
    \addlinespace[0.9pt]
    \mbox{EntityErasure~\citep{zhu2025entityerasure}} & 22.75 & 82.02 & 0.2933 & 0.1330 & 0.7263 & 0.1037 & 0.9376 & 0.8866 \\
    \addlinespace[0.9pt]
    \mbox{Attentive Eraser~\citep{sun2025attentive}} & 26.22 & 59.96 & 0.2048 & 0.1123 & 0.8497 & 0.0889 & 0.9605 & 0.9228 \\
    \addlinespace[0.9pt]
    \mbox{DesignEdit~\citep{jia2025designedit}} & 26.39 & 71.25 & 0.2052 & 0.1302 & 0.8585 & 0.0955 & 0.9485 & 0.8997 \\
    \addlinespace[0.9pt]
    \mbox{OmniEraser~\citep{wei2025omnieraser}} & 25.52 & 85.73 & 0.2131 & 0.0996 & 0.9141 & 0.0896 & 0.9457 & 0.8698 \\
    \addlinespace[0.9pt]
    \mbox{OmniPaint~\citep{yu2025omnipaint}} & 28.79 & \cellcolor{LensAppendixThird}23.68 & 0.0829 & 0.0549 & 0.8575 & 0.0427 & \cellcolor{LensAppendixThird}0.9836 & \cellcolor{LensAppendixSecond}0.9803 \\
    \addlinespace[0.9pt]
    \mbox{GeoRemover~\citep{zhu2026georemover}} & 24.66 & 56.24 & 0.4277 & 0.1423 & 0.7622 & 0.1128 & 0.9305 & 0.9292 \\
    \addlinespace[0.9pt]
    \mbox{PANDORA~\citep{vo2026pandora}} & 25.63 & 79.21 & 0.3854 & 0.1482 & 0.8507 & 0.1056 & 0.9397 & 0.8723 \\
    \addlinespace[0.9pt]
    \mbox{DORS~\citep{tang2026dors}} & 24.13 & 88.09 & 0.4874 & 0.1250 & 0.7492 & 0.0966 & 0.9371 & 0.8714 \\
    \addlinespace[0.9pt]
    \mbox{ObjectClear~\citep{zhao2026precise}} & \cellcolor{LensAppendixSecond}33.05 & \cellcolor{LensAppendixSecond}19.16 & \cellcolor{LensAppendixSecond}0.0132 & \cellcolor{LensAppendixSecond}0.0338 & \cellcolor{LensAppendixThird}0.9353 & \cellcolor{LensAppendixSecond}0.0217 & \cellcolor{LensAppendixSecond}0.9894 & \cellcolor{LensAppendixFirst}\textbf{0.9839} \\
    \addlinespace[0.9pt]
    \mbox{OSOR~\citep{zhou2026osor}} & \cellcolor{LensAppendixThird}30.14 & 27.82 & \cellcolor{LensAppendixThird}0.0674 & \cellcolor{LensAppendixThird}0.0524 & 0.8748 & \cellcolor{LensAppendixThird}0.0377 & 0.9827 & \cellcolor{LensAppendixThird}0.9693 \\
    \arrayrulecolor{LensAppendixDivider}
    \specialrule{0.25pt}{3.1pt}{3.6pt}
    \arrayrulecolor{LensAppendixRule}
    \textbf{LensErase} & \cellcolor{LensAppendixFirst}\textbf{34.96} & \cellcolor{LensAppendixFirst}\textbf{16.97} & \cellcolor{LensAppendixFirst}\textbf{0.0073} & \cellcolor{LensAppendixFirst}\textbf{0.0256} & \cellcolor{LensAppendixFirst}\textbf{0.9543} & \cellcolor{LensAppendixFirst}\textbf{0.0176} & \cellcolor{LensAppendixFirst}\textbf{0.9930} & \cellcolor{LensAppendixFirst}\textbf{0.9839} \\
    \addlinespace[1.8pt]
    \bottomrule[0.6pt]
  \end{tabular}
  \arrayrulecolor{black}
  \endgroup
\end{table}

\begin{table}[!htbp]
  \centering
  \begingroup
  \caption{Complete comparison on RORD-Val-343~\citep{sagong2022rord}.}
  \label{tab:results-rord-val-343}
  \fontsize{8}{10.2}\selectfont
  \setlength{\tabcolsep}{1.7pt}
  \renewcommand{\arraystretch}{1.05}
  \arrayrulecolor{LensAppendixRule}
  \begin{tabular}{@{}>{\raggedright\arraybackslash}m{\dimexpr\linewidth-281.7pt\relax}
    *{2}{>{\columncolor{white}[0pt][0pt]\centering\arraybackslash}m{31.5pt}}
    >{\columncolor{white}[0pt][0pt]\centering\arraybackslash}m{34pt}
    *{5}{>{\columncolor{white}[0pt][0pt]\centering\arraybackslash}m{31.5pt}}@{}}
    \toprule[0.6pt]
    \textbf{Method} & \mbox{\textbf{PSNR}\,$\uparrow$} & \mbox{\textbf{FID}\,$\downarrow$}
    & \mbox{\textbf{CMMD}\,$\downarrow$} & \mbox{\textbf{LPIPS}\,$\downarrow$}
    & \mbox{\textbf{SSIM}\,$\uparrow$} & \mbox{\textbf{DISTS}\,$\downarrow$}
    & \mbox{\textbf{CLIP-I}\,$\uparrow$} & \mbox{\textbf{DINO}\,$\uparrow$} \\
    \midrule[0.35pt]
    \mbox{SmartEraser~\citep{jiang2025smarteraser}} & 24.89 & 35.41 & 0.1104 & 0.1658 & \cellcolor{LensAppendixThird}0.8472 & 0.0950 & 0.9644 & 0.9349 \\
    \addlinespace[0.9pt]
    \mbox{RORem-mixed~\citep{li2025rorem}} & 23.16 & 35.91 & 0.0983 & 0.1906 & 0.6893 & 0.0990 & 0.9482 & 0.9316 \\
    \addlinespace[0.9pt]
    \mbox{EntityErasure~\citep{zhu2025entityerasure}} & 21.25 & 50.71 & 0.2829 & 0.3340 & 0.6142 & 0.1461 & 0.9400 & 0.9082 \\
    \addlinespace[0.9pt]
    \mbox{Attentive Eraser~\citep{sun2025attentive}} & 22.57 & 39.19 & 0.0948 & 0.1743 & 0.6896 & 0.0879 & 0.9501 & 0.9189 \\
    \addlinespace[0.9pt]
    \mbox{DesignEdit~\citep{jia2025designedit}} & 23.28 & 49.55 & 0.1603 & 0.1938 & 0.7268 & 0.1046 & 0.9304 & 0.8837 \\
    \addlinespace[0.9pt]
    \mbox{OmniEraser~\citep{wei2025omnieraser}} & 24.46 & 35.78 & 0.0426 & 0.1189 & 0.8380 & 0.0690 & 0.9580 & 0.9262 \\
    \addlinespace[0.9pt]
    \mbox{OmniPaint~\citep{yu2025omnipaint}} & 23.64 & 27.45 & \cellcolor{LensAppendixSecond}0.0334 & \cellcolor{LensAppendixThird}0.1119 & 0.7823 & \cellcolor{LensAppendixThird}0.0623 & \cellcolor{LensAppendixThird}0.9672 & \cellcolor{LensAppendixThird}0.9541 \\
    \addlinespace[0.9pt]
    \mbox{GeoRemover~\citep{zhu2026georemover}} & 24.62 & 28.13 & 0.1716 & 0.1185 & 0.7659 & 0.0708 & 0.9640 & 0.9536 \\
    \addlinespace[0.9pt]
    \mbox{PANDORA~\citep{vo2026pandora}} & 21.61 & 68.85 & 0.5348 & 0.2274 & 0.7106 & 0.1211 & 0.8886 & 0.7928 \\
    \addlinespace[0.9pt]
    \mbox{DORS~\citep{tang2026dors}} & 21.84 & 50.22 & 0.3032 & 0.2902 & 0.6253 & 0.1412 & 0.9330 & 0.8862 \\
    \addlinespace[0.9pt]
    \mbox{ObjectClear~\citep{zhao2026precise}} & \cellcolor{LensAppendixSecond}26.21 & \cellcolor{LensAppendixSecond}25.42 & 0.0511 & 0.1163 & \cellcolor{LensAppendixSecond}0.8577 & 0.0657 & \cellcolor{LensAppendixSecond}0.9704 & \cellcolor{LensAppendixSecond}0.9566 \\
    \addlinespace[0.9pt]
    \mbox{OSOR~\citep{zhou2026osor}} & \cellcolor{LensAppendixThird}25.67 & \cellcolor{LensAppendixThird}26.33 & \cellcolor{LensAppendixThird}0.0390 & \cellcolor{LensAppendixSecond}0.1037 & 0.8138 & \cellcolor{LensAppendixSecond}0.0542 & 0.9663 & 0.9475 \\
    \arrayrulecolor{LensAppendixDivider}
    \specialrule{0.25pt}{3.1pt}{3.6pt}
    \arrayrulecolor{LensAppendixRule}
    \textbf{LensErase} & \cellcolor{LensAppendixFirst}\textbf{28.23} & \cellcolor{LensAppendixFirst}\textbf{20.53} & \cellcolor{LensAppendixFirst}\textbf{0.0272} & \cellcolor{LensAppendixFirst}\textbf{0.0601} & \cellcolor{LensAppendixFirst}\textbf{0.9053} & \cellcolor{LensAppendixFirst}\textbf{0.0357} & \cellcolor{LensAppendixFirst}\textbf{0.9715} & \cellcolor{LensAppendixFirst}\textbf{0.9568} \\
    \addlinespace[1.8pt]
    \bottomrule[0.6pt]
  \end{tabular}
  \arrayrulecolor{black}
  \endgroup
\end{table}

\begin{table}[!htbp]
  \centering
  \begingroup
  \caption{Complete comparison on RemovalBench~\citep{wei2025omnieraser}.}
  \label{tab:results-removalbench}
  \fontsize{8}{10.2}\selectfont
  \setlength{\tabcolsep}{1.7pt}
  \renewcommand{\arraystretch}{1.05}
  \arrayrulecolor{LensAppendixRule}
  \begin{tabular}{@{}>{\raggedright\arraybackslash}m{\dimexpr\linewidth-281.7pt\relax}
    *{2}{>{\columncolor{white}[0pt][0pt]\centering\arraybackslash}m{31.5pt}}
    >{\columncolor{white}[0pt][0pt]\centering\arraybackslash}m{34pt}
    *{5}{>{\columncolor{white}[0pt][0pt]\centering\arraybackslash}m{31.5pt}}@{}}
    \toprule[0.6pt]
    \textbf{Method} & \mbox{\textbf{PSNR}\,$\uparrow$} & \mbox{\textbf{FID}\,$\downarrow$}
    & \mbox{\textbf{CMMD}\,$\downarrow$} & \mbox{\textbf{LPIPS}\,$\downarrow$}
    & \mbox{\textbf{SSIM}\,$\uparrow$} & \mbox{\textbf{DISTS}\,$\downarrow$}
    & \mbox{\textbf{CLIP-I}\,$\uparrow$} & \mbox{\textbf{DINO}\,$\uparrow$} \\
    \midrule[0.35pt]
    \mbox{SmartEraser~\citep{jiang2025smarteraser}} & 23.49 & 151.88 & 0.4359 & 0.3308 & \cellcolor{LensAppendixThird}0.7695 & 0.1890 & 0.9339 & 0.8239 \\
    \addlinespace[0.9pt]
    \mbox{RORem-mixed~\citep{li2025rorem}} & 23.51 & 90.95 & 0.2044 & 0.2262 & 0.7149 & 0.1175 & 0.9575 & 0.9098 \\
    \addlinespace[0.9pt]
    \mbox{EntityErasure~\citep{zhu2025entityerasure}} & 20.36 & 161.80 & 0.3525 & 0.4287 & 0.6664 & 0.2288 & 0.9116 & 0.7918 \\
    \addlinespace[0.9pt]
    \mbox{Attentive Eraser~\citep{sun2025attentive}} & 23.78 & 80.49 & 0.1666 & \cellcolor{LensAppendixThird}0.1941 & 0.7095 & \cellcolor{LensAppendixThird}0.0969 & 0.9646 & 0.9263 \\
    \addlinespace[0.9pt]
    \mbox{DesignEdit~\citep{jia2025designedit}} & 23.96 & 108.46 & 0.2758 & 0.2343 & 0.7200 & 0.1357 & 0.9457 & 0.8676 \\
    \addlinespace[0.9pt]
    \mbox{OmniEraser~\citep{wei2025omnieraser}} & 22.33 & 147.90 & 0.3822 & 0.2150 & 0.7539 & 0.1285 & 0.9213 & 0.7990 \\
    \addlinespace[0.9pt]
    \mbox{OmniPaint~\citep{yu2025omnipaint}} & 24.57 & \cellcolor{LensAppendixThird}43.25 & \cellcolor{LensAppendixThird}0.0265 & 0.2103 & 0.7535 & 0.0998 & 0.9821 & 0.9663 \\
    \addlinespace[0.9pt]
    \mbox{GeoRemover~\citep{zhu2026georemover}} & 25.35 & \cellcolor{LensAppendixSecond}42.05 & 0.2310 & 0.2316 & 0.7457 & 0.1181 & 0.9777 & \cellcolor{LensAppendixThird}0.9698 \\
    \addlinespace[0.9pt]
    \mbox{PANDORA~\citep{vo2026pandora}} & 23.84 & 98.21 & 0.4140 & 0.2389 & 0.7167 & 0.1413 & 0.9359 & 0.8427 \\
    \addlinespace[0.9pt]
    \mbox{DORS~\citep{tang2026dors}} & 23.07 & 141.61 & 0.4622 & 0.3755 & 0.6987 & 0.1935 & 0.9419 & 0.8629 \\
    \addlinespace[0.9pt]
    \mbox{ObjectClear~\citep{zhao2026precise}} & \cellcolor{LensAppendixSecond}25.99 & 46.61 & \cellcolor{LensAppendixSecond}0.0248 & 0.2853 & \cellcolor{LensAppendixSecond}0.7794 & 0.1428 & \cellcolor{LensAppendixSecond}0.9853 & \cellcolor{LensAppendixSecond}0.9718 \\
    \addlinespace[0.9pt]
    \mbox{OSOR~\citep{zhou2026osor}} & \cellcolor{LensAppendixThird}25.80 & 43.89 & 0.0275 & \cellcolor{LensAppendixSecond}0.1935 & 0.7656 & \cellcolor{LensAppendixSecond}0.0882 & \cellcolor{LensAppendixThird}0.9838 & 0.9597 \\
    \arrayrulecolor{LensAppendixDivider}
    \specialrule{0.25pt}{3.1pt}{3.6pt}
    \arrayrulecolor{LensAppendixRule}
    \textbf{LensErase} & \cellcolor{LensAppendixFirst}\textbf{28.16} & \cellcolor{LensAppendixFirst}\textbf{33.67} & \cellcolor{LensAppendixFirst}\textbf{0.0116} & \cellcolor{LensAppendixFirst}\textbf{0.1125} & \cellcolor{LensAppendixFirst}\textbf{0.8306} & \cellcolor{LensAppendixFirst}\textbf{0.0586} & \cellcolor{LensAppendixFirst}\textbf{0.9874} & \cellcolor{LensAppendixFirst}\textbf{0.9737} \\
    \addlinespace[1.8pt]
    \bottomrule[0.6pt]
  \end{tabular}
  \arrayrulecolor{black}
  \endgroup
\end{table}

\begin{table}[!htbp]
  \centering
  \begingroup
  \caption{Complete comparison on OmniPaint-Bench~\citep{yu2025omnipaint} (captured subset), excluding RORD-derived cases.}
  \label{tab:results-omnipaint-300}
  \fontsize{8}{10.2}\selectfont
  \setlength{\tabcolsep}{1.7pt}
  \renewcommand{\arraystretch}{1.05}
  \arrayrulecolor{LensAppendixRule}
  \begin{tabular}{@{}>{\raggedright\arraybackslash}m{\dimexpr\linewidth-281.7pt\relax}
    *{2}{>{\columncolor{white}[0pt][0pt]\centering\arraybackslash}m{31.5pt}}
    >{\columncolor{white}[0pt][0pt]\centering\arraybackslash}m{34pt}
    *{5}{>{\columncolor{white}[0pt][0pt]\centering\arraybackslash}m{31.5pt}}@{}}
    \toprule[0.6pt]
    \textbf{Method} & \mbox{\textbf{PSNR}\,$\uparrow$} & \mbox{\textbf{FID}\,$\downarrow$}
    & \mbox{\textbf{CMMD}\,$\downarrow$} & \mbox{\textbf{LPIPS}\,$\downarrow$}
    & \mbox{\textbf{SSIM}\,$\uparrow$} & \mbox{\textbf{DISTS}\,$\downarrow$}
    & \mbox{\textbf{CLIP-I}\,$\uparrow$} & \mbox{\textbf{DINO}\,$\uparrow$} \\
    \midrule[0.35pt]
    \mbox{SmartEraser~\citep{jiang2025smarteraser}} & 22.05 & 94.79 & 0.2854 & 0.2787 & 0.8121 & 0.1460 & 0.9189 & 0.8610 \\
    \addlinespace[0.9pt]
    \mbox{RORem-mixed~\citep{li2025rorem}} & 23.14 & 54.66 & 0.1398 & 0.1994 & 0.7730 & 0.1072 & 0.9433 & 0.9226 \\
    \addlinespace[0.9pt]
    \mbox{EntityErasure~\citep{zhu2025entityerasure}} & 20.44 & 96.41 & 0.3472 & 0.3033 & 0.7262 & 0.1583 & 0.9068 & 0.8579 \\
    \addlinespace[0.9pt]
    \mbox{Attentive Eraser~\citep{sun2025attentive}} & 23.31 & 50.60 & 0.1324 & 0.1642 & 0.7865 & 0.0719 & 0.9452 & 0.9322 \\
    \addlinespace[0.9pt]
    \mbox{DesignEdit~\citep{jia2025designedit}} & 23.30 & 68.54 & 0.2245 & 0.1838 & 0.7859 & 0.1033 & 0.9261 & 0.8838 \\
    \addlinespace[0.9pt]
    \mbox{OmniEraser~\citep{wei2025omnieraser}} & 21.73 & 93.71 & 0.2645 & 0.1767 & 0.8123 & 0.0904 & 0.9059 & 0.8212 \\
    \addlinespace[0.9pt]
    \mbox{OmniPaint~\citep{yu2025omnipaint}} & 23.42 & 40.65 & \cellcolor{LensAppendixThird}0.0743 & \cellcolor{LensAppendixThird}0.1484 & \cellcolor{LensAppendixThird}0.8193 & \cellcolor{LensAppendixThird}0.0644 & 0.9562 & 0.9423 \\
    \addlinespace[0.9pt]
    \mbox{GeoRemover~\citep{zhu2026georemover}} & 24.87 & \cellcolor{LensAppendixThird}38.54 & 0.1376 & 0.1575 & 0.8173 & 0.0704 & 0.9524 & 0.9494 \\
    \addlinespace[0.9pt]
    \mbox{PANDORA~\citep{vo2026pandora}} & 23.26 & 83.41 & 0.3606 & 0.1940 & 0.7835 & 0.1122 & 0.9130 & 0.8538 \\
    \addlinespace[0.9pt]
    \mbox{DORS~\citep{tang2026dors}} & 22.20 & 79.22 & 0.3511 & 0.2799 & 0.7479 & 0.1389 & 0.9246 & 0.8949 \\
    \addlinespace[0.9pt]
    \mbox{ObjectClear~\citep{zhao2026precise}} & \cellcolor{LensAppendixThird}24.93 & 48.91 & 0.0981 & 0.2233 & 0.8178 & 0.1054 & \cellcolor{LensAppendixThird}0.9613 & \cellcolor{LensAppendixSecond}0.9549 \\
    \addlinespace[0.9pt]
    \mbox{OSOR~\citep{zhou2026osor}} & \cellcolor{LensAppendixSecond}26.02 & \cellcolor{LensAppendixSecond}37.52 & \cellcolor{LensAppendixSecond}0.0687 & \cellcolor{LensAppendixSecond}0.1336 & \cellcolor{LensAppendixSecond}0.8421 & \cellcolor{LensAppendixSecond}0.0566 & \cellcolor{LensAppendixSecond}0.9617 & \cellcolor{LensAppendixThird}0.9506 \\
    \arrayrulecolor{LensAppendixDivider}
    \specialrule{0.25pt}{3.1pt}{3.6pt}
    \arrayrulecolor{LensAppendixRule}
    \textbf{LensErase} & \cellcolor{LensAppendixFirst}\textbf{26.22} & \cellcolor{LensAppendixFirst}\textbf{33.29} & \cellcolor{LensAppendixFirst}\textbf{0.0245} & \cellcolor{LensAppendixFirst}\textbf{0.1257} & \cellcolor{LensAppendixFirst}\textbf{0.8521} & \cellcolor{LensAppendixFirst}\textbf{0.0544} & \cellcolor{LensAppendixFirst}\textbf{0.9660} & \cellcolor{LensAppendixFirst}\textbf{0.9595} \\
    \addlinespace[1.8pt]
    \bottomrule[0.6pt]
  \end{tabular}
  \arrayrulecolor{black}
  \endgroup
\end{table}

\begin{table}[!htbp]
  \centering
  \begingroup
  \caption{Complete comparison on AnimeEraseBench~\citep{zhou2026osor}.}
  \label{tab:results-anime-erase}
  \fontsize{8}{10.2}\selectfont
  \setlength{\tabcolsep}{1.7pt}
  \renewcommand{\arraystretch}{1.05}
  \arrayrulecolor{LensAppendixRule}
  \begin{tabular}{@{}>{\raggedright\arraybackslash}m{\dimexpr\linewidth-281.7pt\relax}
    *{2}{>{\columncolor{white}[0pt][0pt]\centering\arraybackslash}m{31.5pt}}
    >{\columncolor{white}[0pt][0pt]\centering\arraybackslash}m{34pt}
    *{5}{>{\columncolor{white}[0pt][0pt]\centering\arraybackslash}m{31.5pt}}@{}}
    \toprule[0.6pt]
    \textbf{Method} & \mbox{\textbf{PSNR}\,$\uparrow$} & \mbox{\textbf{FID}\,$\downarrow$}
    & \mbox{\textbf{CMMD}\,$\downarrow$} & \mbox{\textbf{LPIPS}\,$\downarrow$}
    & \mbox{\textbf{SSIM}\,$\uparrow$} & \mbox{\textbf{DISTS}\,$\downarrow$}
    & \mbox{\textbf{CLIP-I}\,$\uparrow$} & \mbox{\textbf{DINO}\,$\uparrow$} \\
    \midrule[0.35pt]
    \mbox{SmartEraser~\citep{jiang2025smarteraser}} & 23.90 & 55.01 & 0.0451 & 0.2116 & 0.8252 & 0.1174 & 0.9837 & 0.9577 \\
    \addlinespace[0.9pt]
    \mbox{RORem-mixed~\citep{li2025rorem}} & 24.42 & 33.68 & 0.0261 & 0.1305 & 0.7926 & 0.0654 & 0.9875 & 0.9756 \\
    \addlinespace[0.9pt]
    \mbox{EntityErasure~\citep{zhu2025entityerasure}} & 20.26 & 83.39 & 0.0906 & 0.3574 & 0.6143 & 0.1776 & 0.9711 & 0.9282 \\
    \addlinespace[0.9pt]
    \mbox{Attentive Eraser~\citep{sun2025attentive}} & 22.60 & 37.77 & 0.0315 & 0.1329 & 0.7321 & 0.0629 & 0.9852 & 0.9699 \\
    \addlinespace[0.9pt]
    \mbox{DesignEdit~\citep{jia2025designedit}} & 24.10 & 49.33 & 0.0666 & 0.1582 & 0.7861 & 0.0846 & 0.9800 & 0.9519 \\
    \addlinespace[0.9pt]
    \mbox{OmniEraser~\citep{wei2025omnieraser}} & 24.05 & 49.39 & 0.0383 & \cellcolor{LensAppendixThird}0.1155 & 0.8569 & \cellcolor{LensAppendixThird}0.0617 & 0.9772 & 0.9273 \\
    \addlinespace[0.9pt]
    \mbox{OmniPaint~\citep{yu2025omnipaint}} & \cellcolor{LensAppendixThird}25.58 & \cellcolor{LensAppendixThird}27.76 & \cellcolor{LensAppendixThird}0.0138 & 0.1195 & \cellcolor{LensAppendixThird}0.8581 & 0.0636 & 0.9874 & 0.9764 \\
    \addlinespace[0.9pt]
    \mbox{GeoRemover~\citep{zhu2026georemover}} & 24.80 & 30.72 & 0.0533 & 0.1361 & 0.8247 & 0.0734 & 0.9828 & \cellcolor{LensAppendixThird}0.9766 \\
    \addlinespace[0.9pt]
    \mbox{PANDORA~\citep{vo2026pandora}} & 23.38 & 54.81 & 0.0968 & 0.1621 & 0.7787 & 0.0911 & 0.9716 & 0.9274 \\
    \addlinespace[0.9pt]
    \mbox{DORS~\citep{tang2026dors}} & 21.59 & 73.28 & 0.0925 & 0.3123 & 0.6459 & 0.1548 & 0.9742 & 0.9460 \\
    \addlinespace[0.9pt]
    \mbox{ObjectClear~\citep{zhao2026precise}} & 25.42 & 36.76 & \cellcolor{LensAppendixSecond}0.0121 & 0.1791 & 0.8417 & 0.0978 & \cellcolor{LensAppendixSecond}0.9914 & \cellcolor{LensAppendixFirst}\textbf{0.9821} \\
    \addlinespace[0.9pt]
    \mbox{OSOR~\citep{zhou2026osor}} & \cellcolor{LensAppendixSecond}27.96 & \cellcolor{LensAppendixSecond}24.45 & \cellcolor{LensAppendixFirst}\textbf{0.0094} & \cellcolor{LensAppendixSecond}0.0967 & \cellcolor{LensAppendixSecond}0.8870 & \cellcolor{LensAppendixSecond}0.0475 & \cellcolor{LensAppendixThird}0.9900 & 0.9751 \\
    \arrayrulecolor{LensAppendixDivider}
    \specialrule{0.25pt}{3.1pt}{3.6pt}
    \arrayrulecolor{LensAppendixRule}
    \textbf{LensErase} & \cellcolor{LensAppendixFirst}\textbf{30.50} & \cellcolor{LensAppendixFirst}\textbf{17.20} & \cellcolor{LensAppendixFirst}\textbf{0.0094} & \cellcolor{LensAppendixFirst}\textbf{0.0527} & \cellcolor{LensAppendixFirst}\textbf{0.9398} & \cellcolor{LensAppendixFirst}\textbf{0.0159} & \cellcolor{LensAppendixFirst}\textbf{0.9920} & \cellcolor{LensAppendixSecond}0.9798 \\
    \addlinespace[1.8pt]
    \bottomrule[0.6pt]
  \end{tabular}
  \arrayrulecolor{black}
  \endgroup
\end{table}

\begin{table}[!htbp]
  \centering
  \begingroup
  \caption{Complete comparison on TextEraseBench~\citep{zhou2026osor}.}
  \label{tab:results-text-erase}
  \fontsize{8}{10.2}\selectfont
  \setlength{\tabcolsep}{1.7pt}
  \renewcommand{\arraystretch}{1.05}
  \arrayrulecolor{LensAppendixRule}
  \begin{tabular}{@{}>{\raggedright\arraybackslash}m{\dimexpr\linewidth-281.7pt\relax}
    *{2}{>{\columncolor{white}[0pt][0pt]\centering\arraybackslash}m{31.5pt}}
    >{\columncolor{white}[0pt][0pt]\centering\arraybackslash}m{34pt}
    *{5}{>{\columncolor{white}[0pt][0pt]\centering\arraybackslash}m{31.5pt}}@{}}
    \toprule[0.6pt]
    \textbf{Method} & \mbox{\textbf{PSNR}\,$\uparrow$} & \mbox{\textbf{FID}\,$\downarrow$}
    & \mbox{\textbf{CMMD}\,$\downarrow$} & \mbox{\textbf{LPIPS}\,$\downarrow$}
    & \mbox{\textbf{SSIM}\,$\uparrow$} & \mbox{\textbf{DISTS}\,$\downarrow$}
    & \mbox{\textbf{CLIP-I}\,$\uparrow$} & \mbox{\textbf{DINO}\,$\uparrow$} \\
    \midrule[0.35pt]
    \mbox{SmartEraser~\citep{jiang2025smarteraser}} & 28.85 & 47.96 & 0.0608 & 0.1686 & 0.8885 & 0.1049 & 0.9430 & 0.9471 \\
    \addlinespace[0.9pt]
    \mbox{RORem-mixed~\citep{li2025rorem}} & 29.50 & 24.83 & \cellcolor{LensAppendixThird}0.0178 & 0.1034 & 0.8292 & 0.0584 & \cellcolor{LensAppendixThird}0.9815 & 0.9744 \\
    \addlinespace[0.9pt]
    \mbox{EntityErasure~\citep{zhu2025entityerasure}} & 24.98 & 53.79 & 0.1175 & 0.2653 & 0.7307 & 0.1502 & 0.9427 & 0.9414 \\
    \addlinespace[0.9pt]
    \mbox{Attentive Eraser~\citep{sun2025attentive}} & 26.31 & 34.70 & 0.0228 & 0.1134 & 0.7956 & 0.0563 & 0.9738 & 0.9613 \\
    \addlinespace[0.9pt]
    \mbox{DesignEdit~\citep{jia2025designedit}} & 27.61 & 49.23 & 0.1002 & 0.1391 & 0.8289 & 0.0883 & 0.9578 & 0.9247 \\
    \addlinespace[0.9pt]
    \mbox{OmniEraser~\citep{wei2025omnieraser}} & \cellcolor{LensAppendixThird}30.70 & \cellcolor{LensAppendixThird}21.97 & 0.0195 & \cellcolor{LensAppendixThird}0.0763 & \cellcolor{LensAppendixThird}0.9005 & \cellcolor{LensAppendixThird}0.0435 & 0.9812 & \cellcolor{LensAppendixThird}0.9758 \\
    \addlinespace[0.9pt]
    \mbox{OmniPaint~\citep{yu2025omnipaint}} & 28.14 & 27.78 & 0.0188 & 0.1085 & 0.8798 & 0.0607 & 0.9772 & 0.9622 \\
    \addlinespace[0.9pt]
    \mbox{GeoRemover~\citep{zhu2026georemover}} & 26.26 & 60.36 & 0.1997 & 0.1346 & 0.8613 & 0.0804 & 0.8907 & 0.9169 \\
    \addlinespace[0.9pt]
    \mbox{PANDORA~\citep{vo2026pandora}} & 26.39 & 53.93 & 0.1185 & 0.1412 & 0.8202 & 0.0914 & 0.9463 & 0.9106 \\
    \addlinespace[0.9pt]
    \mbox{DORS~\citep{tang2026dors}} & 25.75 & 53.04 & 0.0676 & 0.2467 & 0.7484 & 0.1332 & 0.9612 & 0.9343 \\
    \addlinespace[0.9pt]
    \mbox{ObjectClear~\citep{zhao2026precise}} & 30.37 & 27.81 & 0.0193 & 0.1329 & 0.8976 & 0.0822 & 0.9774 & 0.9749 \\
    \addlinespace[0.9pt]
    \mbox{OSOR~\citep{zhou2026osor}} & \cellcolor{LensAppendixSecond}32.04 & \cellcolor{LensAppendixSecond}18.52 & \cellcolor{LensAppendixSecond}0.0093 & \cellcolor{LensAppendixSecond}0.0728 & \cellcolor{LensAppendixSecond}0.9126 & \cellcolor{LensAppendixSecond}0.0412 & \cellcolor{LensAppendixSecond}0.9843 & \cellcolor{LensAppendixSecond}0.9771 \\
    \arrayrulecolor{LensAppendixDivider}
    \specialrule{0.25pt}{3.1pt}{3.6pt}
    \arrayrulecolor{LensAppendixRule}
    \textbf{LensErase} & \cellcolor{LensAppendixFirst}\textbf{36.88} & \cellcolor{LensAppendixFirst}\textbf{13.51} & \cellcolor{LensAppendixFirst}\textbf{0.0091} & \cellcolor{LensAppendixFirst}\textbf{0.0339} & \cellcolor{LensAppendixFirst}\textbf{0.9654} & \cellcolor{LensAppendixFirst}\textbf{0.0135} & \cellcolor{LensAppendixFirst}\textbf{0.9906} & \cellcolor{LensAppendixFirst}\textbf{0.9822} \\
    \addlinespace[1.8pt]
    \bottomrule[0.6pt]
  \end{tabular}
  \arrayrulecolor{black}
  \endgroup
\end{table}
\FloatBarrier
\endgroup

\subsection{Results on LensErase-Bench}
\label{app:full_vlm_results}

Table~\ref{tab:vlm-full} reports mean scores over LensErase-Bench. PANDORA uses 499 cases because one output was consistently rejected by GPT-5.6 Sol's safety filter; all other methods use 500.
HiFi Removal denotes the percentage of cases achieving complete removal ($\mathrm{Removal}=4$) while maintaining Inpainting, Preservation, and
Cleanliness scores of at least 3.

\begingroup
\definecolor{LensFullVLMRule}{HTML}{555B62} \definecolor{LensFullVLMDivider}{HTML}{BDC3C9}
\definecolor{LensFullVLMFirst}{HTML}{F5C6C9}
\definecolor{LensFullVLMSecond}{HTML}{F9DCB9}
\definecolor{LensFullVLMThird}{HTML}{FBF1C3}

\begin{table}[!htbp]
  \centering
  \begingroup
  \caption{VLM comparison on LensErase-Bench.}
  \label{tab:vlm-full}
  \fontsize{8}{10.2}\selectfont
  \setlength{\tabcolsep}{1.5pt}
  \renewcommand{\arraystretch}{1.05}
  \arrayrulecolor{LensFullVLMRule}
  \begin{tabular*}{\linewidth}{@{\extracolsep{\fill}}l*{6}{>{\columncolor{white}[0pt][0pt]}c}@{}}
    \toprule[0.6pt]
    \textbf{Method} & \mbox{\textbf{Removal}\,$\uparrow$} & \mbox{\textbf{Inpainting}\,$\uparrow$}
    & \mbox{\textbf{Preservation}\,$\uparrow$} & \mbox{\textbf{Cleanliness}\,$\uparrow$}
    & \mbox{\textbf{Overall}\,$\uparrow$} & \mbox{\textbf{HiFi Removal (\%)}\,$\uparrow$} \\
    \midrule[0.35pt]
    \mbox{SmartEraser~\citep{jiang2025smarteraser}} & 23.40 & 24.15 & 66.25 & 57.75 & 38.99 & 3.60 \\
    \addlinespace[0.9pt]
    \mbox{RORem-mixed~\citep{li2025rorem}} & 73.95 & 59.70 & 74.05 & 67.10 & 69.75 & 27.40 \\
    \addlinespace[0.9pt]
    \mbox{EntityErasure~\citep{zhu2025entityerasure}} & 59.65 & 48.50 & 48.45 & 60.20 & 55.29 & 4.80 \\
    \addlinespace[0.9pt]
    \mbox{Attentive Eraser~\citep{sun2025attentive}} & \cellcolor{LensFullVLMSecond}78.40 & 52.25 & 61.50 & 55.05 & 65.12 & 20.60 \\
    \addlinespace[0.9pt]
    \mbox{DesignEdit~\citep{jia2025designedit}} & \cellcolor{LensFullVLMThird}76.95 & 60.25 & 63.85 & 62.95 & 68.19 & 25.00 \\
    \addlinespace[0.9pt]
    \mbox{OmniEraser~\citep{wei2025omnieraser}} & 55.65 & 47.95 & \cellcolor{LensFullVLMThird}77.55 & 57.65 & 58.89 & 22.60 \\
    \addlinespace[0.9pt]
    \mbox{OmniPaint~\citep{yu2025omnipaint}} & 69.20 & \cellcolor{LensFullVLMSecond}66.15 & \cellcolor{LensFullVLMSecond}80.20 & \cellcolor{LensFullVLMSecond}83.20 & \cellcolor{LensFullVLMSecond}73.59 & \cellcolor{LensFullVLMSecond}42.80 \\
    \addlinespace[0.9pt]
    \mbox{GeoRemover~\citep{zhu2026georemover}} & 54.30 & 49.85 & 77.05 & \cellcolor{LensFullVLMThird}78.00 & 62.70 & 25.60 \\
    \addlinespace[0.9pt]
    \mbox{PANDORA~\citep{vo2026pandora}} & 41.93 & 39.28 & 58.17 & 49.50 & 46.16 & 7.21 \\
    \addlinespace[0.9pt]
    \mbox{DORS~\citep{tang2026dors}} & 55.20 & 27.40 & 39.10 & 32.60 & 41.90 & 0.60 \\
    \addlinespace[0.9pt]
    \mbox{ObjectClear~\citep{zhao2026precise}} & 73.95 & \cellcolor{LensFullVLMThird}61.45 & 73.80 & 75.20 & \cellcolor{LensFullVLMThird}71.67 & 28.40 \\
    \addlinespace[0.9pt]
    \mbox{OSOR~\citep{zhou2026osor}} & 73.85 & 60.45 & 75.55 & 73.55 & 71.45 & \cellcolor{LensFullVLMThird}32.60 \\
    \arrayrulecolor{LensFullVLMDivider}
    \specialrule{0.25pt}{3.1pt}{3.6pt}
    \arrayrulecolor{LensFullVLMRule}
    \textbf{LensErase} & \cellcolor{LensFullVLMFirst}\textbf{95.15} & \cellcolor{LensFullVLMFirst}\textbf{92.40} & \cellcolor{LensFullVLMFirst}\textbf{96.35} & \cellcolor{LensFullVLMFirst}\textbf{97.05} & \cellcolor{LensFullVLMFirst}\textbf{95.22} & \cellcolor{LensFullVLMFirst}\textbf{82.80} \\
    \addlinespace[1.8pt]
    \bottomrule[0.6pt]
  \end{tabular*}
  \arrayrulecolor{black}
  \endgroup
\end{table}
\endgroup
\FloatBarrier

\FloatBarrier
\section{Reward Component Ablations}
\label{app:posttraining_ablations}
\label{sec:reward_ablation}
\label{app:reward_component_ablations}

We examine five variants of the reward in Section~\ref{sec:removal_rl}, each omitting one component: Removal, Reconstruction, Preservation, Outside Fidelity, or Cleanliness. The remaining component weights are renormalized. The variant without Removal also disables the removal-aware gate, whereas the other four retain it. All variants start from the same SFT checkpoint, with optimization details given in Appendix~\ref{app:nft_implementation}. Table~\ref{tab:reward_components} reports the VLM metrics together with PSNR and LPIPS.

\begin{table}[!htbp]
  \centering
  \caption{\textbf{Reward-component ablations.}  PSNR and LPIPS are equal-weight means over the six public benchmarks.}
  \label{tab:reward_components}
  \begingroup
  \fontsize{7.5}{8.9}\selectfont
  \setlength{\tabcolsep}{1.6pt}
  \renewcommand{\arraystretch}{1.08}
  \begin{tabular*}{\linewidth}{@{\extracolsep{\fill}}l*{6}{c}@{\hspace{6pt}}*{2}{c}@{}}
    \toprule
    & \multicolumn{6}{c}{\textbf{LensErase-Bench}} & \multicolumn{2}{c}{\textbf{Public Benchmarks}} \\
    \cmidrule(lr){2-7}\cmidrule(l){8-9}
    \raisebox{\dimexpr(\arraystretch\ht\strutbox+\arraystretch\dp\strutbox+\aboverulesep+\belowrulesep+\cmidrulewidth)/2\relax}[0pt][0pt]{\textbf{Model}} & \mbox{\textbf{Removal}\,$\uparrow$} & \mbox{\textbf{Inpainting}\,$\uparrow$} & \mbox{\textbf{Preservation}\,$\uparrow$} & \mbox{\textbf{Cleanliness}\,$\uparrow$} & \mbox{\textbf{Overall}\,$\uparrow$} & \mbox{\textbf{HiFi Removal}\,$\uparrow$} & \mbox{\textbf{PSNR}\,$\uparrow$} & \mbox{\textbf{LPIPS}\,$\downarrow$} \\
    \midrule
    RL-Raw (full reward) & 94.95 & 92.65 & 93.35 & 96.15 & 94.41 & 81.60 & 28.70 & 0.0816 \\
    \midrule
    w/o Removal + gate & 43.35 & 46.10 & 98.70 & 98.10 & 65.92 & 38.80 & 21.85 & 0.1446 \\
    w/o Reconstruction & 90.85 & 91.40 & 96.10 & 95.70 & 92.98 & 78.00 & 28.32 & 0.0791 \\
    w/o Preservation & 94.15 & 91.50 & 89.20 & 96.00 & 93.00 & 74.60 & 28.01 & 0.0817 \\
    w/o Outside Fidelity & 92.90 & 91.70 & 94.40 & 96.40 & 93.66 & 78.00 & 26.69 & 0.0888 \\
    w/o Cleanliness & 90.90 & 90.95 & 96.25 & 95.95 & 92.99 & 78.40 & 28.48 & 0.0803 \\
    \bottomrule
  \end{tabular*}
  \endgroup
\end{table}

\paragraph{Removal component and gate.}
Jointly removing the Removal component and the removal-aware gate causes the largest degradation, with Removal dropping from 94.95 to 43.35 and HiFi Removal from 81.60\% to 38.80\%.

\paragraph{Complementary reward components.}
The remaining components provide complementary benefits. Removing Preservation yields the lowest HiFi Removal among these variants (74.60\%), while removing Outside Fidelity causes the largest PSNR drop, from 28.70 to 26.69\,dB. Removing Reconstruction or Cleanliness also reduces HiFi Removal to 78.00\% and 78.40\%, respectively. Overall, the full reward achieves the best HiFi Removal.

\FloatBarrier

\FloatBarrier
\section{Limitations and Conclusion}
\label{app:limitations}

Despite its broad coverage, LensErase still has several limitations. First, scaling training to ultra-high resolutions remains challenging. Our large-scale data construction relies in part on existing general-purpose image editing models, which remain less reliable for ultra-high-resolution editing, particularly at resolutions such as 4K. This makes it difficult to construct sufficiently large and consistent 4K-scale training data. Future work could explore more reliable high-resolution data construction and more challenging erase scenarios. Overall, LensErase establishes a scalable framework for general-purpose mask-based erasing by combining large-scale data, generative-prior adaptation, reward-based post-training, and high-fidelity post-processing. We hope it provides a useful foundation for future erase-related models and visual editing systems.

\clearpage

\begingroup
\setlength{\intextsep}{6pt}
\makeatletter
\setlength{\@fptop}{0pt}
\makeatother

\section{Qualitative Comparisons}
\label{app:qualitative_comparisons}

\subsection{Comparisons with Mask-based GPT-Image-2.5-Sunburst \citep{openai2026gptimage25} and Flux2-Erase \citep{blackforestlabs2026fluxerase}}
\label{app:qualitative_generative_editors}

\begin{figure}[!ht]
  \centering
  \includegraphics[width=\linewidth]{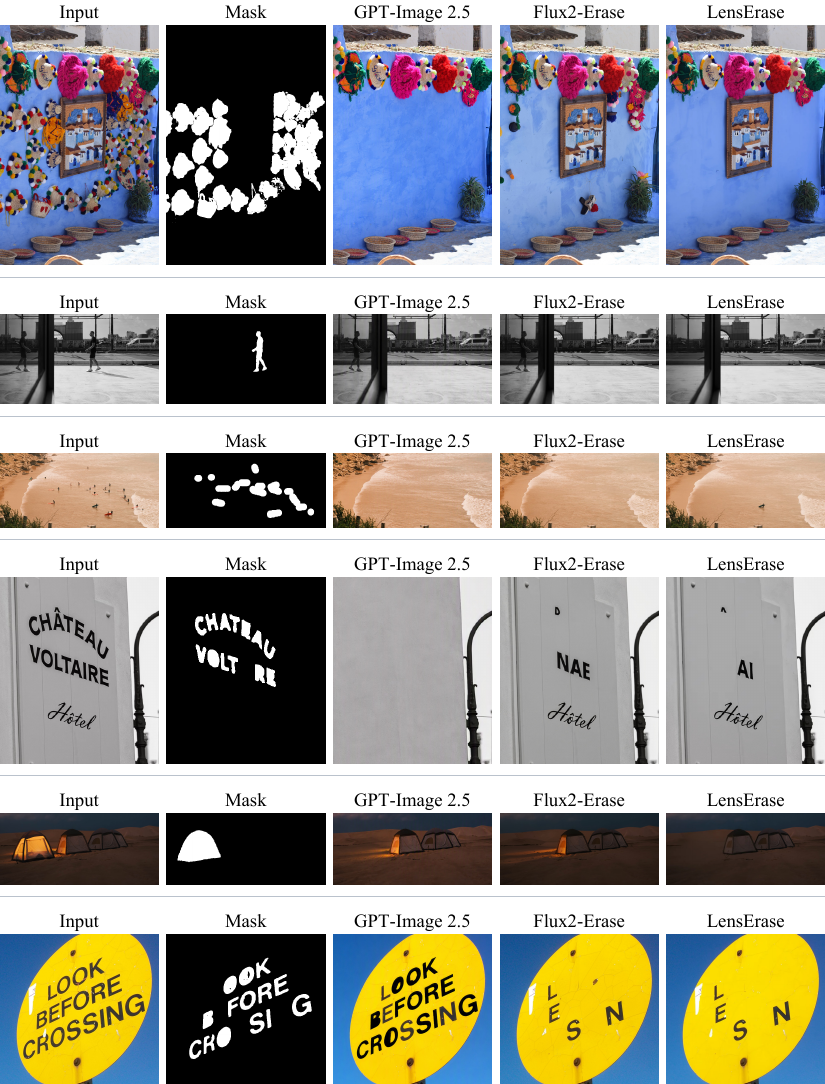}
  \caption{Qualitative comparisons with GPT-Image 2.5 and Flux2-Erase.}
  \label{fig:qualitative_commercial_01}
\end{figure}

\clearpage

\begin{figure}[!ht]
  \centering
  \includegraphics[width=\linewidth]{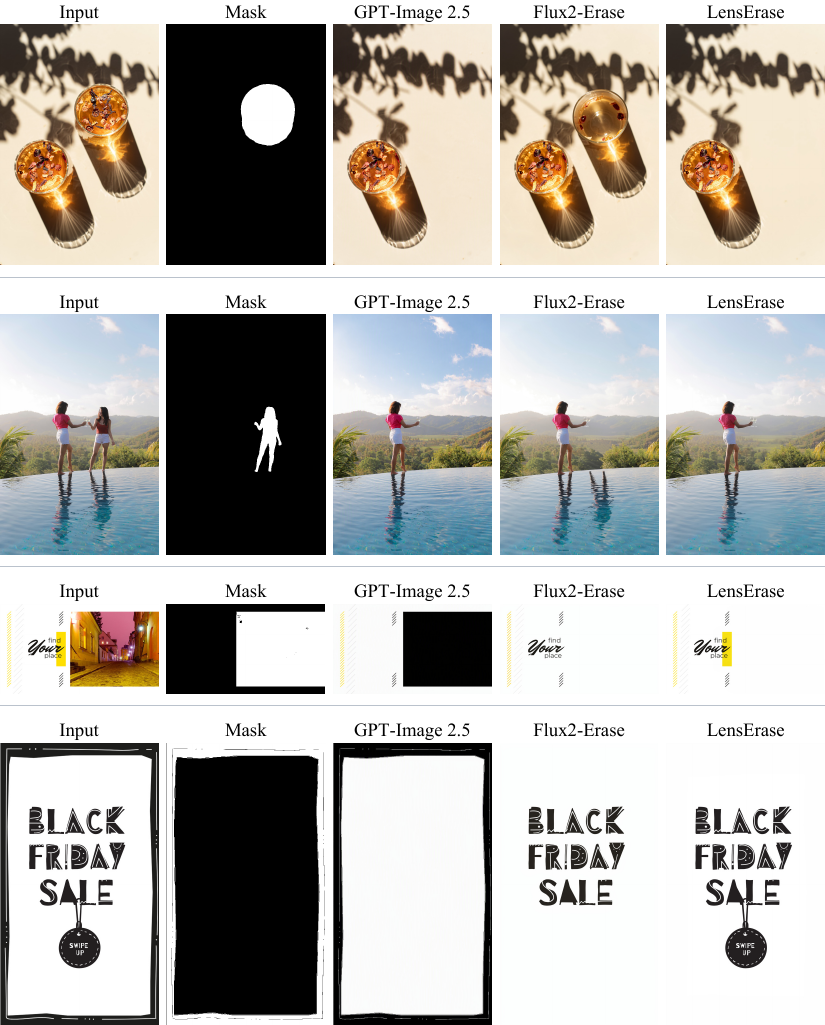}
  \caption{Qualitative comparisons with GPT-Image 2.5 and Flux2-Erase.}
  \label{fig:qualitative_commercial_02}
\end{figure}

\clearpage

\begin{figure}[!ht]
  \centering
  \includegraphics[width=\linewidth]{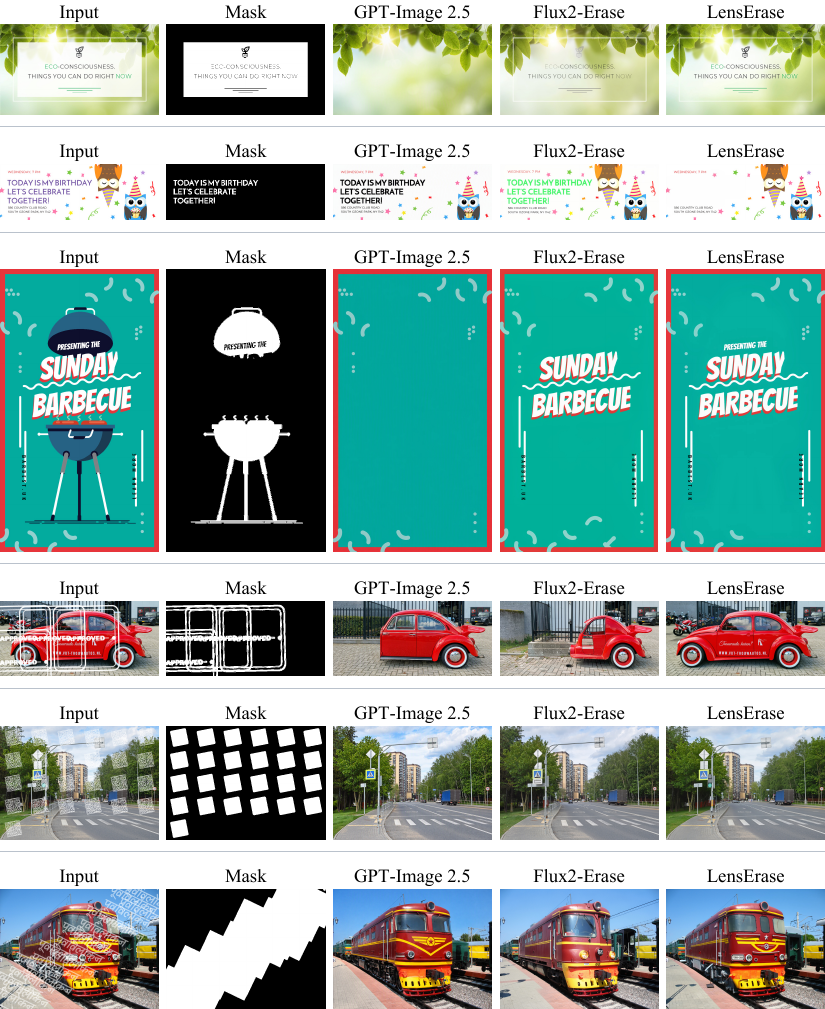}
  \caption{Qualitative comparisons with GPT-Image 2.5 and Flux2-Erase.}
  \label{fig:qualitative_commercial_03}
\end{figure}

\clearpage

\begin{figure}[!ht]
  \centering
  \includegraphics[width=\linewidth]{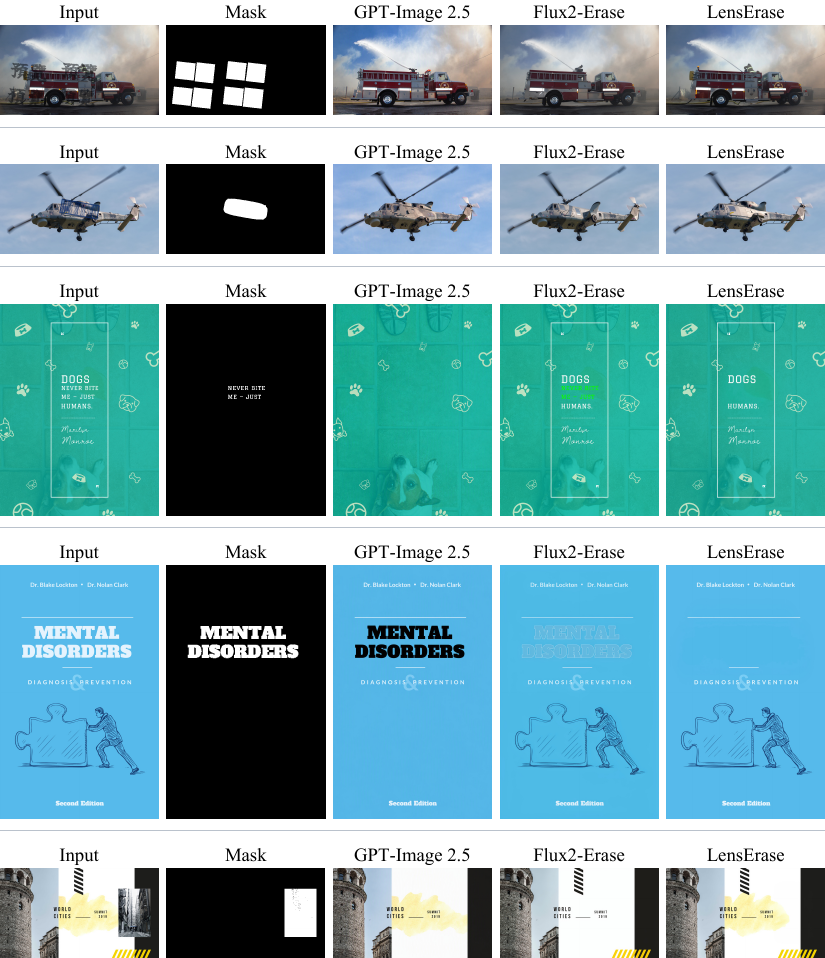}
  \caption{Qualitative comparisons with GPT-Image 2.5 and Flux2-Erase.}
  \label{fig:qualitative_commercial_04}
\end{figure}

\clearpage

\subsection{Comparisons with Object-Removal Baselines}
\label{app:qualitative_removal_models}

\begin{figure}[!ht]
  \centering
  \includegraphics[width=\linewidth]{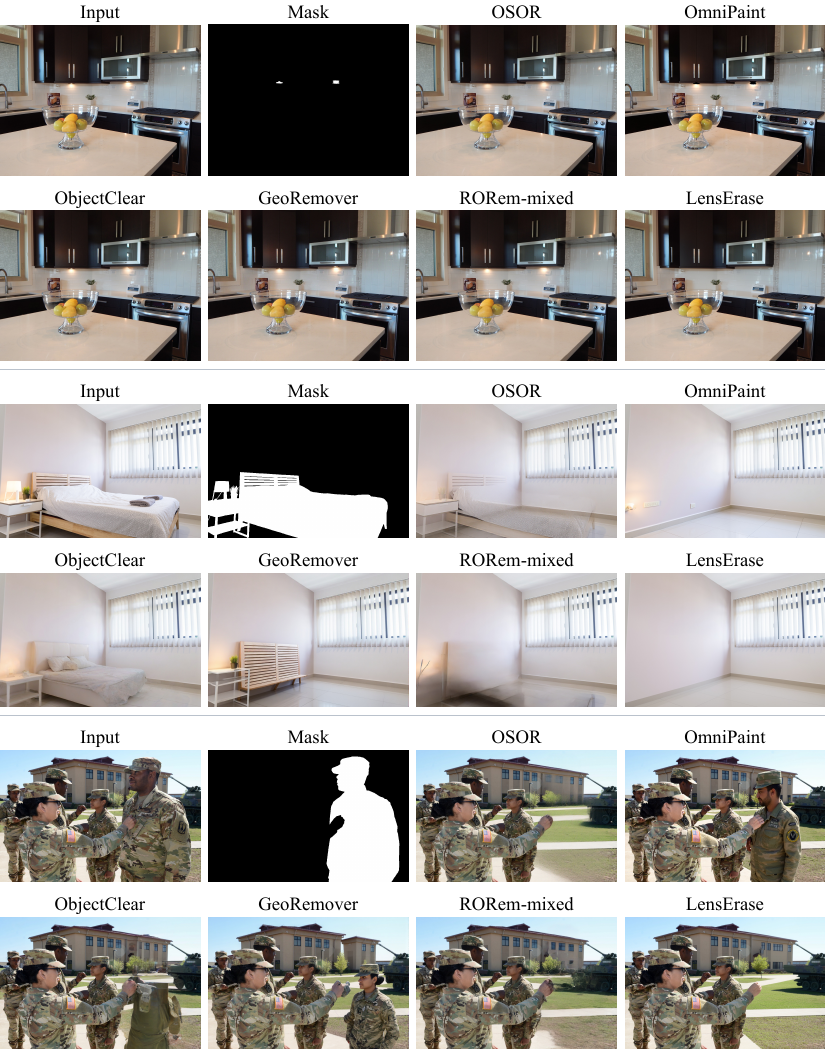}
  \caption{Qualitative comparisons on the General subset.}
  \label{fig:qualitative_open_01}
\end{figure}

\clearpage

\begin{figure}[!ht]
  \centering
  \includegraphics[width=\linewidth]{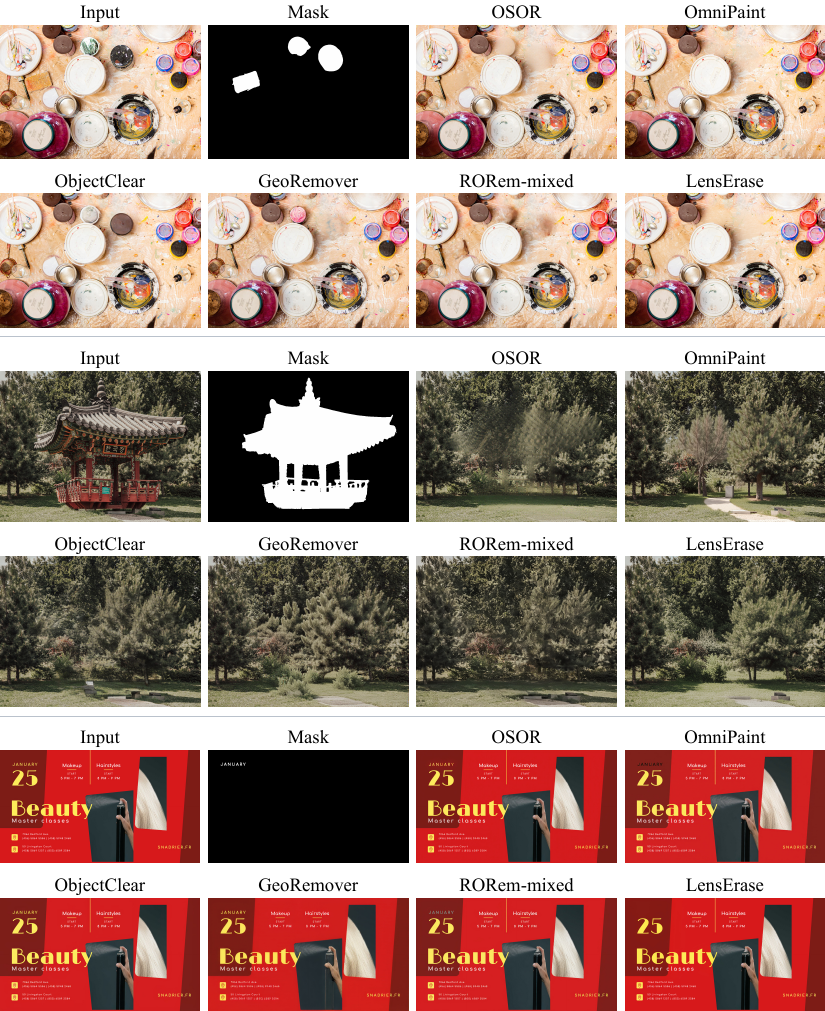}
  \caption{Qualitative comparisons on General images and posters.}
  \label{fig:qualitative_open_02}
\end{figure}

\clearpage

\begin{figure}[!ht]
  \centering
  \includegraphics[width=\linewidth]{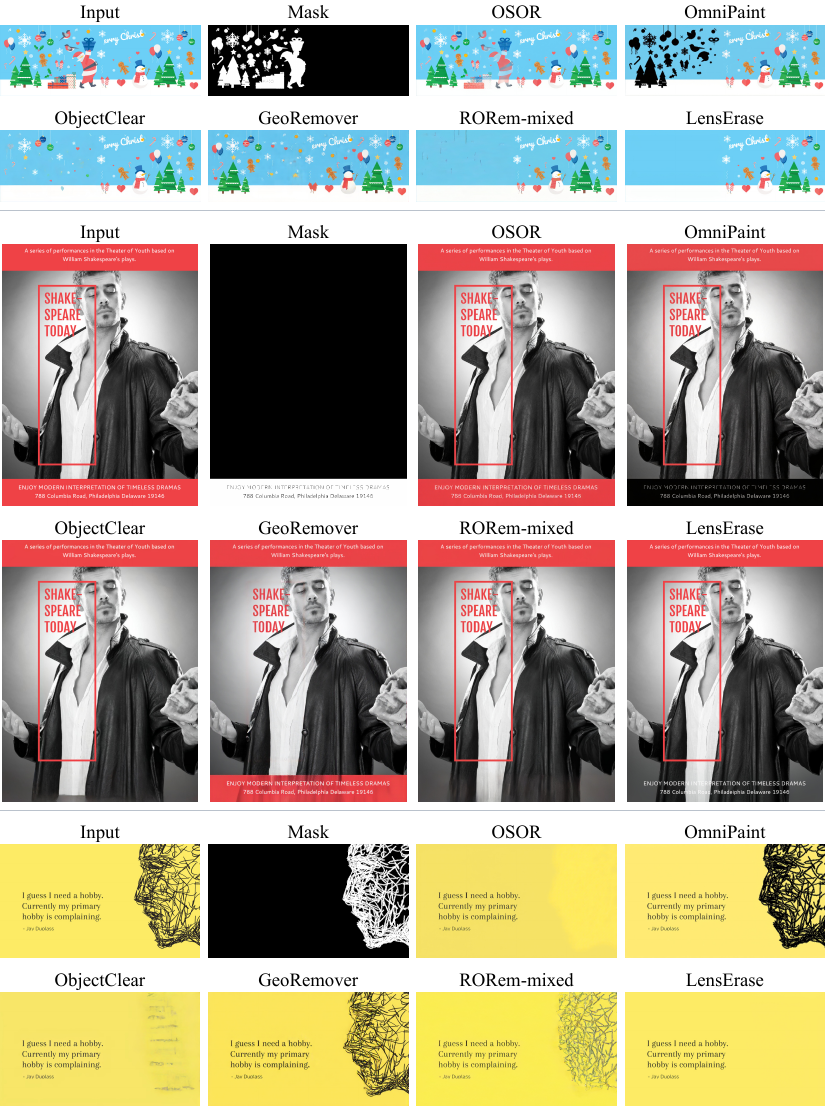}
  \caption{Qualitative comparisons on posters.}
  \label{fig:qualitative_open_03}
\end{figure}

\clearpage

\begin{figure}[!ht]
  \centering
  \includegraphics[width=\linewidth]{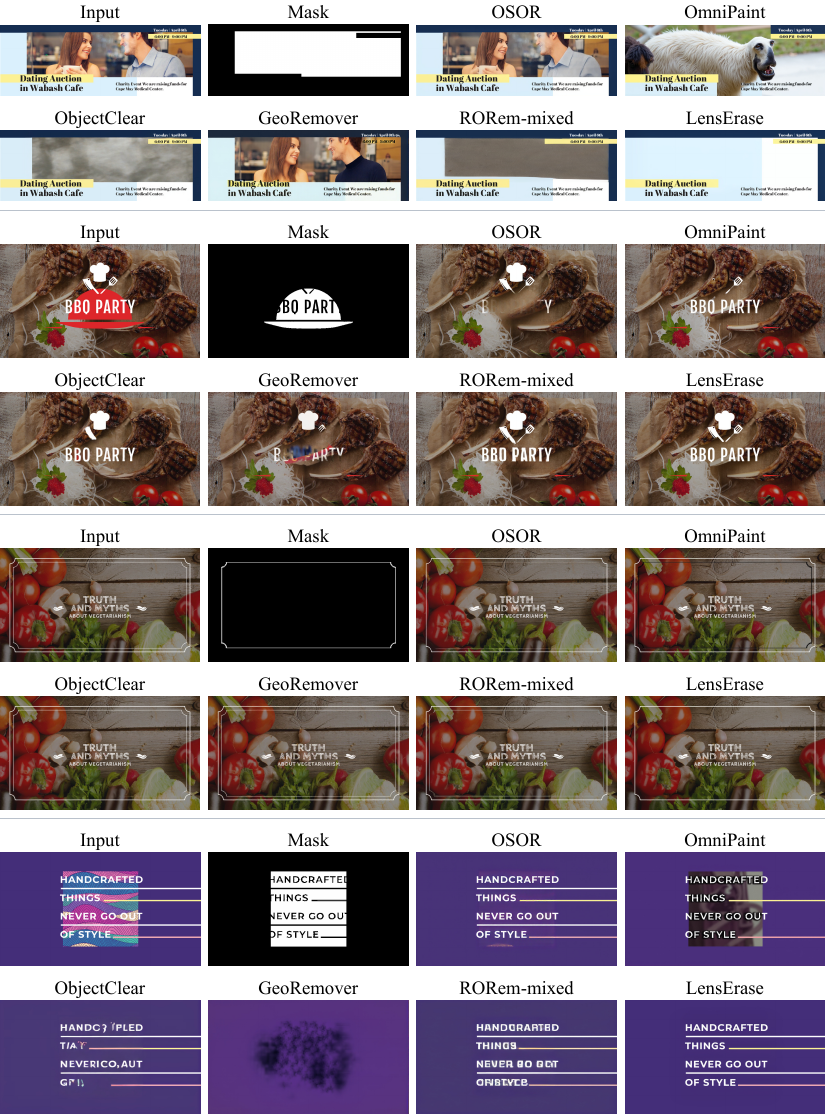}
  \caption{Qualitative comparisons on posters.}
  \label{fig:qualitative_open_04}
\end{figure}

\clearpage

\begin{figure}[!ht]
  \centering
  \includegraphics[width=\linewidth]{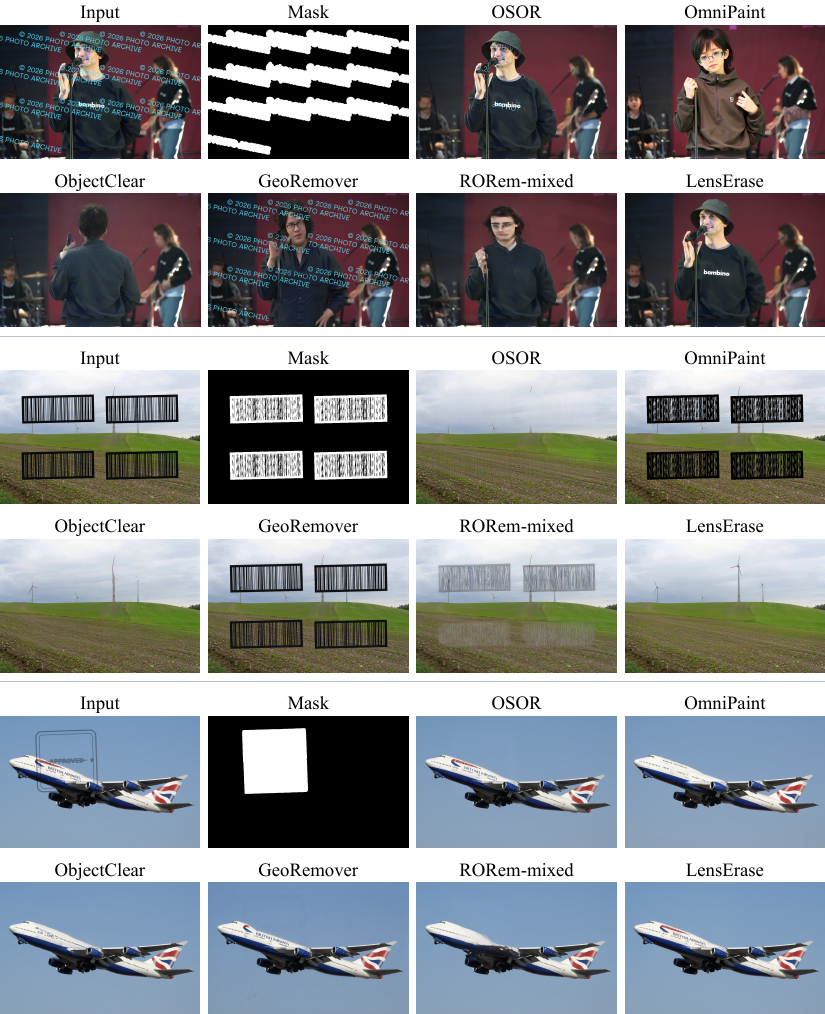}
  \caption{Qualitative comparisons on watermark removal.}
  \label{fig:qualitative_open_05}
\end{figure}

\clearpage

\begin{figure}[!ht]
  \centering
  \includegraphics[width=\linewidth]{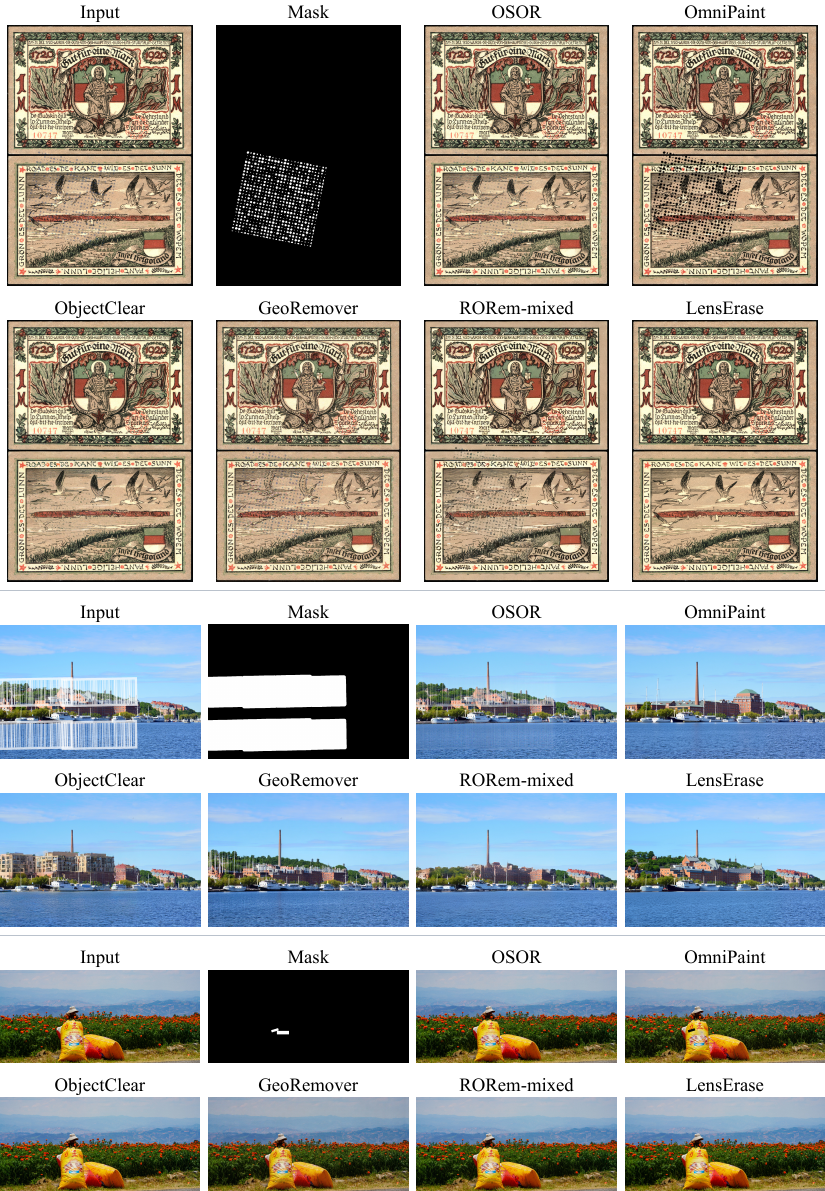}
  \caption{Qualitative comparisons on watermark and visual text removal.}
  \label{fig:qualitative_open_06}
\end{figure}

\clearpage

\begin{figure}[!ht]
  \centering
  \includegraphics[width=\linewidth]{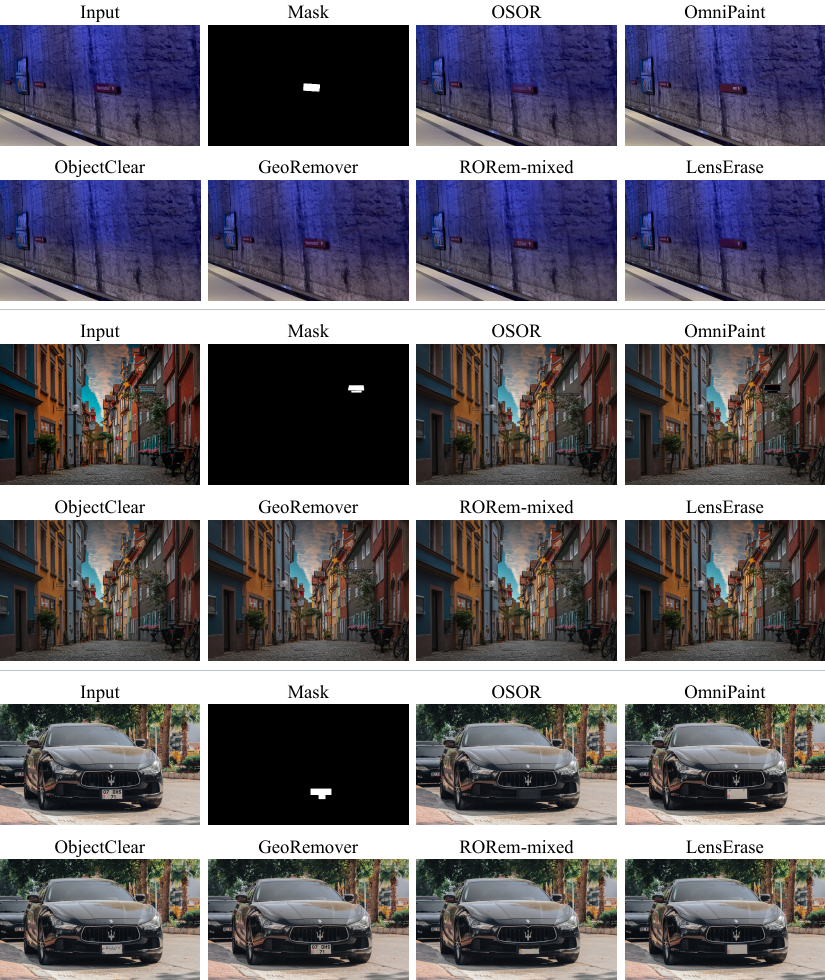}
  \caption{Qualitative comparisons on visual text removal.}
  \label{fig:qualitative_open_07}
\end{figure}

\clearpage

\endgroup

\clearpage

\begingroup
\setlength{\intextsep}{6pt}
\makeatletter
\setlength{\@fptop}{0pt}
\makeatother

\section{Removal with Hand-Drawn and SAM Masks}
\label{app:mask_variants}

Figures~\ref{fig:mask_variants_01}--\ref{fig:mask_variants_04} present additional LensErase results using hand-drawn masks and SAM~2 segmentations~\citep{ravi2024sam2}. The examples include text, transparent objects, multiple targets, and associated shadows or reflections. Each row shows the input image, mask, and LensErase output.

\begin{figure}[!ht]
  \centering
  \includegraphics[width=\linewidth,height=0.92\textheight,keepaspectratio]{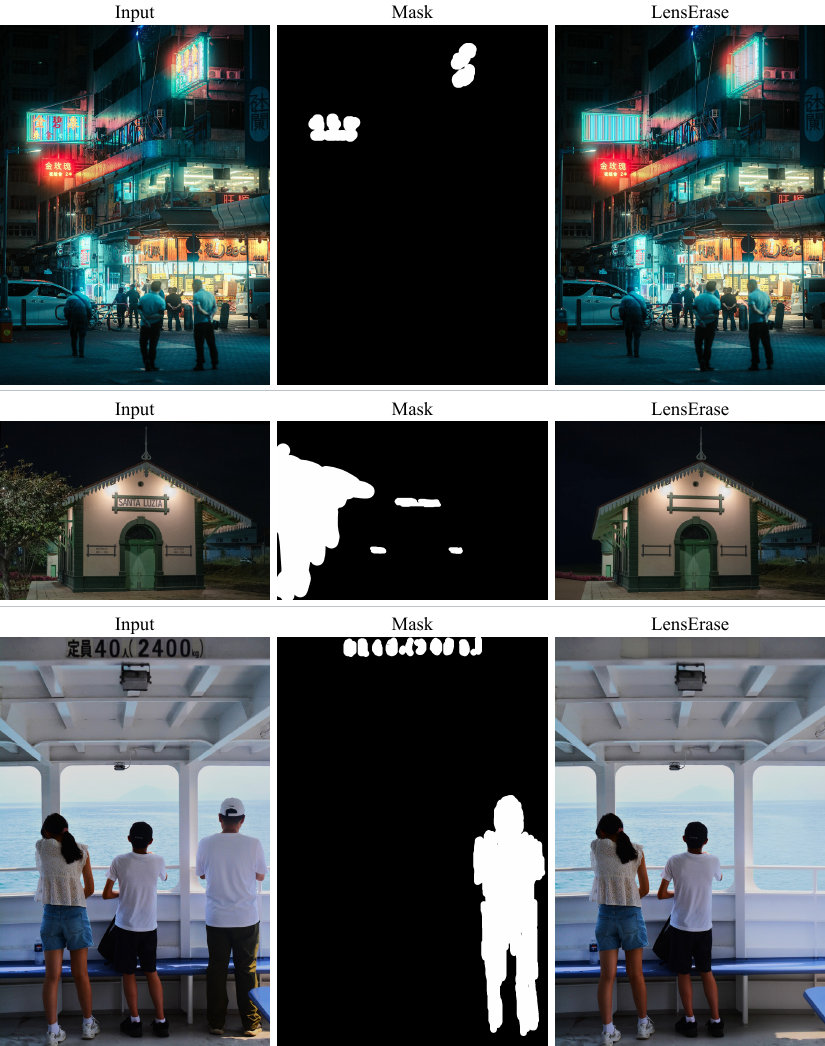}
  \caption{LensErase results with user-provided masks.}
  \label{fig:mask_variants_01}
\end{figure}

\clearpage

\begin{figure}[!ht]
  \centering
  \includegraphics[width=\linewidth,height=0.92\textheight,keepaspectratio]{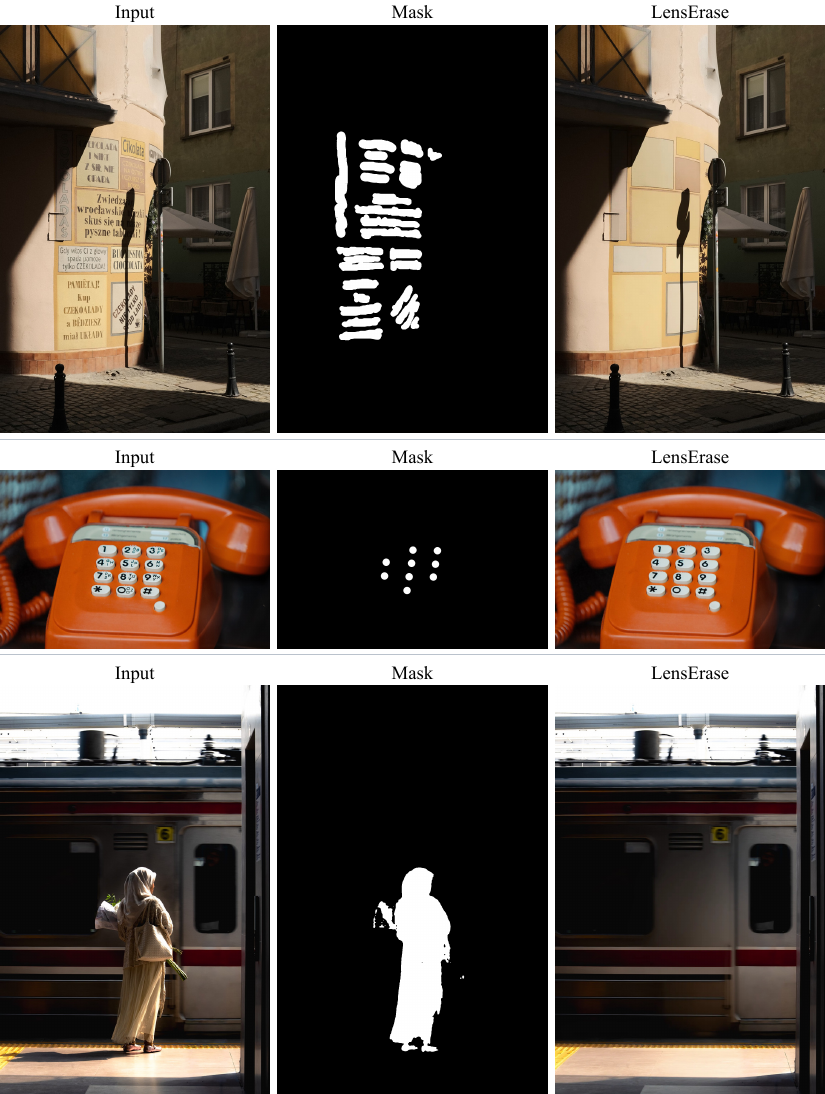}
  \caption{LensErase results with user-provided masks.}
  \label{fig:mask_variants_02}
\end{figure}

\clearpage

\begin{figure}[!ht]
  \centering
  \includegraphics[width=\linewidth,height=0.92\textheight,keepaspectratio]{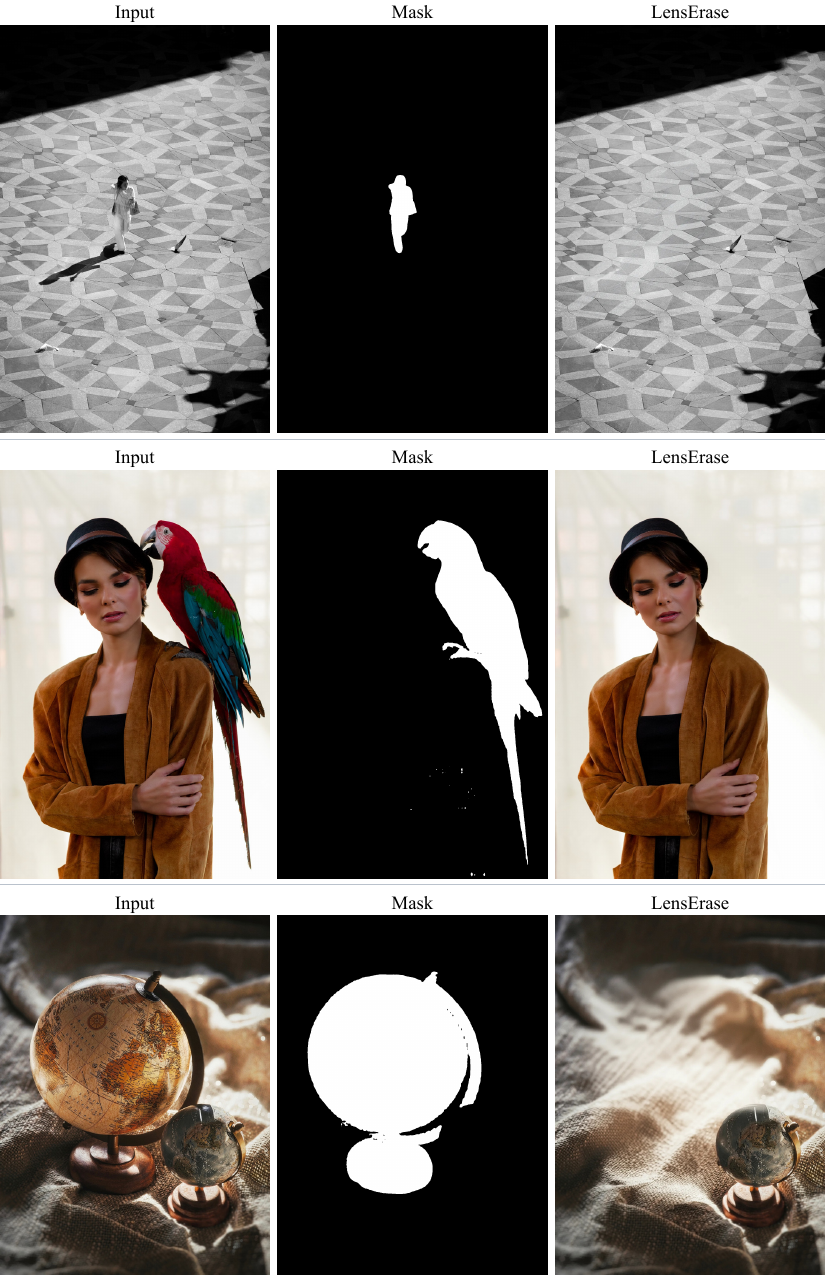}
  \caption{LensErase results with user-provided masks.}
  \label{fig:mask_variants_03}
\end{figure}

\clearpage

\begin{figure}[!ht]
  \centering
  \includegraphics[width=\linewidth,height=0.92\textheight,keepaspectratio]{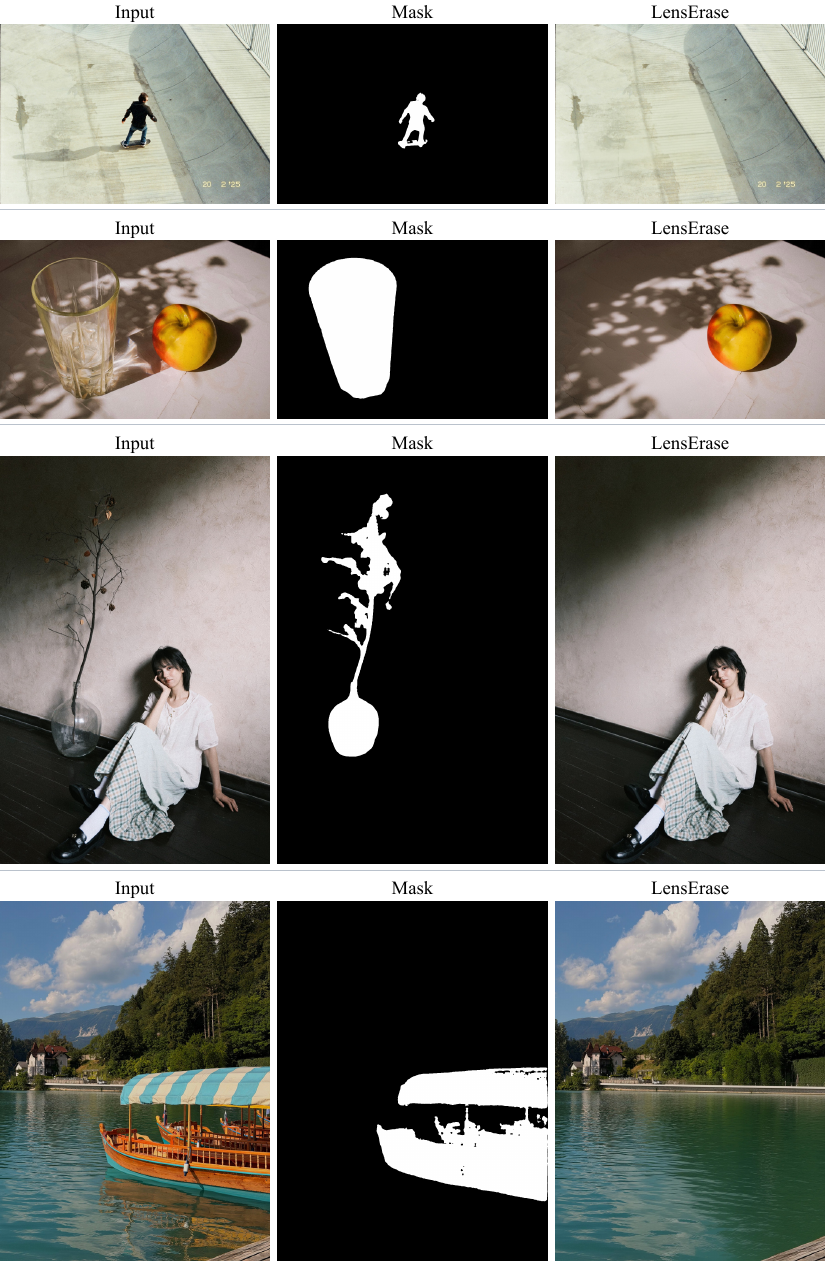}
  \caption{LensErase results with user-provided masks.}
  \label{fig:mask_variants_04}
\end{figure}

\clearpage

\endgroup

\end{document}